%% file: neurips_2026.tex
\documentclass{article}

\PassOptionsToPackage{numbers,sort&compress}{natbib}
\usepackage[preprint]{neurips_2026}

\usepackage[utf8]{inputenc} % allow utf-8 input
\usepackage[T1]{fontenc}    % use 8-bit T1 fonts
\usepackage{hyperref}       % hyperlinks
\usepackage{graphicx}       % resizebox
\usepackage{url}            % simple URL typesetting
\usepackage{booktabs}       % professional-quality tables
\usepackage{amsmath}       % advanced math
\usepackage{amsfonts}       % blackboard math symbols
\usepackage{nicefrac}       % compact symbols for 1/2, etc.
\usepackage{microtype}      % microtypography
\usepackage[table]{xcolor}  % colors

\usepackage{hyperref}
\usepackage[nameinlink,capitalize,noabbrev]{cleveref}

\usepackage{multirow}
\usepackage{wrapfig}
\usepackage{longtable}
\definecolor{ETGreen}{RGB}{190,216,186}

\usepackage{booktabs}
\usepackage{multirow}
\usepackage{colortbl}
\usepackage{xcolor}
\usepackage{caption}
\usepackage{subcaption}   % for sub-captions if needed

\definecolor{groupbg}{RGB}{242,242,242}   % light-grey row for section headers
\definecolor{toolbg}{RGB}{248,248,248}    % subtle grey row for w/ tool
\definecolor{toprule}{RGB}{40,40,40}      % near-black heavy rule
\usepackage{graphicx}
\usepackage{array}
\usepackage{tabularx}
\usepackage{makecell}
\usepackage{colortbl}
\usepackage[table]{xcolor}
\usepackage{enumitem}

\definecolor{groupbg}{RGB}{242,242,242}
\definecolor{headbg}{RGB}{250,250,250}

\usepackage{booktabs, colortbl, xcolor, array, makecell}
\usepackage{longtable}
\usepackage{ragged2e}
\usepackage{xurl}
\usepackage{pdflscape}   % 如果希望横向页面放长表
\usepackage{placeins}    % 用于 \FloatBarrier
\usepackage[most]{tcolorbox}
\tcbuselibrary{listings,breakable}

\definecolor{groupbg}{RGB}{242,242,242}

\newcolumntype{L}[1]{>{\RaggedRight\arraybackslash}p{#1}}
\newcolumntype{C}[1]{>{\Centering\arraybackslash}p{#1}}
\usepackage{caption}
\definecolor{awblue}{HTML}{185FA5}
\definecolor{segreen}{HTML}{0F6E56}
\definecolor{usorange}{HTML}{854F0B}
\definecolor{copurple}{HTML}{534AB7}

\title{Enabling Extensible Embodied Capabilities with Tools}

\author{%
{\small
\begin{tabular}{c}
\textbf{Xueyang Zhou}$^{1}$,
\textbf{Zijia Wang}$^{1}$,
\textbf{Qianjiang Li}$^{2}$,
\textbf{Yibo Hu}$^{1}$,
\textbf{Guiyao Tie}$^{1}$, 
\textbf{Li Wan}$^{1}$,
\textbf{Yidan Liu}$^{3}$, \\
\textbf{Pan Zhou}$^{1,*}$,
\textbf{Lichao Sun}$^{4}$,
\textbf{Yongchao Chen}$^{5,*}$ \\[0.3em]
$^{1}$Huazhong University of Science and Technology \\
$^{2}$Hebei University of Technology,
$^{3}$Tianjin University,
$^{4}$Lehigh University \\
$^{5}$College of AI, Tsinghua University \\[0.3em]
\texttt{\{d202480819, m202572276, u202413536, tgy\}@hust.edu.cn} \\
\texttt{\{u202315903, panzhou\}@hust.edu.cn} \\
\texttt{255273@stu.hebut.edu.cn,
motianjiu@tju.edu.cn} \\
\texttt{lis221@lehigh.edu,
yongchaochen12@gmail.com} \\[0.3em]
$^{*}$Corresponding authors: Pan Zhou and Yongchao Chen
\end{tabular}
}%
}

\begin{document}
\raggedbottom

\maketitle

\vspace{-0.8em}
\begin{center}
\small
\textbf{GitHub:} \url{https://github.com/Zxy-MLlab/EmbodiedTool} \\
\textbf{Webpage:} \url{https://racingemperor.github.io/manip-tool-hub/}
\end{center}
\vspace{0.6em}

\begin{abstract}
  Most existing embodied intelligence methods formulate perception, reasoning, planning, and control within a unified parameterized policy. Yet these capabilities are inherently hierarchical and heterogeneous, making them difficult to reliably learn and modularize within a single model. We propose a \textit{capability externalization} approach that decouples heterogeneous capabilities into independently optimized tools, dynamically invoked at inference time. To this end, we introduce \textbf{E}mbodied \textbf{T}ool \textbf{P}rotocol (\textbf{ETP}), a standardized protocol for embodied tool registration, discovery, invocation, and execution, and curate 100+ validated tools spanning perception, cognition, reasoning, and execution as the tool base. Building on this, we construct \textbf{EmbodiedToolBench} to evaluate both whether tool augmentation improves embodied performance and how well current models use tools across tool-necessity recognition, tool selection, tool execution, and tool-chain composition. Experiments across simulation and real-world platforms confirm that capability externalization consistently improves embodied performance (avg. gain 31\% on EB-ALFRED and 36\% on EB-Navigation), yet reveal a clear boundary: gains are substantial for cognition and perception but are limited for execution-type capabilities. Moreover, our analysis reveals that knowing when, which, and how to invoke tools remains a persistent challenge across all models, thereby highlighting embodied tool competence as a critical direction for future research.
\end{abstract}

\input{sections/introduction}
\input{sections/preliminary}
\input{sections/embodiedtools}
\input{sections/experiments}
\input{sections/related_works}
\input{sections/conclusion}
\nocite{huang2023metatool}
\bibliographystyle{plainnat}
\bibliography{references}
\clearpage
\appendix
\input{sections/appendix}
% \newpage
% \input{checklist.tex}

\end{document}

%% file: sections/introduction.tex
\section{Introduction}

Embodied intelligence requires agents to perceive, reason, and act in 
physical environments through coupled perception, planning, and 
control~\cite{Zen23,Lia25,Sal25}. Recent advances have centered on 
parameterized embodied policies, including vision-language-action 
models~\cite{openvla2024,octo2024,black2024pi0} and hierarchical 
vision-language systems~\cite{Ahn22}, which have been strengthened 
through improved spatial grounding, richer perception, and enhanced 
reasoning~\cite{song2025robospatial,Ao24,chi2023diffusionpolicy}. Despite substantial gains on standard benchmarks, these monolithic 
architectures underperform on long-horizon, compositional, and 
safety-critical tasks, a gap that points to a structural, rather than 
a capacity, limitation.

The core difficulty is that embodied decision-making is inherently 
\textit{hierarchical and heterogeneous}~\cite{Zen23,Wan24e,Lia25,Ma24}: 
high-level planning and low-level control differ fundamentally in their 
functional roles and optimization objectives. Encoding both within a single 
shared parameterization introduces three intrinsic limitations. 
\textit{Opaque coupling}: capabilities are jointly encoded without clear 
attribution, preventing selective invocation or diagnosis. 
\textit{Capability isolation}: competencies acquired by one model remain siloed and cannot be transferred or reused across systems. 
\textit{Entangled optimization}: improving one capability risks degrading 
others due to conflicting gradient signals over shared parameters. Together, 
these limitations motivate \textit{tool-based embodied intelligence}~\cite{yao2023react,schick2023toolformer,huang2023metatool}, an 
approach that externalizes specific capabilities as callable, independently 
optimizable tools that can be registered, discovered, and composed across 
models and tasks.

Yet research on tool-based embodied intelligence remains fragmented, 
lacking a unifying framework to guide design and evaluation. Three fundamental questions remain 
open: \textbf{(1) Protocol:} how should embodied capabilities be 
represented, registered, and invoked in a principled, extensible manner? 
\textbf{(2) Efficacy:} can current models reliably use external tools, and 
how much does tool augmentation improve embodied performance in practice? 
\textbf{(3) Externalization boundary:} which capabilities benefit from 
decoupling into tools, and where does this benefit diminish?

To address these questions, we establish a unified framework centered on 
the \textbf{Embodied Tool Protocol} (ETP), which provides a principled 
foundation for embodied tool registration, discovery, invocation, and 
execution. Grounded in ETP, we curate a tool base of 100+ validated tools spanning 
perception and grounding, cognition and state modeling, reasoning and 
planning, and execution and control. Building on this, we construct \textbf{EmbodiedToolBench} to systematically 
evaluate the tool-use competence of current models across tool-need 
recognition, tool selection, tool execution, and tool-chain composition. Through systematic 
experiments across simulation and real-world platforms, we characterize 
the externalization boundary and identify the key bottlenecks that current 
models face in embodied tool use.

\begin{itemize}[leftmargin=1.2em,labelsep=0.4em,itemsep=2pt,topsep=3pt,
  parsep=0pt,partopsep=0pt]
  \item We introduce \textbf{ETP}, a standardized protocol for embodied 
  tool registration, discovery, invocation, and execution, and for the 
  first time curate 
  100+ validated tools spanning perception, cognition, reasoning, and 
  execution as a unified tool base.

  \item We present \textbf{EmbodiedToolBench}, a benchmark evaluating 
embodied tool use competence across tool-need recognition, tool selection, tool 
execution, and tool-chain composition in planning, navigation, and 
manipulation scenarios.

  \item Through extensive experiments across simulation and real-world 
  platforms, we demonstrate that tool augmentation consistently improves 
  embodied task performance, and reveal systematic bottlenecks in how 
  current models autonomously use embodied tools.
\end{itemize}

%% file: sections/preliminary.tex
\section{Preliminary}
\subsection{Embodied Agent}

We formulate embodied tasks as sequential decision-making problems.
Let $o_t \in \mathcal{O}$ be the observation at timestep $t$,
$l \in \mathcal{L}$ the language instruction,
and $\tau_t := (o_0, a_0, \dots, o_t)$ the interaction history,
with history space $\mathcal{H}_t := \mathcal{O}^{\leq t} \times \mathcal{A}^{<t}$.
A \emph{monolithic} embodied agent defines a single policy:
\begin{equation}
    \pi_{\theta}:\; \mathcal{H}_t \times \mathcal{L} \;\longrightarrow\; \Delta(\mathcal{A}),
    \qquad a_t \sim \pi_{\theta}(\cdot \mid \tau_t,\, l),
    \label{eq:monolithic}
\end{equation}
where $\theta$ jointly encodes \emph{all} capabilities at every decision step.
Let $\mathcal{C} = \{c_1, \ldots, c_K\}$ be the full capability set required across tasks
(e.g.\ low-level control, spatial reasoning, memory),
and $\mathcal{C}(d) \subseteq \mathcal{C}$ the subset demanded by task $d$.
The agent maximizes the expected discounted return via:
\begin{equation}
    \mathcal{J}_{\mathrm{mono}}(\theta) \;:=\;
    \mathbb{E}_{d \sim p(d)}\!\left[
    \mathbb{E}_{\tau \sim p(\tau \mid \pi_\theta,\, d)}\!
    \left[\,\sum_{t=0}^{T} \gamma^{t} r_t\right]\right],\;
    \gamma \in (0, 1],\quad
    \theta^* = \operatorname*{arg\,max}_{\theta}\, \mathcal{J}_{\mathrm{mono}}(\theta).
    \label{eq:mono-obj}
\end{equation}
Because $\theta$ must handle all tasks in $\mathcal{D}$,
the heterogeneous capabilities $\mathcal{C}$ are \emph{entangled} within a single
parameter set, regardless of which subset $\mathcal{C}(d)$ is actually needed per step.

\subsection{Embodied Agent with Tools}

To overcome capability entanglement, we externalize each capability as an
independently optimized \emph{tool} invoked on demand.

\paragraph{Tool collection.}
Define the tool collection
$\mathcal{Z} := \{z_m(\,\cdot\,;\,\phi_m)\}_{m=1}^{M}$,
where each tool $z_m$, parameterized by $\phi_m$, realizes a capability subset
$\mathcal{C}(z_m) \subseteq \mathcal{C}$.
Tools are optimized independently of one another,
and their collective coverage
$\mathcal{C}_{\mathcal{Z}} := \bigcup_{m=1}^{M} \mathcal{C}(z_m)$
replaces the entangled encoding inside $\theta$.

\paragraph{Tool-augmented decision process.}
At each step $t$, the agent selects a tool
$g_t \in \bar{\mathcal{Z}} := \mathcal{Z} \cup \{\perp\}$
(where $\perp$ denotes invoking no tool),
queries it with a generated query $q_\theta(\tau_t, l, g_t)$,
and conditions its action on the returned observation $y_t$.
Formally, the three-stage transition is:
\begin{equation}
    g_t \sim \mu_{\theta}(\cdot \mid \tau_t,\, l), \qquad
    y_t \;:=\; \mathcal{T}_{g_t}\!\left(q_{\theta}(\tau_t,\, l,\, g_t)\right), \qquad
    a_t \sim \pi_{\theta}(\cdot \mid \tau_t,\, l,\, y_t).
    \label{eq:tool-aug}
\end{equation}

\paragraph{Bi-level optimization.}
Learning decomposes into two decoupled objectives.
The \emph{lower-level} objective trains each tool independently
on its designated capability dataset $\mathcal{D}_m$:
\begin{equation}
    \mathcal{J}_{\mathrm{tool}}(\phi_m) \;:=\;
    \mathcal{L}_m(\phi_m;\,\mathcal{D}_m),
    \qquad
    \phi_m^* \;=\; \operatorname*{arg\,min}_{\phi_m}\;
    \mathcal{J}_{\mathrm{tool}}(\phi_m), \quad m \in [M].
    \label{eq:lower}
\end{equation}
The \emph{upper-level} objective trains the orchestration policy $\theta$
to maximize expected return over the full task distribution:
\begin{equation}
    \mathcal{J}_{\mathrm{orch}}(\theta) \;:=\;
    \mathbb{E}_{d \sim p(d)}\!\left[
    \mathbb{E}_{\tau \sim p_{\theta}(\tau \mid d)}\!
    \left[\,\sum_{t=0}^{T} \gamma^{t} r_t\right]\right],
    \qquad
    \theta^* \;=\; \operatorname*{arg\,max}_{\theta}\;
    \mathcal{J}_{\mathrm{orch}}(\theta).
    \label{eq:upper}
\end{equation}
Each tool $\phi_m^*$ is optimized independently for its designated capability
(\cref{eq:lower}), while $\theta^*$ learns to orchestrate the tool collection
for downstream task completion (\cref{eq:upper}).
This yields two key advantages:
\emph{(i)}~capability-specific optimization without cross-task interference, and
\emph{(ii)}~on-demand invocation that avoids embedding all capabilities into $\theta$ at every step.

%% file: sections/embodiedtools.tex
%embodiedtools
\section{Embodied Tools}

\begin{figure*}[t]
\centering
\includegraphics[width=0.999\textwidth]{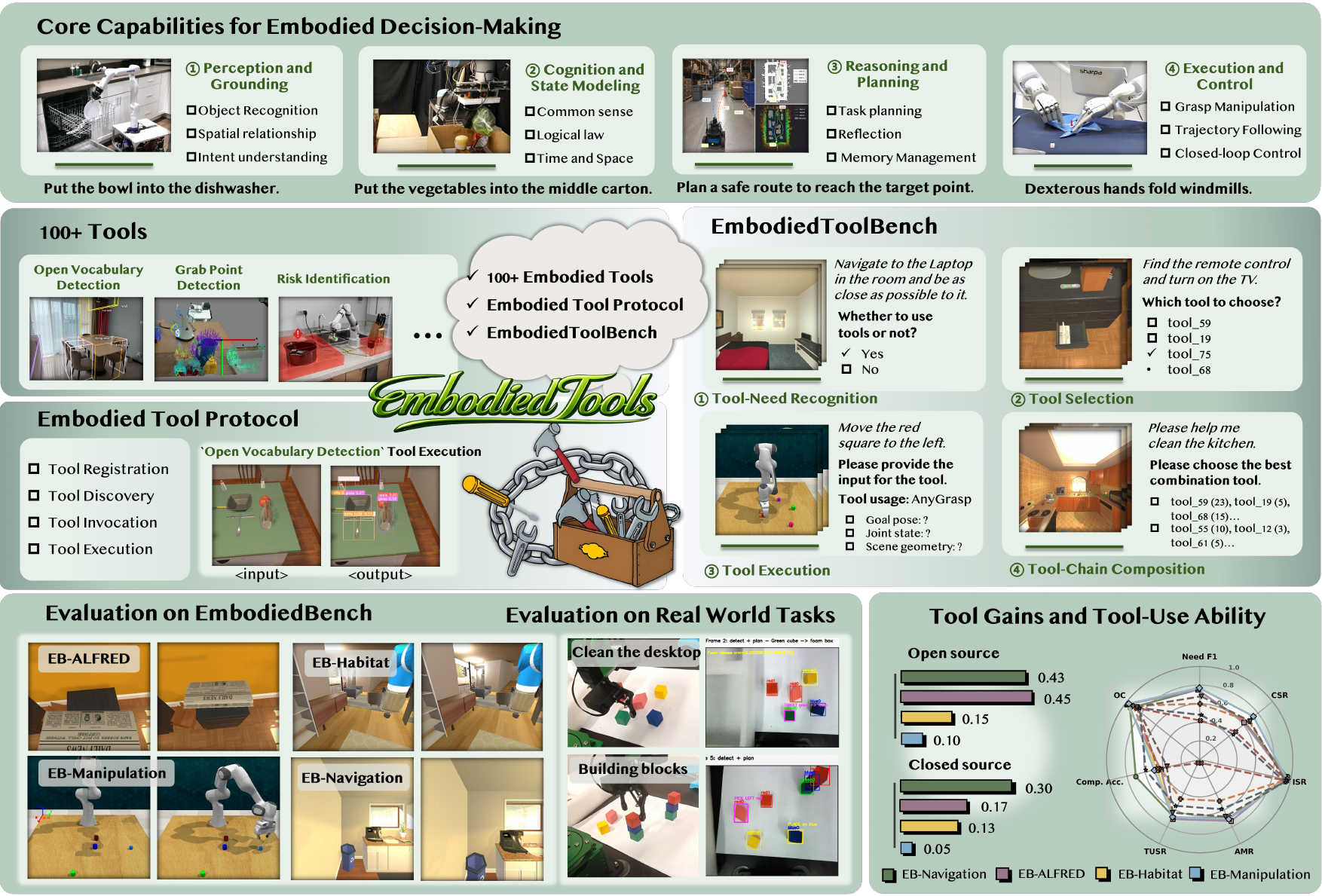}
\caption{Overview of the \textbf{EmbodiedTool}.}
\vspace{-10pt}
\label{fig:embodiedtools_overview}
\end{figure*}

% In this section, we present a protocol-driven framework for externalizing embodied capabilities. We first define embodied tools and introduce the \textit{Embodied Tool Protocol} (ETP) for standardized tool use in embodied decision-making, and then propose \textbf{EmbodiedToolBench} to evaluate models' awareness of and capability in embodied tool use.

\subsection{Embodied Tool Protocol}

Heterogeneous embodied capabilities, ranging from low-level motor control to
high-level spatial reasoning, differ in their interfaces, parameterization, and
execution requirements. Composing them within a single agent therefore requires a
shared abstraction that decouples \emph{what} a capability does from \emph{how}
the agent invokes it. We introduce the \emph{Embodied Tool Protocol} (ETP), which
standardizes capability representation, discovery, and invocation under a unified
framework.

\paragraph{Tool as a capability unit.}
ETP treats each embodied capability as a callable unit with a declared interface.
Formally, a tool $z_m$ is characterized by its input--output spaces $(X_m, Y_m)$,
a realized capability subset $\mathcal{C}(z_m) \subseteq \mathcal{C}$, and an
executable mapping $f_m(\cdot;\phi_m): X_m \to Y_m$.
This interface contract separates capability from implementation: $f_m$ can be
instantiated as a learned model, a classical algorithm, an executable program,
or any hybrid thereof, without changing how the agent interacts with it.

% \paragraph{Registry-based discovery and invocation.}
% Rather than hard-coding tool selection, ETP maintains a capability registry
% \begin{equation}
%     \mathcal{R} = \bigl\{\,(z_m,\,\rho_m,\,\kappa_m)\,\bigr\}_{m=1}^{M},
%     \label{eq:registry}
% \end{equation}
% where $\rho_m$ describes the capability and applicability conditions of $z_m$,
% and $\kappa_m$ specifies its input--output interface constraints.
% Given a decision context, the agent queries $\mathcal{R}$ via $\rho_m$ to
% identify relevant tools, then constructs a query satisfying $\kappa_m$ before
% invoking the selected tool: \emph{discover by capability, invoke by interface}.
% This separation ensures that tool selection is driven by declared functionality
% rather than tool-specific calling conventions, so new tools can be registered
% without modifying the agent policy.

\paragraph{Registry-based discovery and invocation.}
ETP maintains a capability registry
$\mathcal{R} = \{(z_m, \rho_m, \kappa_m)\}_{m=1}^{M}$, where $\rho_m$
describes the capability and applicability conditions of $z_m$, and $\kappa_m$
specifies schema constraints over its input--output spaces $(X_m,Y_m)$.
Given a decision context, the agent queries $\mathcal{R}$ via $\rho_m$ to
identify relevant tools, then constructs a query satisfying $\kappa_m$ before
invoking the selected tool. In other words, the agent discovers tools by
capability and invokes them through their declared interfaces. This separation
ensures that tool selection is driven by declared functionality rather than
tool-specific calling conventions, so new tools can be registered without
modifying the agent policy.

\paragraph{Isolated execution with runtime feedback.}
Each invocation runs in an isolated session that captures the returned output
and any runtime feedback, which the agent can use to detect failures and adapt
subsequent decisions. Session isolation additionally supports concurrent calls
and remote execution, enabling ETP to scale to large and heterogeneous tool
collections without coupling individual tool implementations to the agent.

\subsection{EmbodiedToolBench}

\label{sec:embodiedtoolbench}
\begin{figure*}[t]
\centering
\includegraphics[width=0.99\textwidth]{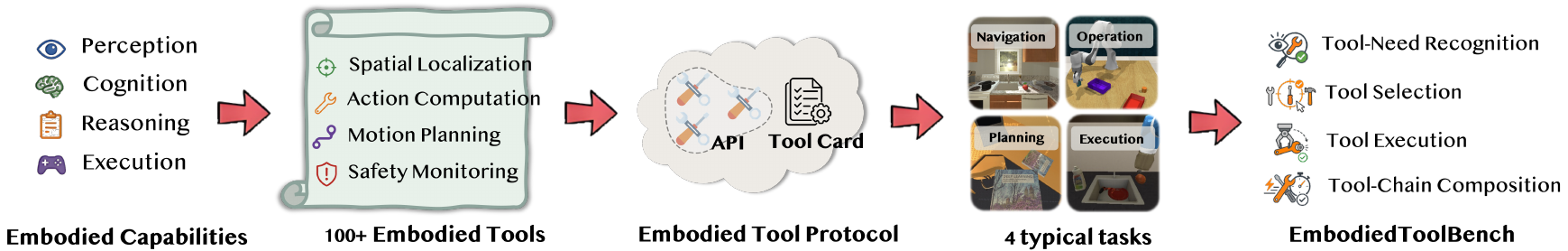}
\caption{Overview of the EmbodiedToolBench collection process.}
\label{fig:tool_collection_pipeline}
\vspace{-10pt}
\end{figure*}
\paragraph{Tool suite construction.} To systematically construct an embodied tool suite, we distill four core capability dimensions from prior work~\cite{driess2023palme,Ahn22}: \textbf{Perception and Grounding}, \textbf{Cognition and State Modeling}, \textbf{Reasoning and Planning}, and \textbf{Execution and Control} (see Appendix~\ref{app:capability_taxonomy} for detailed taxonomy). We then decompose each dimension into fine-grained embodied problems, such as object recognition and spatial grounding, and identify corresponding state-of-the-art methods for each. Each validated method is encapsulated as a unified API, with an LLM generating a structured tool card specifying its functionality, input--output interface, applicability conditions, and usage constraints. This bottom-up process yields a validated suite of more than~100 embodied tools (see Appendix~\ref{app:embodied_tool_collection} for details on the collection process).

\paragraph{Overview.}
Building on the above tool suite and representative embodied environments for
navigation, planning, and manipulation, we introduce \textbf{EmbodiedToolBench}
to evaluate models' embodied tool-use capability from four complementary
perspectives: \emph{tool-need recognition}, \emph{tool selection},
\emph{tool execution}, and \emph{tool-chain composition}
(detailed dataset design principles are provided in
Appendix~\ref{app:embodiedtoolbench}). Each evaluation instance is defined by
a decision state $(l, \tau_t, \mathcal{Z}_{\mathrm{cand}})$, where
$l \in \mathcal{L}$ is the task instruction, $\tau_t$ is the interaction
history, and $\mathcal{Z}_{\mathrm{cand}} \subseteq \mathcal{Z}$ is the
candidate tool set.

\paragraph{Task 1: Tool-Need Recognition.}
This task evaluates whether a model can recognize when tool invocation is
necessary. Given $(l, \tau_t, \mathcal{Z}_{\mathrm{cand}})$, the model outputs
a binary decision $\hat{u} \in \{0,1\}$, where $\hat{u}=1$ indicates that
external tools should be invoked. Positive instances require tool assistance for
successful completion, whereas negative instances can be solved directly.
Performance is reported as accuracy and F1 score against the ground-truth label
$u^\star \in \{0,1\}$.

\paragraph{Task 2: Tool Selection.}
This task evaluates whether a model can identify the most appropriate embodied tool
from a fixed candidate set. Given $(l, \tau_t, \mathcal{Z}_{\mathrm{cand}})$, the model selects the single most appropriate tool $\hat{z} \in \mathcal{Z}_{\mathrm{cand}}$ from four candidates, which include semantically similar tools of the same category and irrelevant tools as distractors, with exactly one ground-truth answer $z^\star$. Performance is measured by the Correct Selection Rate $\mathrm{CSR} = (1/N)\sum_{i=1}^{N}\mathbf{1}(\hat{z}_i = z_i^\star)$.

\paragraph{Task 3: Tool Execution.}
This task evaluates whether a model can correctly invoke a tool and act on its output, and is decomposed into two stages. In \textbf{Stage 1}, given the current task context $(l, \tau_t)$ and the specification of a designated tool $z^\star \in \mathcal{Z}$, the model constructs a query $\hat{x} \in X_{z^\star}$; success requires a well-formed invocation, i.e., $\mathrm{ISR}=\mathrm{Valid}(\hat{x}, z^\star) = 1$. In \textbf{Stage 2}, the model receives the task context together with the tool output $y \in Y_{z^\star}$ and must predict the next action $\hat{a} \in \mathcal{A}$; success requires consistency with the reference action $a^\star$, i.e., $\mathrm{AMR}=\mathrm{Match}(\hat{a}, a^\star) = 1$. We report the success rate of each stage independently, as well as the overall Tool Usage Success Rate $\mathrm{TUSR} = (1/N)\sum_{i=1}^{N}\mathbf{1}\!\left(\mathrm{Valid}(\hat{x}_i, z_i^\star) = 1 \wedge \mathrm{Match}(\hat{a}_i, a_i^\star) = 1\right)$.

\paragraph{Task 4: Tool-Chain Composition.}
This task evaluates whether a model can select the minimal required tools and arrange them in the correct dependency order. Given the task context $(l,\tau_t,\mathcal{Z}_{\mathrm{cand}})$, the model outputs an ordered tool sequence $\hat{P}=(\hat{z}_1,\ldots,\hat{z}_R)$ where each $\hat{z}_r\in \mathcal{Z}_{\mathrm{cand}}$, and $\hat{S}$ denotes the induced tool set. Let $S^\star\subseteq\mathcal{Z}$ denote the ground-truth minimal tool set and $\Omega^\star=\{(z_a,z_b)\}$ the ground-truth ordering constraints, where $(z_a,z_b)$ indicates that $z_a$ must be invoked before $z_b$. We report tool selection quality via \textbf{Accuracy} $\mathrm{ACC}=(1/N)\sum_{i=1}^{N}\mathbf{1}(\hat{S}_i = S_i^\star)$ and \textbf{F1} $\mathrm{F1}=(1/N)\sum_{i=1}^{N}{2|\hat{S}_i\cap S_i^\star|}/({|\hat{S}_i|+|S_i^\star|})$, and the \textbf{Order Consistency Rate} $\mathrm{OCR}=(1/N)\sum_{i=1}^{N}(1/|\Omega_i^\star|)\sum_{(z_a,z_b)\in\Omega_i^\star}\mathbf{1}(z_a,z_b\in\hat{P}_i \land \mathrm{pos}_{\hat{P}_i}(z_a)<\mathrm{pos}_{\hat{P}_i}(z_b))$.

% \paragraph{Task 5: tool efficiency.}
% This task evaluates whether a model can accomplish the task with minimal tool
% overhead. Given $(l, \tau_t, o_t, L)$ with runtime costs provided in each tool's
% description, the model outputs a predicted tool set $\hat{S} \subseteq L$. Let
% $S_{\min}^\star$ denote the oracle-minimal sufficient tool set. The Tool Efficiency
% Score $\mathrm{TES} = (1/N)\sum_{i=1}^{N} \mathbf{1}(\mathrm{Succ}(\tau_i) = 1)
% \cdot |S_{\min,i}^\star| / \max(|\hat{S}_i|, |S_{\min,i}^\star|)$ rewards task
% success while penalizing unnecessary tool calls; the efficiency ratio equals~$1$
% if and only if $\hat{S}_i = S_{\min,i}^\star$.

%% file: sections/experiments.tex
%experiments
\section{Experiments}
In this section, we conduct experiments organized around the following 
research questions, with implementation details provided in 
Appendix~\ref{app:implementation}. We evaluate eight representative open- 
and closed-source models on Embodiedbench~\cite{yang2025embodiedbench} (EB-ALFRED, 
EB-Habitat, EB-Navigation, and EB-Manipulation) and real-world robotic 
platforms. To ensure result reliability, we report multi-run statistical analysis 
on a subset of evaluations in Appendix~\ref{app:statistical_significance}.

\begingroup
\setlength{\tabcolsep}{4.4pt}
\renewcommand{\arraystretch}{1.16}

% requires: \usepackage{booktabs,multirow,colortbl,xcolor,graphicx}
\definecolor{groupbg}{RGB}{244,244,244}
\definecolor{toolbg}{RGB}{247,250,255}
\definecolor{avgbg}{RGB}{218,248,211}

\newcommand{\best}[1]{\textbf{#1}}
\newcommand{\worst}[1]{\underline{#1}}
\newcommand{\gainup}[1]{\makebox[3.2em][c]{$\uparrow\,#1$}}
\newcommand{\gaindown}[1]{\makebox[3.2em][c]{$\downarrow\,#1$}}

\begin{table*}[t]
\centering
\caption{%
Performance comparison across four embodied benchmarks.
\textbf{Bold}: best result per column; \underline{underlined}: worst result; \protect\colorbox{avgbg}{Avg. Gain}: mean improvement of ``w/ tool'' over ``w/o tool''.
}
\label{tab:eb_results}

% ============================================================
% (a) EB-ALFRED and EB-Habitat
% ============================================================
\vspace{2pt}
{\small\textbf{(a)} EB-ALFRED and EB-Habitat}
\vspace{3pt}

\resizebox{\textwidth}{!}{%
\begin{tabular}{@{} l @{\hspace{6pt}}
                c c c c c c c @{\hspace{2pt}}
                c @{\hspace{2pt}}
                c c c c c c c @{}}

\specialrule{1.2pt}{0pt}{2pt}

\multirow{2}{*}{\textbf{Model}}
  & \multicolumn{7}{c}{\textbf{EB-ALFRED}}
  & \multicolumn{1}{c}{}
  & \multicolumn{7}{c}{\textbf{EB-Habitat}} \\

\cmidrule(lr){2-8} \cmidrule(lr){10-16}

& \textbf{Avg} & \textbf{Base} & \textbf{Common} & \textbf{Complex} & \textbf{Visual} & \textbf{Spatial} & \textbf{Long}
&
& \textbf{Avg} & \textbf{Base} & \textbf{Common} & \textbf{Complex} & \textbf{Visual} & \textbf{Spatial} & \textbf{Long} \\

\specialrule{0.8pt}{1pt}{2pt}

\rowcolor{groupbg}
\multicolumn{16}{c}{\textit{Closed-Source MLLMs}} \\[1pt]

GPT-4o
  & 0.48 & 0.58 & 0.52 & 0.52 & 0.42 & 0.50 & 0.36
  & & 0.48 & 0.92 & 0.32 & 0.50 & 0.46 & 0.30 & 0.40 \\
\rowcolor{toolbg}
\quad w/ tool
  & 0.76 & 0.90 & 0.90 & 0.88 & 0.76 & 0.74 & 0.36
  & & 0.64 & 0.96 & 0.72 & 0.60 & 0.52 & 0.62 & 0.40 \\[1pt]

GPT-5
  & 0.62 & 0.70 & 0.60 & 0.66 & 0.56 & 0.68 & 0.52
  & & 0.57 & 0.84 & 0.42 & 0.64 & 0.66 & 0.40 & \best{0.46} \\
\rowcolor{toolbg}
\quad w/ tool
  & 0.81 & 0.90 & 0.90 & 0.88 & 0.78 & 0.84 & 0.54
  & & 0.75 & 0.98 & 0.82 & 0.72 & 0.74 & \best{0.76} & \best{0.46} \\[1pt]

Claude-3.7
  & 0.82 & 0.86 & 0.82 & 0.80 & 0.80 & 0.76 & 0.86
  & & 0.67 & 0.98 & 0.78 & 0.76 & 0.68 & 0.40 & 0.43 \\
\rowcolor{toolbg}
\quad w/ tool
  & 0.91 & \best{0.96} & \best{0.96} & 0.92 & \best{0.90} & 0.86 & 0.86
  & & 0.76 & \best{1.00} & 0.94 & \best{0.82} & 0.70 & 0.66 & 0.43 \\[1pt]

Claude-4.6
  & 0.81 & 0.86 & 0.76 & 0.80 & 0.78 & 0.80 & 0.88
  & & 0.67 & 0.98 & 0.72 & 0.70 & 0.74 & 0.40 & \best{0.46} \\
\rowcolor{toolbg}
\quad w/ tool
  & \best{0.92} & \best{0.96} & \best{0.96} & \best{0.94} & 0.86 & \best{0.90} & \best{0.92}
  & & \best{0.77} & \best{1.00} & \best{0.96} & 0.76 & \best{0.76} & 0.68 & \best{0.46} \\

\midrule

\rowcolor{groupbg}
\multicolumn{16}{c}{\textit{Open-Source MLLMs}} \\[1pt]

Qwen3-32B
  & 0.37 & 0.44 & 0.32 & 0.44 & 0.36 & 0.30 & 0.38
  & & 0.30 & 0.78 & 0.22 & 0.30 & 0.10 & 0.36 & 0.06 \\
\rowcolor{toolbg}
\quad w/ tool
  & 0.77 & 0.92 & 0.88 & 0.88 & 0.78 & 0.78 & 0.38
  & & 0.44 & 0.88 & 0.46 & 0.46 & 0.14 & 0.66 & 0.06 \\[1pt]

Qwen3-8B
  & \worst{0.20} & \worst{0.14} & \worst{0.18} & \worst{0.28} & \worst{0.20} & \worst{0.14} & 0.26
  & & 0.23 & 0.44 & 0.14 & 0.24 & 0.24 & 0.18 & 0.12 \\
\rowcolor{toolbg}
\quad w/ tool
  & 0.72 & 0.88 & 0.86 & 0.88 & 0.72 & 0.70 & 0.26
  & & 0.39 & 0.74 & 0.38 & 0.38 & 0.24 & 0.46 & 0.12 \\[1pt]

Qwen2.5-32B
  & 0.24 & 0.34 & 0.24 & 0.30 & \worst{0.20} & 0.24 & \worst{0.12}
  & & 0.34 & 0.68 & 0.26 & 0.48 & 0.26 & 0.20 & 0.18 \\
\rowcolor{toolbg}
\quad w/ tool
  & 0.66 & 0.88 & 0.74 & 0.88 & 0.66 & 0.70 & \worst{0.12}
  & & 0.49 & 0.90 & 0.46 & 0.58 & 0.28 & 0.56 & 0.18 \\[1pt]

Qwen3.5-35B
  & 0.24 & 0.30 & 0.28 & 0.32 & 0.22 & 0.22 & \worst{0.12}
  & & \worst{0.11} & \worst{0.30} & \worst{0.06} & \worst{0.12} & \worst{0.06} & \worst{0.10} & \worst{0.02} \\
\rowcolor{toolbg}
\quad w/ tool
  & 0.70 & 0.86 & 0.86 & 0.88 & 0.72 & 0.74 & \worst{0.12}
  & & 0.26 & 0.52 & 0.22 & 0.26 & 0.10 & 0.46 & \worst{0.02} \\

\midrule
\rowcolor{avgbg}
\textbf{Avg. Gain}
  & \gainup{0.31} & \gainup{0.38} & \gainup{0.42} & \gainup{0.38} & \gainup{0.33} & \gainup{0.33} & \gainup{0.01}
  & & \gainup{0.14} & \gainup{0.13} & \gainup{0.25} & \gainup{0.10} & \gainup{0.03} & \gainup{0.32} & \gainup{0.00} \\

\specialrule{1.2pt}{1pt}{0pt}
\end{tabular}%
}

\vspace{8pt}

% ============================================================
% (b) EB-Navigation and EB-Manipulation
% ============================================================
{\small\textbf{(b)} EB-Navigation and EB-Manipulation}
\vspace{3pt}

\resizebox{\textwidth}{!}{%
\begin{tabular}{@{} l @{\hspace{6pt}}
                c c c c c c @{\hspace{4pt}}
                c @{\hspace{4pt}}
                c c c c c c @{}}

\specialrule{1.2pt}{0pt}{2pt}

\multirow{2}{*}{\textbf{Model}}
  & \multicolumn{6}{c}{\textbf{EB-Navigation}}
  & \multicolumn{1}{c}{}
  & \multicolumn{6}{c}{\textbf{EB-Manipulation}} \\

\cmidrule(lr){2-7} \cmidrule(lr){9-14}

& \textbf{Avg} & \textbf{Base} & \textbf{Common} & \textbf{Complex} & \textbf{Visual} & \textbf{Long}
&
& \textbf{Avg} & \textbf{Base} & \textbf{Common} & \textbf{Complex} & \textbf{Visual} & \textbf{Spatial} \\

\specialrule{0.8pt}{1pt}{2pt}

\rowcolor{groupbg}
\multicolumn{14}{c}{\textit{Closed-Source MLLMs}} \\[1pt]

GPT-4o
  & 0.46 & 0.58 & 0.50 & 0.47 & 0.33 & 0.43
  & & 0.31 & 0.31 & 0.29 & 0.27 & 0.36 & 0.33 \\
\rowcolor{toolbg}
\quad w/ tool
  & 0.83 & 0.80 & 0.85 & \best{0.85} & 0.80 & 0.83
  & & 0.37 & 0.44 & 0.31 & 0.40 & 0.36 & 0.35 \\[1pt]

GPT-5
  & 0.57 & 0.62 & 0.62 & 0.68 & 0.50 & 0.42
  & & 0.27 & 0.21 & 0.25 & 0.31 & 0.22 & 0.35 \\
\rowcolor{toolbg}
\quad w/ tool
  & \best{0.85} & \best{0.85} & \best{0.87} & \best{0.85} & 0.83 & 0.85
  & & 0.35 & 0.42 & 0.27 & 0.44 & 0.31 & 0.31 \\[1pt]

Claude-3.7
  & 0.55 & 0.67 & 0.62 & 0.63 & 0.50 & 0.35
  & & 0.38 & 0.33 & 0.38 & 0.46 & 0.33 & \best{0.38} \\
\rowcolor{toolbg}
\quad w/ tool
  & 0.83 & \best{0.85} & 0.82 & 0.83 & \best{0.85} & 0.82
  & & \best{0.42} & \best{0.46} & \best{0.44} & \best{0.50} & 0.36 & 0.33 \\[1pt]

Claude-4.6
  & 0.62 & 0.68 & 0.63 & 0.77 & 0.55 & 0.48
  & & 0.37 & 0.40 & 0.35 & 0.40 & 0.36 & 0.37 \\
\rowcolor{toolbg}
\quad w/ tool
  & \best{0.85} & \best{0.85} & 0.85 & \best{0.85} & \best{0.85} & 0.85
  & & 0.38 & 0.40 & 0.35 & 0.42 & \best{0.40} & 0.31 \\

\midrule

\rowcolor{groupbg}
\multicolumn{14}{c}{\textit{Open-Source MLLMs}} \\[1pt]

Qwen3-32B
  & 0.40 & 0.52 & 0.47 & 0.55 & 0.35 & 0.10
  & & 0.20 & 0.21 & 0.25 & 0.17 & 0.19 & 0.19 \\
\rowcolor{toolbg}
\quad w/ tool
  & 0.84 & 0.83 & 0.82 & \best{0.85} & 0.82 & \best{0.87}
  & & 0.25 & 0.27 & 0.25 & 0.31 & 0.17 & 0.22 \\[1pt]

Qwen3-8B
  & 0.42 & 0.62 & 0.48 & 0.45 & 0.32 & 0.22
  & & 0.19 & 0.15 & 0.15 & 0.25 & \worst{0.06} & 0.31 \\
\rowcolor{toolbg}
\quad w/ tool
  & 0.80 & \best{0.85} & 0.85 & 0.77 & 0.75 & 0.77
  & & 0.27 & 0.29 & 0.35 & 0.31 & 0.19 & 0.19 \\[1pt]

Qwen2.5-32B
  & \worst{0.35} & \worst{0.48} & \worst{0.42} & \worst{0.33} & \worst{0.30} & 0.23
  & & 0.17 & \worst{0.13} & \worst{0.13} & 0.23 & 0.11 & 0.25 \\
\rowcolor{toolbg}
\quad w/ tool
  & 0.81 & 0.75 & 0.83 & 0.83 & 0.82 & 0.83
  & & 0.29 & 0.27 & 0.35 & 0.33 & 0.21 & 0.28 \\[1pt]

Qwen3.5-35B
  & 0.39 & 0.50 & \worst{0.42} & 0.48 & 0.47 & \worst{0.07}
  & & \worst{0.15} & \worst{0.13} & 0.17 & \worst{0.15} & 0.17 & \worst{0.17} \\
\rowcolor{toolbg}
\quad w/ tool
  & 0.81 & 0.80 & 0.80 & 0.80 & 0.83 & 0.82
  & & 0.27 & 0.33 & 0.29 & 0.27 & 0.22 & 0.23 \\

\midrule
\rowcolor{avgbg}
\textbf{Avg. Gain}
  & \gainup{0.36} & \gainup{0.24} & \gainup{0.32} & \gainup{0.28} & \gainup{0.40} & \gainup{0.54}
  & & \gainup{0.07} & \gainup{0.13} & \gainup{0.08} & \gainup{0.09} & \gainup{0.05} & \gainup{0.02} \\

\specialrule{1.2pt}{1pt}{0pt}
\end{tabular}%
}

\vspace{-10pt}
\end{table*}

% \vspace{-10pt}
\endgroup
\begin{itemize}[leftmargin=1.2em,labelsep=0.4em,itemsep=2pt,topsep=3pt,parsep=0pt,partopsep=0pt]

\item \textbf{RQ1: Effectiveness of Tool Augmentation.} Do embodied agents equipped with external tools consistently outperform tool-free baselines across both simulation environments and real-world robotic platforms?

\item \textbf{RQ2: Proficiency of Tool Use Behavior.} What extent do current models demonstrate competence in tool-use awareness, tool selection, tool execution, and multi-tool composition?

\item \textbf{RQ3: Task-Specific Amenability to Tool Offloading.} Which types of embodied capabilities derive the greatest benefit from being externalized as tools, and which resist such decomposition?

\end{itemize}

% \subsection{Settings}
% We evaluate embodied tool use on EmbodiedBench, a comprehensive benchmark spanning planning, navigation, and manipulation tasks. Each task is assessed under two conditions: without tools, where the model solves tasks directly, and with tools, where external tools for perception, reasoning, planning, or control are available. Beyond simulation, we conduct three real-world robotic experiments to validate practical effectiveness, evaluating eight representative open- and closed-source models.

\subsection{Does Tool Augmentation Improve Embodied Performance?}
Through the experiment results in Table~\ref{tab:eb_results}, we have gained the following conclusions:

\paragraph{Tool augmentation consistently improves embodied task performance.}
Across the four benchmarks, tool augmentation brings consistent gains, with the largest improvements on EB-ALFRED ($+0.31$) and EB-Navigation ($+0.36$). These tasks require instruction following, visual grounding, planning, and long-horizon decision making, which can be effectively supported by external tools. In EB-ALFRED, all models improve after tool augmentation, with especially large gains for open-source models such as Qwen3-8B ($0.20 \rightarrow 0.72$) and Qwen3-32B ($0.37 \rightarrow 0.77$). EB-Navigation shows a similar pattern, where gains are particularly strong in Visual ($+0.40$) and Long-horizon ($+0.54$) subtasks. These results suggest that tools are most beneficial when embodied tasks can be decomposed into perception, planning, and structured reasoning components.

\paragraph{Tool augmentation narrows the gap between open-source and closed-source models.}
Tool augmentation substantially improves the competitiveness of open-source MLLMs. For example, Qwen3-8B with tools reaches $0.72$ on EB-ALFRED, outperforming GPT-4o without tools ($0.48$), while Qwen2.5-32B with tools achieves $0.81$ on EB-Navigation, surpassing GPT-5 without tools ($0.57$). Meanwhile, stronger closed-source models also benefit from tools: Claude-4.6 with tools achieves the best average scores on EB-ALFRED ($0.92$) and EB-Habitat ($0.77$), and matches the best result on EB-Navigation ($0.85$). This indicates that tool use does not merely compensate for weaker base models, but also complements stronger models by enabling more effective integration of external embodied capabilities.

\paragraph{Tool augmentation remains limited in fine-grained manipulation settings.}
The benefit of tool augmentation is more limited on EB-Habitat and EB-Manipulation. EB-Habitat shows a moderate average gain of $+0.14$, while EB-Manipulation improves by only $+0.07$, much lower than EB-ALFRED and EB-Navigation. In EB-Manipulation, the Spatial subtask improves only slightly ($+0.02$), and some individual models even show minor drops after tool augmentation. A similar saturation effect appears in Long-horizon subtasks of EB-ALFRED ($+0.01$) and EB-Habitat ($+0.00$). These results suggest that current tool interfaces are less effective for tasks requiring precise spatial grounding, contact-rich interaction, and low-level physical control. We discuss these limitations further in Section~\ref{error_analysis}.

\begin{wraptable}{r}{0.58\textwidth}
\vspace{0pt}
\centering
\setlength{\tabcolsep}{4.2pt}
\renewcommand{\arraystretch}{1.12}
\caption{%
  Performance comparison in real-world tasks.
}
\label{tab:real_world_results}
\scriptsize

\resizebox{\linewidth}{!}{%
\begin{tabular}{@{}
  l
  cc
  @{\hspace{4pt}}
  cc
  @{\hspace{4pt}}
  cc
@{}}

\specialrule{1.1pt}{0pt}{2pt}

\multirow{2}{*}{\textbf{Model}}
  & \multicolumn{2}{c}{\textbf{Desktop Cleaning}}
  & \multicolumn{2}{c}{\textbf{Balance Scale}}
  & \multicolumn{2}{c}{\textbf{Building Blocks}} \\

\cmidrule(lr){2-3}\cmidrule(lr){4-5}\cmidrule(lr){6-7}

& \textbf{w/} & \textbf{w/o}
& \textbf{w/} & \textbf{w/o}
& \textbf{w/} & \textbf{w/o} \\

\specialrule{0.8pt}{1pt}{2pt}

GPT-4o        & 8/10 & 0/10 & 6/10 & 0/10 & 6/10 & 0/10 \\
\addlinespace[1pt]
GPT-5         & 7/10 & 0/10 & 8/10 & 0/10 & 6/10 & 0/10 \\
\addlinespace[1pt]
Claude-3.7    & 6/10 & 0/10 & 8/10 & 0/10 & 8/10 & 0/10 \\
\addlinespace[1pt]
Claude-4.6    & 6/10 & 0/10 & 9/10 & 0/10 & 8/10 & 0/10 \\

\midrule

Qwen3-8B      & 6/10 & 0/10 & 4/10 & 0/10 & 5/10 & 0/10 \\
\addlinespace[1pt]
Qwen3-32B     & 6/10 & 0/10 & 8/10 & 0/10 & 5/10 & 0/10 \\
\addlinespace[1pt]
Qwen2.5-32B   & 5/10 & 0/10 & 6/10 & 0/10 & 7/10 & 0/10 \\
\addlinespace[1pt]
Qwen3.5-35B   & 4/10 & 0/10 & 7/10 & 0/10 & 5/10 & 0/10 \\

\specialrule{1.1pt}{1pt}{0pt}

\end{tabular}%
}

\vspace{-10pt}
\end{wraptable}

\begin{figure*}[t]
\centering
\includegraphics[width=1.0\textwidth]{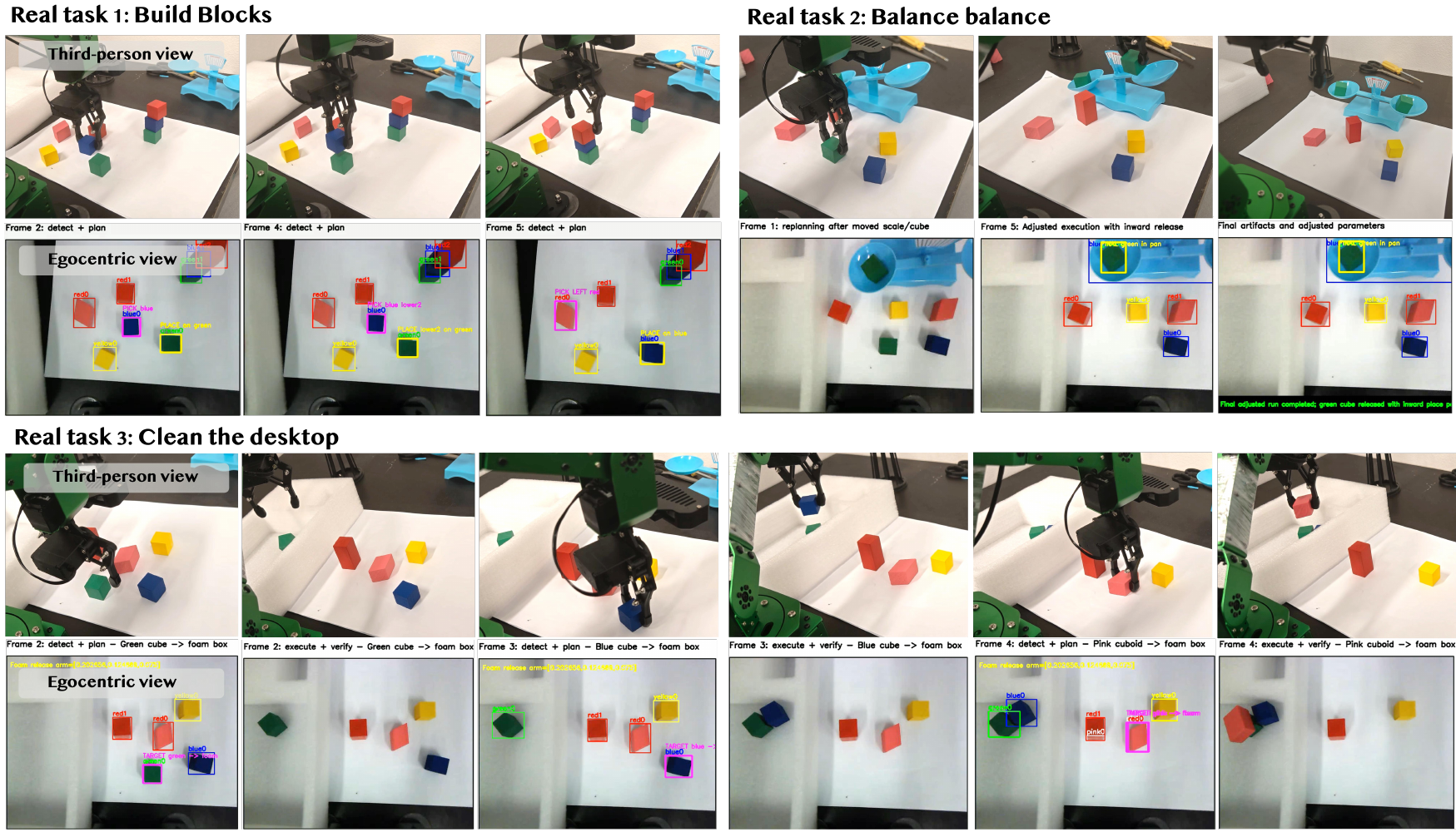}
\caption{Examples of the real-world robot tasks.}
\label{fig:realworld_example}
\vspace{-10pt}
\end{figure*}

\paragraph{Real-world robot experiments.}
We further evaluate tool augmentation on three real-world robotic tasks:
desktop cleaning, balance scale manipulation, and building blocks
(Figure~\ref{fig:realworld_example}). As shown in
Table~\ref{tab:real_world_results}, all models fail without tools, obtaining
0/10 success across all tasks, whereas tool-augmented agents achieve substantial
success rates. GPT-4o performs best on desktop cleaning (8/10), Claude-4.6 leads
on balance scale manipulation (9/10), and Claude-3.7/Claude-4.6 achieve the best
results on structure copying (8/10). These results expose the limitations of current general-purpose models in real-world embodied decision-making, while demonstrating that tool augmentation can substantially improve real-world robotic performance even without any training.

\subsection{How Well Do Current Models Use Embodied Tools?}
% ============================================================
%  Top-conference style table  (NeurIPS / CVPR / ACL)
%  Requirements: booktabs, multirow, colortbl, xcolor, graphicx
% ============================================================

We report the evaluation results on EmbodiedToolBench in 
Table~\ref{tab:tool_eval}, from which we draw the following conclusions. \textbf{(1) Tool-Need Recognition.}
Recognizing when to invoke tools remains challenging across all models. 
Open-source models in particular frequently miss necessary tool invocations, 
with recall dropping as low as 0.31, indicating that current models have 
not yet developed reliable judgment on tool necessity in embodied settings. \textbf{(2) Tool Selection.}
Tool selection remains a non-trivial challenge across all models. The best 
results are achieved by Claude-3.7 and Claude-4.6 (0.74), while weaker 
open-source models drop to as low as 0.45, reflecting difficulty in grounding 
the correct tool among semantically similar candidates. \textbf{(3) Tool Execution.}
Most models achieve consistently high input construction scores (0.94--0.96), 
yet result utilization emerges as the primary bottleneck: GPT-5 achieves 
the highest tool-use success rate (0.70), while several open-source models 
struggle to interpret and act upon returned tool outputs effectively. \textbf{(4) Tool-Chain Composition.}
Composition accuracy correlates with general model capability, ranging from 
0.36 to 0.71. A dominant failure pattern is over-composition: stronger 
models achieve high recall (0.97--0.99) but precision varies widely (0.81--0.93), 
indicating that models cover required tools but frequently include 
superfluous ones, highlighting the difficulty of precise multi-tool 
coordination.

\begingroup
\setlength{\tabcolsep}{5.5pt}
\renewcommand{\arraystretch}{1.20}

\begin{table*}[htb]
\centering
\caption{%
  Evaluation results on EmbodiedToolBench. Abbreviations are defined in Sec.~\ref{sec:embodiedtoolbench}.
}
\label{tab:tool_eval}
\scriptsize

\resizebox{\textwidth}{!}{%
\begin{tabular}{@{} l cccc c ccc ccccc @{}}

\specialrule{1.2pt}{0pt}{2pt}

\multirow{2}{*}{\textbf{Model}}
  & \multicolumn{4}{c}{\textbf{Tool-Need Recognition}}
  & \multicolumn{1}{c}{\textbf{Tool Selection}}
  & \multicolumn{3}{c}{\textbf{Tool Execution}}
  & \multicolumn{5}{c}{\textbf{Tool-Chain Composition}} \\

\cmidrule(lr){2-5}
\cmidrule(lr){6-6}
\cmidrule(lr){7-9}
\cmidrule(lr){10-14}

& \textbf{Acc.} & \textbf{Prec.} & \textbf{Rec.} & \textbf{F1}
& \textbf{CSR}
& \textbf{ISR} & \textbf{AMR} & \textbf{TUSR}
& \textbf{Acc.} & \textbf{Prec.} & \textbf{Rec.} & \textbf{F1.} & \textbf{OC} \\

\specialrule{0.8pt}{1pt}{3pt}

\multicolumn{14}{c}{\cellcolor{groupbg}\textit{Closed-Source MLLMs}} \\[2pt]

GPT-4o
  & 0.65 & \underline{0.67} & \textbf{0.85} & 0.75
  & 0.63
  & \textbf{0.96} & 0.70 & 0.68
  & \textbf{0.71} & \textbf{0.93} & 0.97 & \textbf{0.94} & 0.93 \\

GPT-5
  & 0.66 & 0.68 & 0.84 & 0.75
  & 0.62
  & \underline{0.94} & \textbf{0.74} & \textbf{0.70}
  & 0.43 & \underline{0.81} & \textbf{0.99} & 0.88 & \textbf{0.98} \\

Claude-3.7
  & \textbf{0.69} & 0.72 & 0.82 & \textbf{0.77}
  & \textbf{0.74}
  & 0.95 & 0.71 & 0.68
  & 0.47 & 0.83 & \textbf{0.99} & 0.89 & \textbf{0.98} \\

Claude-4.6
  & \textbf{0.69} & 0.71 & 0.84 & \textbf{0.77}
  & \textbf{0.74}
  & \textbf{0.96} & 0.71 & 0.68
  & 0.50 & 0.83 & \textbf{0.99} & 0.89 & 0.97 \\[2pt]

\midrule

\multicolumn{14}{c}{\cellcolor{groupbg}\textit{Open-Source MLLMs}} \\[2pt]

Qwen3-32B
  & 0.63 & \textbf{0.90} & 0.45 & 0.60
  & 0.67
  & 0.95 & 0.49 & 0.47
  & 0.44 & 0.83 & 0.96 & 0.87 & 0.92 \\

Qwen3-8B
  & \underline{0.48} & 0.68 & \underline{0.31} & \underline{0.42}
  & 0.49
  & \textbf{0.96} & \underline{0.04} & \underline{0.04}
  & 0.46 & 0.84 & 0.92 & \underline{0.86} & 0.87 \\

Qwen2.5-32B
  & 0.63 & 0.74 & 0.63 & 0.68
  & 0.68
  & 0.95 & 0.67 & 0.64
  & \underline{0.36} & 0.83 & 0.94 & 0.87 & 0.88 \\
  
Qwen3.5-35B
  & 0.55 & 0.84 & 0.34 & 0.48
  & \underline{0.45}
  & 0.95 & 0.58 & 0.56
  & 0.59 & 0.91 & \underline{0.91} & 0.90 & \underline{0.85} \\[2pt]

\specialrule{1.2pt}{0pt}{0pt}

\end{tabular}%
}
\end{table*}
\endgroup

\subsection{Which Capabilities Benefit Most from Externalization?}
\begin{figure*}[htb]
\centering
\includegraphics[width=0.999\textwidth]{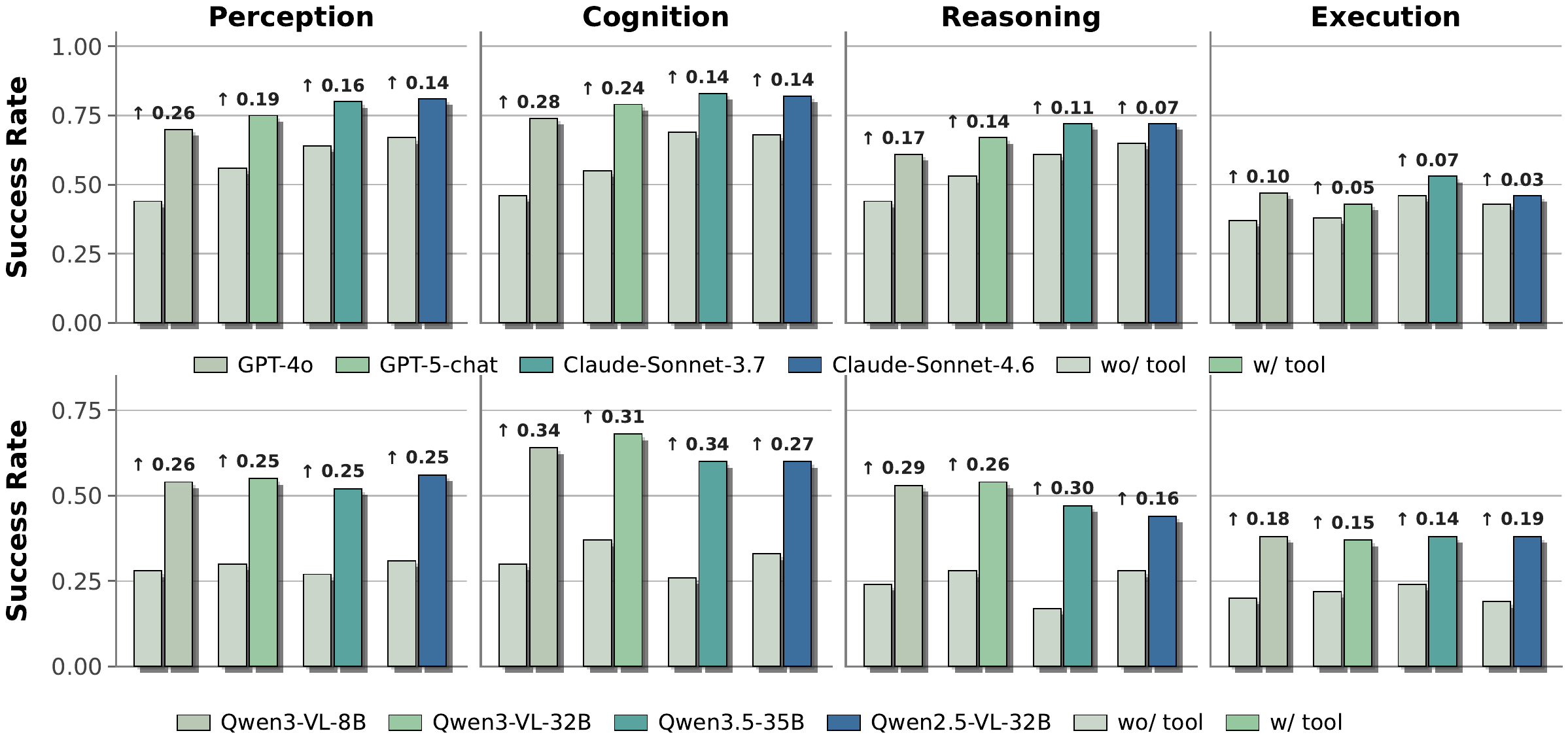}
\caption{Distribution of EmbodiedToolBench across capability dimensions.}
\label{fig:capability_analysis}
\vspace{-15pt}
\end{figure*}

As shown in Figure~\ref{fig:capability_analysis}, tool augmentation improves performance across all four capability dimensions for both closed-source and open-source models. The largest gains occur in Perception and Cognition, reaching +0.26/+0.28 for closed-source models and +0.26/+0.34 for open-source models, respectively, suggesting that external tools are especially effective for visual parsing and knowledge-driven tasks. Reasoning also improves consistently, e.g., +0.17 for GPT-4o and +0.29 for Qwen2.5-VL-32B, indicating that tools can complement internal inference. In contrast, Execution shows smaller and more variable gains, ranging from +0.03 to +0.10 for closed-source models and +0.14 to +0.19 for open-source models, reflecting its reliance on precise input construction and tool generalization.

\subsection{Inference Time Overhead Analysis}
% Requires: booktabs, multirow, xcolor, colortbl
Table~\ref{tab:task_level_tool_comparison} reports the average per-task inference time across four embodied benchmarks, comparing settings with and without tool augmentation. Overall, the latency overhead introduced by tool usage remains acceptable in most cases. On EB-ALFRED and EB-Habitat, the additional cost is generally modest, typically within 10 to 30 seconds (e.g., Claude-3.7 on EB-ALFRED: 38.88s $\rightarrow$ 63.36s; Claude-4.6 on EB-Habitat: 39.60s $\rightarrow$ 49.68s), suggesting that tool augmentation can be integrated without substantially disrupting inference efficiency in these settings. By contrast, EB-Manipulation presents a more challenging profile, where several models incur latency increases exceeding 30 seconds. This is mainly because manipulation tasks invoke tools with higher inference costs and are more affected by network latency during execution. These results point to an important direction for future work: given the stringent real-time requirements of embodied tasks, agents must not only select tools accurately but also reason about their temporal cost.

\begingroup
\setlength{\tabcolsep}{5.2pt}
\renewcommand{\arraystretch}{1.18}
\newcommand{\toolbg}{\cellcolor{blue!8}}

\begin{table*}[htb]
\centering
\caption{%
  Task-level inference time on EmbodiedToolBench with and without tool augmentation.
}
\label{tab:task_level_tool_comparison}
\scriptsize

\resizebox{\textwidth}{!}{%
\begin{tabular}{@{} ll cccc cccc @{}}

\specialrule{1.2pt}{0pt}{2pt}

\multirow{2}{*}{\textbf{Task}} 
& \multirow{2}{*}{\textbf{Setting}}
& \multicolumn{4}{c}{\textbf{Closed-Source MLLMs}}
& \multicolumn{4}{c}{\textbf{Open-Source MLLMs}} \\

\cmidrule(lr){3-6}
\cmidrule(lr){7-10}

&
& \textbf{GPT-4o}
& \textbf{GPT-5}
& \textbf{Claude-3.7}
& \textbf{Claude-4.6}
& \textbf{Qwen3-8B}
& \textbf{Qwen3-32B}
& \textbf{Qwen3.5-35B}
& \textbf{Qwen2.5-32B} \\

\specialrule{0.8pt}{1pt}{3pt}

\multirow{2}{*}{EB-ALFRED}
& w/o tool
& 50.40s & 67.68s & 38.88s & 38.88s
& 51.12s & 63.36s & 59.76s & 56.16s \\
& \toolbg w/ tool
& \toolbg 69.12s & \toolbg 106.56s & \toolbg 63.36s & \toolbg 54.00s
& \toolbg 76.32s & \toolbg 87.84s & \toolbg 87.84s & \toolbg 72.00s \\[2pt]

\midrule

\multirow{2}{*}{EB-Habitat}
& w/o tool
& 46.80s & 67.68s & 38.88s & 39.60s
& 36.72s & 61.92s & 46.80s & 56.16s \\
& \toolbg w/ tool
& \toolbg 57.60s & \toolbg 75.60s & \toolbg 44.64s & \toolbg 49.68s
& \toolbg 53.28s & \toolbg 74.16s & \toolbg 72.72s & \toolbg 87.12s \\[2pt]

\midrule

\multirow{2}{*}{EB-Navigation}
& w/o tool
& 64.08s & 74.16s & 53.28s & 55.44s
& 60.48s & 59.76s & 66.24s & 64.80s \\
& \toolbg w/ tool
& \toolbg 61.20s & \toolbg 66.24s & \toolbg 74.16s & \toolbg 76.32s
& \toolbg 57.60s & \toolbg 63.36s & \toolbg 59.04s & \toolbg 54.00s \\[2pt]

\midrule

\multirow{2}{*}{EB-Manipulation}
& w/o tool
& 46.51s & 88.85s & 98.14s & 102.17s
& 39.89s & 50.40s & 50.83s & 79.63s \\
& \toolbg w/ tool
& \toolbg 56.95s & \toolbg 55.58s & \toolbg 91.66s & \toolbg 92.52s
& \toolbg 72.29s & \toolbg 99.65s & \toolbg 86.47s & \toolbg 87.19s \\[2pt]

\specialrule{1.2pt}{0pt}{0pt}

\end{tabular}%
}
\end{table*}
\endgroup

\subsection{Error Analysis}
\label{error_analysis}
\begin{figure*}[htb]
\centering
\includegraphics[width=0.999\textwidth]{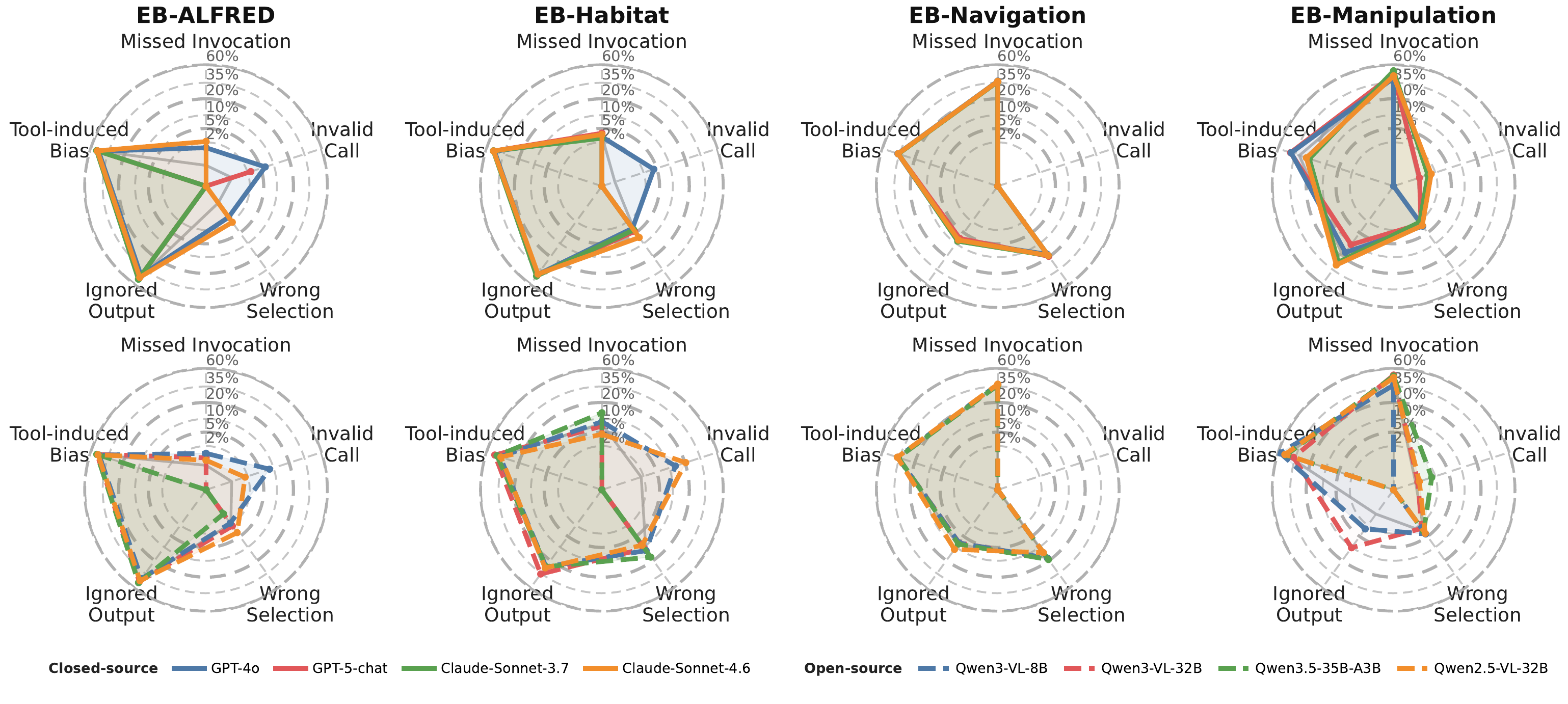}
\caption{Error analysis on EmbodiedBench.}
\label{fig:error_analysis}
\vspace{-15pt}
\end{figure*}

We conduct a fine-grained error analysis across four benchmarks, categorizing failures into five types: Missed Tool Invocation, Invalid Tool Call, Wrong Selection, Ignored Output, and Tool-induced Bias (Figure~\ref{fig:error_analysis}). Missed Tool Invocation is most severe in action-intensive tasks (36\% and 45\% of error cases in EB-Navigation and EB-Manipulation), while Ignored Output dominates in perception-heavy settings (47\% and 43\% in EB-ALFRED and EB-Habitat). Tool-induced Bias remains consistently non-trivial, peaking at 56.5\% for Qwen3-VL-8B in EB-Manipulation. Wrong Selection and Invalid Tool Call are relatively minor overall, though open-source models show notably higher rates in EB-Habitat (e.g., Qwen3.5-35B-A3B: 17.0\%; Qwen2.5-VL-32B: 20.6\%), reflecting weaker tool-interface understanding than closed-source counterparts.

%% file: sections/related_works.tex
%related_works
\section{Related Work}
\label{app:related_work}

Embodied agents require heterogeneous capabilities, including perception, reasoning, planning, control, memory, and adaptation, to operate in open-world environments \cite{Zen23,Wan24e,Lia25,Sal25}. Prior work enhances these capabilities through hierarchical decision-making \cite{Ahn22,Sin22,Hua22b}, scene and spatial representations \cite{Che22,Hua22,Ran23,Wer24,Boo24}, and integrated policy, planning, and control frameworks \cite{Ma24,Mon25,Liu23b,Che23c,Ao24}. However, such model-centric methods remain limited in coverage, reuse, robustness, and composition, especially in long-horizon and safety-critical settings \cite{Wan24e,Lia25,Sal25}. Inspired by LLM-agent tool use \cite{Qu24}, recent embodied systems expose robot capabilities as callable tools or protocol-based services \cite{Mow24,Roy24,Gua25,Ma25}. Yet they lack a standardized embodied tool protocol, a validated full-stack tool suite, and systematic evaluation of tool selection and coordination, motivating our \emph{EmbodiedTool} framework (see Appendix~\ref{app:related_work} for an extended discussion of related work).

%% file: sections/conclusion.tex
%conclusion
\section{Conclusion}
\label{conclusion}
In this paper, we introduce \textbf{ETP}, a standardized framework for 
embodied tool registration, discovery, invocation, and execution, and curate 
100+ validated tools spanning perception, cognition, reasoning, and execution. 
Building on this, we present \textbf{EmbodiedToolBench} to systematically 
evaluate tool-use competence across tool-need recognition, tool selection, 
tool execution, and tool-chain composition. Experiments demonstrate that 
capability externalization consistently improves embodied performance, yet 
reveal that tool competence rather than tool availability is the central 
bottleneck toward truly capable embodied agents.

\paragraph{Limitation.} Despite consistent gains, embodied tool use remains challenging due to failures in tool selection, invocation timing, and multi-tool coordination in dynamic environments. Future work should focus on adaptive strategies that balance task success, invocation cost, and error propagation.

%% file: sections/appendix.tex
\section{Limitations}
\label{app:limitations}
Embodied tool use remains challenging for current models. Models may still fail to decide when tools are needed, select appropriate tools, construct valid inputs, or coordinate multiple tools in long-horizon tasks. These failures are more critical in embodied settings, where observations are dynamic, actions affect the physical world, and tool outputs must be grounded in changing environments. Future work should study how embodied tools can be independently optimized, verified, and adapted across diverse tasks and environments. Tool invocation also introduces extra execution overhead. Developing adaptive tool-use strategies that balance task success, invocation cost, and error propagation is an important direction for future research.

\section{Broader Impacts}
\label{app:broader_impacts}

Tool-augmented embodied agents may improve the modularity, reproducibility, and reliability of embodied AI systems by allowing agents to reuse specialized external capabilities rather than relying solely on a monolithic policy. This can help diagnose model failures, support more transparent evaluation, and facilitate the development of safer embodied systems. Our released assets are intended for research and evaluation in controlled settings, and all tools used in our experiments are validated before evaluation.

\section{Related Work Supplement}
\label{app:related_work}
\paragraph{Embodied agents.}
Embodied agents must coordinate a broad spectrum of capabilities to operate in open-world environments, including perception and scene understanding, spatial reasoning, task planning, action sequencing, motion control, memory, and feedback-driven adaptation \cite{Zen23,Wan24e,Lia25,Sal25}. Recent embodied-agent research has therefore moved beyond isolated perception or control modules toward hierarchical decision-making systems that couple high-level cognition with low-level execution \cite{Zen23,Lia25}. Representative works such as SayCan, ProgPrompt, and Inner Monologue illustrate this trend by combining language understanding with affordance grounding, task decomposition, and interactive replanning for long-horizon embodied tasks \cite{Ahn22,Sin22,Hua22b}. These studies collectively suggest that embodied decision-making is fundamentally a heterogeneous capability integration problem rather than a single-task prediction problem.

\paragraph{Expanding embodied capabilities within the model.}
A dominant line of work seeks to expand embodied capabilities by improving the model itself through pretraining, post-training, architectural specialization, or tighter perception-action integration. For spatial understanding, prior work has developed open-vocabulary scene representations, visual-language maps, 3D scene graphs, and retrieval-based world grounding to improve object semantics, geometric reasoning, and environment querying \cite{Che22,Hua22,Ran23,Wer24,Boo24}. For action and control, another major direction is to train more capable vision-language-action models and embodied policies that more tightly couple perception, reasoning, and action generation \cite{Ma24,Mon25}. Planning-oriented approaches further enhance the policy core by integrating language models with symbolic planners, task-and-motion planning, and behavior-tree generation \cite{Liu23b,Che23c,Ao24}. While these efforts substantially improve embodied competence, they still face persistent limitations in coverage, modular reuse, robustness, and the flexible composition of heterogeneous capabilities, especially in long-horizon, compositional, and safety-critical settings \cite{Wan24e,Lia25,Sal25}.

\paragraph{Tool usage.}
In parallel, the LLM agent literature has shown that outsourcing capabilities to external tools can significantly improve scalability, flexibility, and task performance, leading to a mature paradigm of tool selection, tool calling, and multi-step orchestration \cite{Qu24}. Embodied AI has already exhibited early signs of this direction: ROS-LLM exposes robot actions and services to language-model-based reasoning, ROSA wraps ROS functionality into callable tools, RoboNeuron uses MCP to expose robot capabilities as agent-callable tools, and ECP-style work argues for protocol-level coordination across embodied subsystems \cite{Mow24,Roy24,Gua25,Ma25}. Recent surveys also recognize modular tool use and protocol-oriented integration as emerging themes in embodied AI \cite{Lia25,Sal25}. However, despite these promising early efforts, there is still no standardized embodied tool protocol for unified capability access, no systematically collected and validated embodied tool suite spanning the full embodied decision stack, and no clear understanding of how effectively current embodied models can select, coordinate, and benefit from outsourced tools in embodied settings. These gaps motivate our \emph{EmbodiedTool} framework.

\section{EmbodiedTool}
\label{app:embodiedtool}
\subsection{Taxonomy of Embodied Capabilities}
\label{app:capability_taxonomy}

To justify the capability decomposition adopted in this work, we organize embodied intelligence around four core capability groups: \textbf{perception and grounding}, \textbf{cognition and state modeling}, \textbf{reasoning and planning}, and \textbf{execution and control}. This taxonomy is consistent with recent surveys that analyze embodied systems through the full perception-cognition-planning-action loop, arguing that embodied intelligence should be examined as an integrated process from environmental sensing to physical execution~\cite{hou2026survey}.

Our decomposition follows the same high-level logic while adapting it to the goal of tool construction. Specifically, we focus on capability groups that can be externalized as independently parameterized embodied tools and invoked on demand during decision-making. The resulting taxonomy provides a principled basis for deciding what kinds of tools should be built, how they should be organized, and what role they play in embodied agents.

\paragraph{Perception and grounding.}
This capability group concerns how an agent acquires environmental information and grounds task instructions in multimodal observations. It includes visual perception, object recognition, open-vocabulary detection, spatial localization, scene understanding, and language-vision grounding. The goal is not merely to sense raw inputs, but to convert observations and instructions into structured, task-relevant representations that can support subsequent reasoning, planning, and control.

\paragraph{Cognition and state modeling.}
This capability group concerns how an agent interprets perceived information and maintains an internal representation of the current situation. It includes state estimation, relational modeling, affordance understanding, task-constraint interpretation, risk identification, and commonsense-based situation assessment. In embodied settings, this capability enables the agent to determine what the current environment means, which entities and relations are relevant, and what latent state or dependency should be modeled for downstream decision-making.

\paragraph{Reasoning and planning.}
This capability group concerns how an agent derives decisions and organizes future behavior based on the modeled state and task goal. It includes goal decomposition, subgoal generation, causal reasoning, long-horizon planning, action sequencing, contingency handling, and replanning under environmental changes. While cognition and state modeling focus on understanding the current situation, reasoning and planning focus on deciding what should be done next and how individual decisions should be composed into a coherent executable plan.

\paragraph{Execution and control.}
This capability group concerns how an agent realizes planned decisions through physically grounded actions. It includes trajectory generation, grasp planning, manipulation control, navigation control, low-level actuation, and execution monitoring. This capability ensures that abstract decisions and planned action sequences can be translated into stable, feasible, and effective behavior in the physical world.

\subsection{Embodied Tool Collection}
\label{app:embodied_tool_collection}
\paragraph{Collection objective.}
Our embodied tool collection is designed to externalize the major capabilities required by embodied decision-making into callable and independently deployable tools. Starting from the capability taxonomy in Section~\ref{app:capability_taxonomy}, we organize the collection around four major dimensions: \textbf{perception and grounding}, \textbf{cognition and state modeling}, \textbf{reasoning and planning}, and \textbf{execution and control}. For tool construction, each capability dimension is further decomposed into finer-grained embodied problems, such as object detection, segmentation, spatial grounding, memory retrieval, affordance reasoning, task decomposition, action grounding, and closed-loop execution monitoring. This decomposition allows us to move from abstract capabilities to concrete toolable functions.

\paragraph{Problem decomposition and source collection.}
For each fine-grained problem, we use AI-assisted literature exploration tools, including deep researcher, to collect representative state-of-the-art methods from the corresponding domains, such as computer vision, robot learning, multimodal reasoning, and motion planning. The collected candidates include both widely adopted research models and practically useful system modules. At this stage, our goal is not simply to maximize the number of candidate methods, but to identify methods whose input-output behavior, execution assumptions, and deployment cost are compatible with embodied use.

\paragraph{Human review and tool selection.}
All collected candidates are then manually reviewed before being admitted into the tool suite. In particular, we prioritize methods that are 1) \emph{lightweight}, so that they are practical for embodied use, 2) \emph{easy to deploy}, so that they can be maintained within a unified serving framework, and 3) \emph{environmentally robust}, so that they can adapt across different environments and operating conditions. We also examine whether a method matches the target embodied subproblem, whether its interface can be exposed as a reusable tool, and whether it provides sufficiently stable outputs for downstream decision-making. This review step is important because not every strong method on its original benchmark is suitable as an embodied tool: some methods are too heavy for efficient deployment, some are difficult to integrate into real systems, and some do not generalize reliably beyond a narrow evaluation setting.

\paragraph{Deployment, validation, and API wrapping.}
For each selected method, we first implement or adapt a deployable version and validate it in our embodied setting. Importantly, we do not rely only on the method's original benchmark. Instead, beyond its native evaluation setup, we additionally verify each tool on two to three extra benchmarks or testing scenarios whenever possible, so as to ensure that the tool remains effective across different environments, task distributions, and input conditions. After validation, each tool is deployed on our servers and encapsulated as an accessible API. We further attach a structured tool card to each deployed tool, specifying its functionality, input-output schema, applicability conditions, and usage constraints, so that the tool can be registered under ETP and invoked consistently by embodied agents.

\begin{figure*}[t]
\centering
\includegraphics[width=0.99\textwidth]{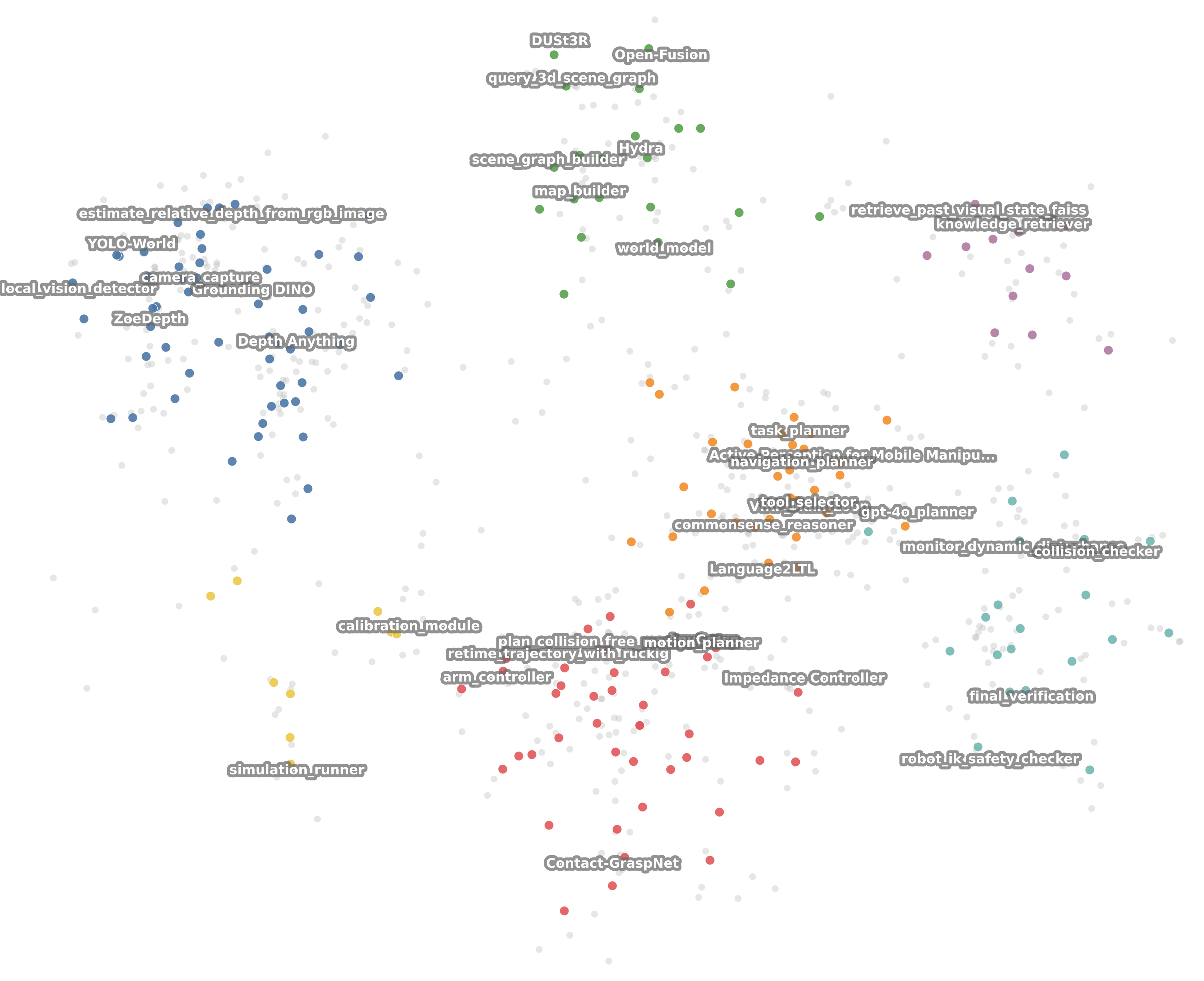}
\caption{Embodied tool embedding visualization.}
\label{fig:tools_embedding}
\end{figure*}
\paragraph{Representative tools.}
Figure~\ref{fig:tools_embedding} visualizes the word embeddings of over 100 tools, showing how they span diverse embodied capabilities in the tool space. Table~\ref{tab:tool_catalogue} lists a representative subset of tools from the broader collection. We select examples that are already deployed or operationally validated, and that together cover memory retrieval, intent grounding, closed-loop monitoring, grasp generation, motion planning, inverse kinematics, and contact-aware control.

% ============================================================
%  Top-conference style tool catalogue table (NeurIPS / CVPR)
%  22 representative tools across 6 capability categories
%  Requirements: booktabs, multirow, colortbl, xcolor, array,
%                makecell, rotating (for sideways header if needed)
% ============================================================

% ---- Preamble additions (place once in main.tex) -----------
% \usepackage{booktabs, multirow, colortbl, xcolor, array, makecell}
% \definecolor{groupbg}{RGB}{242,242,242}
% \newcolumntype{L}[1]{>{\raggedright\arraybackslash}p{#1}}
% \newcolumntype{C}[1]{>{\centering\arraybackslash}p{#1}}
% ------------------------------------------------------------

\begingroup
\setlength{\tabcolsep}{1.8pt}
\setlength{\LTleft}{0pt}
\setlength{\LTright}{0pt}
\renewcommand{\arraystretch}{1.04}
\scriptsize

\begin{longtable}{@{}
  C{0.45cm}
  L{1.55cm}
  L{1.15cm}
  L{2.30cm}
  L{1.45cm}
  L{1.45cm}
  L{1.55cm}
  L{1.00cm}
  L{1.95cm}
@{}}
\caption{%
  Representative tool catalogue of the embodied benchmark suite.
  Tools are grouped by capability unit.
  \textbf{Mode}: \textit{On-demand} = invoked once per query;
  \textit{Continuous} = runs in background loop;
  \textit{Event} = triggered by runtime condition.
  \textbf{Source} lists the originating paper or framework.
}\label{tab:tool_catalogue}\\
\specialrule{1.2pt}{0pt}{2pt}
\textbf{ID}
  & \textbf{Tool Name}
  & \textbf{Capability Unit}
  & \textbf{Description}
  & \textbf{Input}
  & \textbf{Output}
  & \textbf{Trigger Condition}
  & \textbf{Mode}
  & \textbf{Source} \\
\specialrule{0.8pt}{1pt}{2pt}
\endfirsthead

\multicolumn{9}{@{}l@{}}{\tablename\ \thetable\ -- continued from previous page}\\
\specialrule{1.2pt}{0pt}{2pt}
\textbf{ID}
  & \textbf{Tool Name}
  & \textbf{Capability Unit}
  & \textbf{Description}
  & \textbf{Input}
  & \textbf{Output}
  & \textbf{Trigger Condition}
  & \textbf{Mode}
  & \textbf{Source} \\
\specialrule{0.8pt}{1pt}{2pt}
\endhead

\specialrule{0.8pt}{2pt}{2pt}
\multicolumn{9}{r@{}}{Continued on next page}\\
\endfoot

\specialrule{1.2pt}{1pt}{0pt}
\endlastfoot

% ============================================================
%  GROUP 1 — Queryable Memory
% ============================================================
\multicolumn{9}{@{}l@{}}{%
  \cellcolor{groupbg}\hspace{4pt}%
  \textit{Queryable Memory}\hspace{4pt}} \\[1pt]

T01
  & \nolinkurl{query_3d_scene_graph}
  & Queryable Memory
  & Queries historical position and spatial relations of objects observed in a 3-D scene graph.
  & Scene graph; object/relation query
  & Object location; neighbor relations; confidence
  & Target lost from current view
  & On-demand
  & \cite{wu2021scenegraphfusion} \\[2pt]

T02
  & \nolinkurl{STM}
  & Queryable Memory
  & Maintains a fixed-size space-time memory bank; reads and writes target features frame by frame to combat long-term drift.
  & Current frame features; memory bank query
  & Memory readout; attention map; confidence score
  & Target blurred; VLM needs early-frame feature anchor
  & Continuous
  & \cite{oh2019stm} \\[2pt]

T03
  & \nolinkurl{Action Genome}
  & Queryable Memory
  & Infers the current physical state of objects from their motion trajectory via a spatio-temporal scene graph network.
  & Short video clip; target object list
  & State graph; temporal relations
  & VLM needs exact object state for next-step decision
  & On-demand
  & \cite{ji2020actiongenome} \\

\midrule

% ============================================================
%  GROUP 2 — Intent Grounding
% ============================================================
\multicolumn{9}{@{}l@{}}{%
  \cellcolor{groupbg}\hspace{4pt}%
  \textit{Intent Grounding}\hspace{4pt}} \\[1pt]

T04
  & \nolinkurl{Language2LTL}
  & Intent Grounding
  & Translates natural-language SOP into a formal LTL formula; rejects step-skipping or unsafe hallucinated plans before execution.
  & VLM action plan; safety LTL formulas
  & Validation result; violation feedback
  & VLM completes abstract planning; output about to be queued
  & On-demand
  & \cite{shah2023language2ltl} \\[2pt]

T05
  & \nolinkurl{LINGO-Space}
  & Intent Grounding
  & Incrementally estimates a probability distribution over goal positions given composite spatial-referring instructions and scene geometry.
  & Composite instruction; scene objects
  & Target location distribution; confidence
  & Instruction contains spatial referring expressions
  & On-demand
  & \cite{agrawal2023lingospace} \\

\midrule

% ============================================================
%  GROUP 3 — Detection \& Segmentation
% ============================================================
\multicolumn{9}{@{}l@{}}{%
  \cellcolor{groupbg}\hspace{4pt}%
  \textit{Detection \& Segmentation}\hspace{4pt}} \\[1pt]

T06
  & \nolinkurl{YOLO-World}
  & Det.\ \& Seg.
  & Real-time open-vocabulary object detector; finds arbitrary text-described objects in RGB frames.
  & RGB image; text query
  & Bounding boxes; object categories
  & Open-vocabulary detection required
  & On-demand
  & \cite{cheng2024yoloworld} \\[2pt]

T07
  & \nolinkurl{FastSAM}
  & Det.\ \& Seg.
  & CNN-based real-time solution for the Segment Anything task; supports point, box, and text prompts.
  & RGB image; point / box / text prompts
  & Segmentation masks
  & Fast instance segmentation required
  & On-demand
  & \cite{zhao2023fastsam} \\[2pt]

T08
  & \nolinkurl{GigaPose}
  & Det.\ \& Seg.
  & Estimates the 6-DoF pose of known CAD-model objects via one-correspondence matching.
  & RGB image; CAD model templates
  & 6-D pose (R, t)
  & 6-DoF pose of known object needed
  & On-demand
  & \cite{nguyen2024gigapose} \\[2pt]

T09
  & \nolinkurl{CenterPoint}
  & Det.\ \& Seg.
  & Center-based 3-D object detection and multi-object tracking on LiDAR point clouds.
  & 3-D point cloud
  & 3-D bounding boxes; tracking IDs; velocities
  & 3-D detection and tracking needed
  & On-demand
  & \cite{yin2021centerpoint} \\[2pt]

T10
  & \nolinkurl{Cutie}
  & Det.\ \& Seg.
  & Video object segmentation with long-term memory; tracks and segments target from first-frame mask.
  & Video frames; initial mask
  & Per-frame segmentation masks
  & Continuous video tracking needed
  & Continuous
  & \cite{cheng2023cutie} \\

\midrule

% ============================================================
%  GROUP 4 — Localization \& World Modeling
% ============================================================
\multicolumn{9}{@{}l@{}}{%
  \cellcolor{groupbg}\hspace{4pt}%
  \textit{Localization \& World Modeling}\hspace{4pt}} \\[1pt]

T11
  & \nolinkurl{Open-Fusion}
  & Loc.\ \& World
  & Builds a real-time open-vocabulary queryable 3-D semantic field from RGB-D sequences.
  & RGB-D sequence; camera poses; query text
  & Open-vocabulary 3-D semantic map
  & World model refresh or semantic query needed
  & Map update + query
  & \cite{yamazaki2024openfusion} \\[2pt]

T12
  & \nolinkurl{Hydra}
  & Loc.\ \& World
  & Organises a 3-D metric map into a room-region-object hierarchy for high-level spatial reasoning.
  & 3-D map; semantic observations; robot poses
  & Hierarchical 3-D scene graph
  & High-level planning or semantic query
  & Background + query
  & \cite{hughes2022hydra} \\[2pt]

T13
  & \nolinkurl{DUSt3R}
  & Loc.\ \& World
  & Reconstructs dense 3-D point maps and camera poses from an unconstrained image collection without prior calibration.
  & Unconstrained RGB images
  & 3-D point maps; camera poses
  & 3-D scene reconstruction from unposed images needed
  & On-demand
  & \cite{wang2024dust3r} \\[2pt]

T14
  & \nolinkurl{ZoeDepth}
  & Loc.\ \& World
  & Predicts metric-scale depth from a single RGB image using relative-to-metric transfer; provides geometric prior for grasping and navigation.
  & RGB image / batched tensor
  & Metric depth map; 16-bit depth image; colorized visualization
  & Depth sensor absent; downstream module needs geometry
  & On-demand
  & \cite{bhat2023zoedepth} \\[2pt]

T15
  & \nolinkurl{build_query_3d_occupancy_map}
  & Loc.\ \& World
  & Fuses depth / point-cloud streams into an OctoMap octree encoding occupied, free, and unknown voxels for navigation and arm workspace planning.
  & Depth stream; sensor pose; map resolution
  & Octree map; occupied / free / unknown voxel queries
  & 3-D obstacle or free-space query needed
  & Continuous update
  & \cite{hornung2013octomap} \\

\midrule

% ============================================================
%  GROUP 5 — Closed-loop Execution, Recovery \& Safety
% ============================================================
\multicolumn{9}{@{}l@{}}{%
  \cellcolor{groupbg}\hspace{4pt}%
  \textit{Closed-loop Execution, Recovery \& Safety}\hspace{4pt}} \\[1pt]

T16
  & \nolinkurl{TAPIR}
  & Closed-loop Exec.
  & Lightweight point-tracker that outputs up-to-date 2-D coordinates of a displaced target, driving visual servo without heavy pose estimation.
  & Query point; live video stream
  & Tracked point; occlusion flag
  & Arm approaching target; target may have moved
  & Continuous stream
  & \cite{doersch2023tapir} \\[2pt]

T17
  & \nolinkurl{R3M}
  & Closed-loop Exec.
  & Computes cross-modal distance between post-action frame features and language instruction; outputs a Boolean completion assertion.
  & Post-action frame; goal instruction text
  & Completion score; completion flag
  & Discrete action (grasp, place) just executed
  & On-demand
  & \cite{nair2022r3m} \\[2pt]

T18
  & \nolinkurl{VIRF_Main_Loop}
  & Closed-loop Exec.
  & Orchestrates a plan-verify-diagnose-correct closed loop; invokes LLM planner and safety verifier iteratively until a safe plan is found or task is rejected.
  & User instruction; world knowledge core
  & Safe plan; task rejection signal
  & Embodied task instruction received
  & Continuous loop
  & \cite{virf2026} \\

\midrule

% ============================================================
%  GROUP 6 — Task-to-Action Bridging
% ============================================================
\multicolumn{9}{@{}l@{}}{%
  \cellcolor{groupbg}\hspace{4pt}%
  \textit{Task-to-Action Bridging}\hspace{4pt}} \\[1pt]

T19
  & \nolinkurl{AnyGrasp}
  & Task-to-Action
  & Translates coarse 2-D semantic intent into a precise 6-DoF gripper pose from point cloud input; bridges semantic and kinematic gap.
  & Goal pose; joint state; scene point cloud
  & 6-DoF pose; gripper width; confidence
  & VLM locks target; physical grasp action to be generated
  & On-demand
  & \cite{fang2023anygrasp} \\[2pt]

T20
  & \nolinkurl{Contact-GraspNet}
  & Task-to-Action
  & Regresses 6-DoF gripper pose and width directly from a cluttered point cloud; refines semantic grasp to physical contact mechanics.
  & Cluttered scene depth / point cloud; optional segmap
  & 6-DoF grasp candidate set; contact point distribution
  & Dense-packing grasp planning triggered
  & On-demand
  & \cite{sundermeyer2021contactgraspnet} \\[2pt]

T21
  & \nolinkurl{navigate_to_goal_pose}
  & Task-to-Action
  & Translates a high-level goal pose into a navigation action stream via Nav2 stack; handles path planning and obstacle avoidance.
  & Goal pose; map; obstacle state
  & Path; navigation action stream
  & High-level navigation goal pose provided
  & On-demand + exec.
  & \cite{macenski2020nav2} \\[2pt]

T22
  & \nolinkurl{plan_collision_free_manipulation}
  & Task-to-Action
  & Generates a collision-free joint trajectory for manipulation via MoveIt\,2; resolves goal pose into executable waypoints.
  & Goal pose; robot state; collision scene
  & Collision-free joint trajectory
  & Manipulation goal pose and grasp intent available
  & On-demand
  & \cite{coleman2014moveit} \\

\end{longtable}
\endgroup

\section{EmbodiedToolBench Design}
\label{app:embodiedtoolbench}

\subsection{Task 1: Tool-Need Recognition}
\label{app:tool_need_recognition}
\paragraph{Data Collection and Annotation.}
The dataset for Tool-Need Recognition consists of two categories of samples: positive instances (requiring tool invocation) and negative instances (solvable without tools). Positive instances are directly sourced from complete embodied interaction trajectories: decision states $(l, \tau_t, o_t, L)$ in which a tool invocation occurs are labeled $u^\star = 1$. Negative instances are collected from two types of scenarios:
\begin{enumerate}
    \item \textbf{Directly solvable states}, where the model can produce a valid action based solely on the current observation $o_t$ and interaction history $\tau_t$, without resorting to any external tool;
    \item \textbf{Tool-redundant states}, where the candidate tool set $L$ is non-empty, yet no tool invocation is required at the current stage of the task.
\end{enumerate}

\paragraph{Class Balance.}
To prevent systematic prediction bias, we maintain a positive-to-negative sample ratio of $1{:}1$ and apply stratified sampling across task types (navigation, planning, and manipulation), ensuring that each scenario is uniformly represented in the dataset.

\paragraph{Distractor Difficulty Control.}
To increase the difficulty of negative instances, we deliberately introduce \emph{tool-inducing} negative samples: these samples contain task descriptions with keywords strongly associated with tool functionality (\emph{e.g.}, ``detect'', ``grasp'', ``navigate to''), yet based on the current observation and interaction history, no tool invocation is actually required. This design prevents models from relying solely on superficial textual cues to make decisions.

\subsection{Task 2: Tool Selection}
\label{app:tool_selection}
\paragraph{Candidate Tool Set Construction.}
For each query $q$, the corresponding candidate tool set $L_t$ consists of four tools: one ground-truth tool $z^\star$ and three distractors. Distractor tools are selected according to the following hierarchical strategy:

\begin{itemize}
    \item \textbf{Semantically Similar Distractors (Hard Negatives).}
    We employ the \texttt{text-embedding-ada-002} model to embed the descriptions of all tools in the tool library $\mathcal{Z}$, compute the cosine similarity between each tool and $z^\star$, and select the top-$(n-1)$ most similar tools as the candidate distractor pool. These tools are functionally close to $z^\star$ in their descriptions, effectively preventing models from completing the selection task via simple keyword matching.

    \item \textbf{Uniqueness of the Ground-Truth Answer.}
    Filtering distractors by semantic similarity alone is insufficient, as some similar tools may functionally overlap with $z^\star$, potentially allowing multiple tools to accomplish the current task and thereby undermining annotation uniqueness. To address this, we perform a \textbf{precondition check} for each candidate distractor: if a tool $z'$ has its preconditions satisfiable under the current state $(l, \tau_t, o_t)$ and its effects can fulfill the task objective, it is excluded from the distractor pool. Only tools that \textbf{cannot be successfully executed in the current state} due to precondition constraints are retained as valid distractors.

    \item \textbf{Intra-category and Inter-category Distractor Mix.}
    Among the three final distractors, at least two are intra-category distractors drawn from the same tool category as $z^\star$ (\emph{e.g.}, all from the perception, manipulation, or navigation category), and at most one is an inter-category distractor from a different category. This design ensures that the evaluation tests both the model's ability to distinguish fine-grained functional differences and its capacity to filter out irrelevant tools.
\end{itemize}

\paragraph{Example.}
Consider the task ``detect the target object on the table'', where the ground-truth tool is \texttt{ObjectDetector} (supporting object detection from the current viewpoint). Distractors may include: \texttt{SceneSegmentor} (semantic segmentation; functionally similar but with output incompatible with subsequent actions), \texttt{DepthEstimator} (depth estimation; in the same perception category but does not satisfy the classification requirement of the task), and \texttt{GraspPlanner} (from the manipulation category; its precondition requires a known target location, which is not satisfied in the current state).

\subsection{Task 3: Tool Execution}
\label{app:tool_execution}
\paragraph{Stage 1 (Query Construction) Data Design.}
Each Stage~1 sample contains the current task context $(l, \tau_t, o_t)$ and the complete specification of the designated tool $z^\star$ (API name, parameter list, parameter types, and value ranges). The reference query $x^\star$ is constructed by human annotators based on the actual environment state and is automatically verified by a format validator ($\mathrm{Valid}(x^\star, z^\star) = 1$). To increase difficulty, we specifically select tool invocation scenarios where \textbf{parameter values depend on the current observation}, such as coordinates or object IDs that must be read from $o_t$, rather than simple invocations that can rely on default values.

\paragraph{Stage 2 (Output Comprehension) Data Design.}
Stage~2 samples extend Stage~1 by appending the real tool output $y$ (obtained from actual tool execution), requiring the model to predict the next action $\hat{a}$. The reference action $a^\star$ is likewise sourced from expert trajectories and validated through semantic equivalence matching ($\mathrm{Match}$) to accommodate variations in action descriptions. We ensure that the output format of $y$ is diverse, including structured JSON, natural language descriptions, and numerical results, so as to comprehensively evaluate the model's ability to interpret and utilize heterogeneous tool outputs.

\paragraph{Joint Evaluation Logic for TUSR.}
Since TUSR requires both Stage~1 and Stage~2 to succeed, the dataset construction deliberately ensures that the error patterns of the two stages are independent of each other: Stage~1 errors arise primarily from failures in parameter construction, while Stage~2 errors stem mainly from misinterpretation of tool outputs. There is no systematic co-occurrence bias between the two types of failures.

\subsection{Task 4: Tool-Chain Composition}
\label{app:tool_chain_composition}

\paragraph{Source of Tool Chains.}
The reference tool sequence $P^\star = (z_1^\star, \ldots, z_K^\star)$ in the Tool-Chain Composition task is derived from consecutive tool invocation segments in complete expert trajectories, manually annotated as the \textbf{minimal tool set} required to accomplish a specific sub-goal. Trajectory segments containing redundant tool invocations are excluded to ensure the minimality of $S^\star$.

\paragraph{Dependency Constraint Annotation.}
The execution order constraint set $\mathcal{C}^\star = \{(z_a, z_b)\}$ is constructed from two types of dependency relations:
\begin{enumerate}
    \item \textbf{Data dependencies}: the input parameters of tool $z_b$ are derived from the output of tool $z_a$, forming a direct dataflow dependency;
    \item \textbf{State dependencies}: the preconditions of $z_b$ require the environmental state produced by the execution of $z_a$ (\emph{e.g.}, \texttt{GraspPlanner} can only plan a grasping path after \texttt{ObjectDetector} has successfully localized the target).
\end{enumerate}
All constraints are formally verified to ensure their logical necessity and completeness.

\paragraph{Candidate Tool Set Construction (Distractor Strategy).}
The candidate tool set $L$ is constructed by augmenting $S^\star$ with two additional categories of distractor tools:
\begin{itemize}
    \item \textbf{Functionally Overlapping Distractors}: tools that are functionally similar to some tool in $S^\star$ but would cause dataflow disruption or state inconsistency if inserted into the current tool chain. The model must identify the inapplicability of such tools in the specific chain context.
    \item \textbf{Redundancy-Inducing Distractors}: tools that are functionally correct but redundant given the minimal tool set constraint (\emph{e.g.}, introducing an additional perception tool when the target location is already known). These distractors are designed to test the model's understanding of the \emph{minimality} principle.
\end{itemize}

\paragraph{Robustness Design for OCR.}
To prevent models from achieving inflated OCR scores by memorizing common tool ordering patterns, we deliberately include samples with \textbf{non-canonical dependency orders}, \emph{i.e.}, tool invocation sequences that are counter-intuitive yet logically valid (\emph{e.g.}, performing local perception before global planning). Such samples account for approximately $20\%$ of the total Task~4 instances, effectively preventing models from completing the ordering task through prior biases rather than genuine reasoning.

\section{Statistical Distribution}
\label{app:statical}
\subsection{EmbodiedTools}
\label{app:statical_embodiedtools}
\begin{figure*}[htb]
\centering
\includegraphics[width=0.99\textwidth]{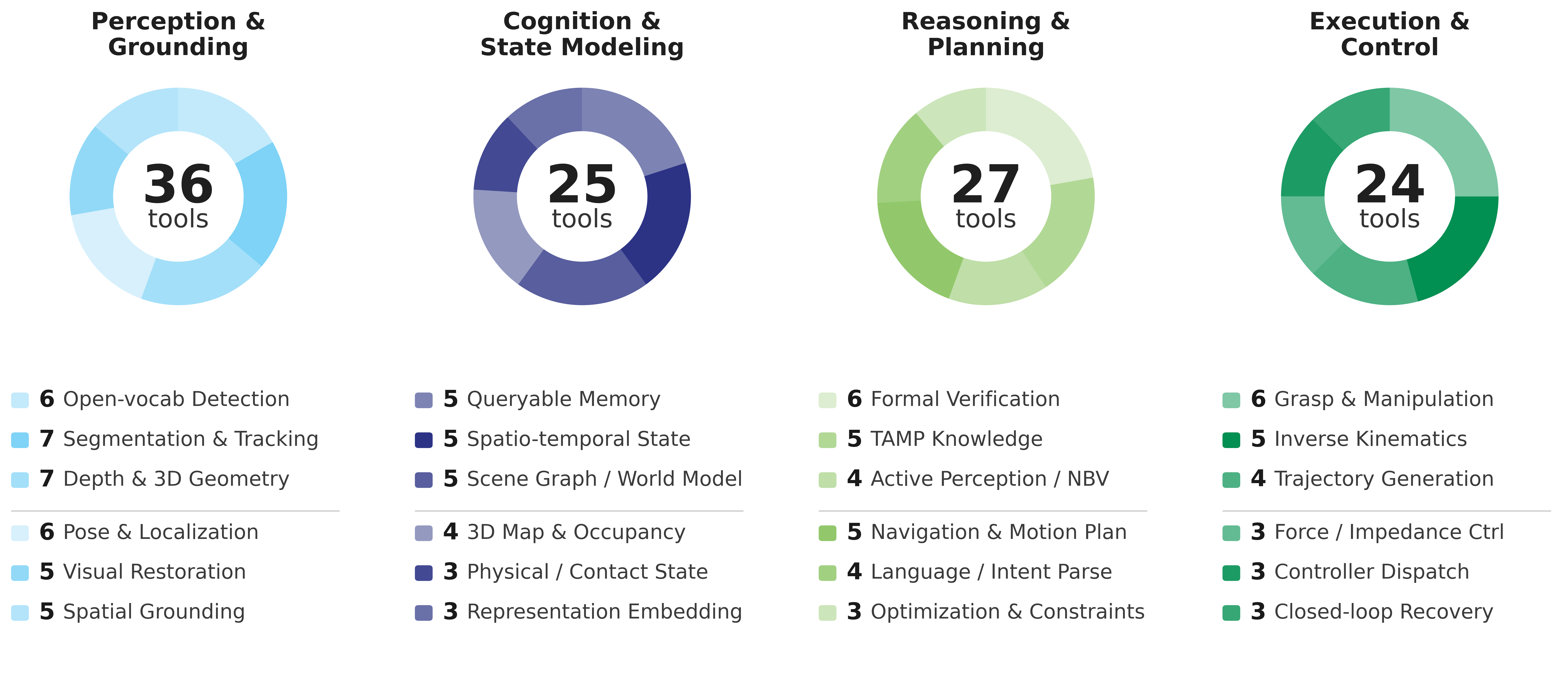}
\caption{Distribution of tools across macro capability groups and fine-grained subcategories.}
\label{fig:tools_macro_subcategory_distribution}
\end{figure*}

Figure~\ref{fig:tools_macro_subcategory_distribution} summarizes the composition of our collected tool pool. In total, our collection comprises 112 tools distributed across four macro capability groups, spanning the full pipeline of an embodied agent: perception and grounding (36 tools), cognition and state modeling (25), reasoning and planning (27), and execution and control (24). Crucially, no single stage dominates the collection. The four groups are broadly balanced, ensuring that our benchmark does not inadvertently favor agents with narrow, stage-specific strengths.

Beyond this macro-level balance, coverage within each group is equally diverse. Rather than concentrating tools around one or two dominant subcategories, each macro group distributes its tools across six coherent subcategories. For instance, perception and grounding spans low-level geometric understanding, such as depth and 3D geometry, pose and localization, as well as high-level semantic grounding, such as open-vocabulary detection, segmentation and tracking, and spatial grounding. Cognition and state modeling captures the internal state-building capabilities required by embodied agents, including queryable memory, spatio-temporal state modeling, scene graph and world modeling, 3D map and occupancy modeling, physical/contact state estimation, and representation embedding. Reasoning and planning bridges formal symbolic methods, such as formal verification and TAMP knowledge, with planning-oriented capabilities, including active perception, navigation and motion planning, language and intent parsing, and optimization under constraints. Execution and control covers both the kinematic layer, such as inverse kinematics and trajectory generation, and the adaptive control layer, such as force/impedance control, controller dispatch, and closed-loop recovery.

\subsection{EmbodiedToolBench}
\label{app:statical_embodiedtoolbench}
Table~\ref{tab:dataset_statistics} summarizes the statistical coverage of \textsc{EmbodiedToolBench}. The benchmark comprises four complementary tool-use tasks, each consisting of 100 samples spanning four distinct embodied environments, yielding a total of 400 samples with consistent cross-environment coverage. The four tasks are carefully structured to reflect a progressive hierarchy of cognitive demand, from binary tool-presence judgments in \textit{Tool-Need Recognition} to ordered tool-chain composition in \textit{Tool-Chain Composition}, with each task containing samples of varying difficulty levels to ensure fine-grained evaluation within each capability tier.
Each subset is also designed with broad and diverse coverage across multiple dimensions. \textit{Tool-Need Recognition} and \textit{Tool Selection} each cover over 60 candidate tools, while \textit{Tool-Chain Composition} encompasses as many as 112 candidates alongside 51 gold tools, reflecting the rich combinatorial space required for multi-step reasoning. The average history context ranges from 1.06 to 2.26 turns with 1.44 to 2.02 accompanying images per sample, providing graded multimodal contextual signals across tasks.
These statistics suggest that \textsc{EmbodiedToolBench} provides a comprehensive and well-structured testbed that spans a wide spectrum of embodied tool-use scenarios, varying systematically in task complexity, tool space coverage, and contextual richness.

\newcommand{\taskbadge}[2]{%
  \begingroup
  \setlength{\fboxsep}{2pt}%
  \colorbox{#1!10}{\textcolor{#1}{\scriptsize\textbf{#2}}}%
  \endgroup
}

\begin{table*}[htb]
\centering
\caption{
Overview of EmbodiedToolBench. Each subset contains 100 samples across four embodied environments.
Tasks progress from binary tool-awareness decisions to ordered tool-chain composition.
\textbf{Diff.} denotes task difficulty, and ``--'' indicates no fixed candidate set for Tool Usage.
}
\label{tab:dataset_statistics}
\scriptsize
\setlength{\tabcolsep}{4.8pt}
\renewcommand{\arraystretch}{1.28}

\resizebox{\textwidth}{!}{%
\begin{tabular}{@{}lcccccccc@{}}
\toprule
\textbf{Task}
& \textbf{Samples}
& \textbf{Envs.}
& \textbf{Diff.}
& \makecell{\textbf{Task}\\\textbf{Types}}
& \makecell{\textbf{Gold}\\\textbf{Tools}}
& \makecell{\textbf{Candidate}\\\textbf{Tools}}
& \makecell{\textbf{Avg. History}\\\textbf{Turns}}
& \makecell{\textbf{Avg. History}\\\textbf{Images}} \\
\midrule
\taskbadge{awblue}{Tool-Need Recognition}
& 100 & 4 & 2 & 1  & 36 & 62 & 2.02 & 2.02 \\
\taskbadge{segreen}{Tool Selection}
& 100 & 4 & 3 & 1  & 35  & 66 & 2.26 & 1.62 \\
\taskbadge{usorange}{Tool Execution}
& 100 & 4 & 3 & 2  & 25  & -- & 1.65 & 1.60 \\
\taskbadge{copurple}{Tool-Chain Composition}
& 100 & 4 & 3 & 2 & 51  & 112 & 1.06 & 1.44 \\
\bottomrule
\end{tabular}%
}
\end{table*}

\begin{figure*}[htb]
\centering
\includegraphics[width=0.99\textwidth]{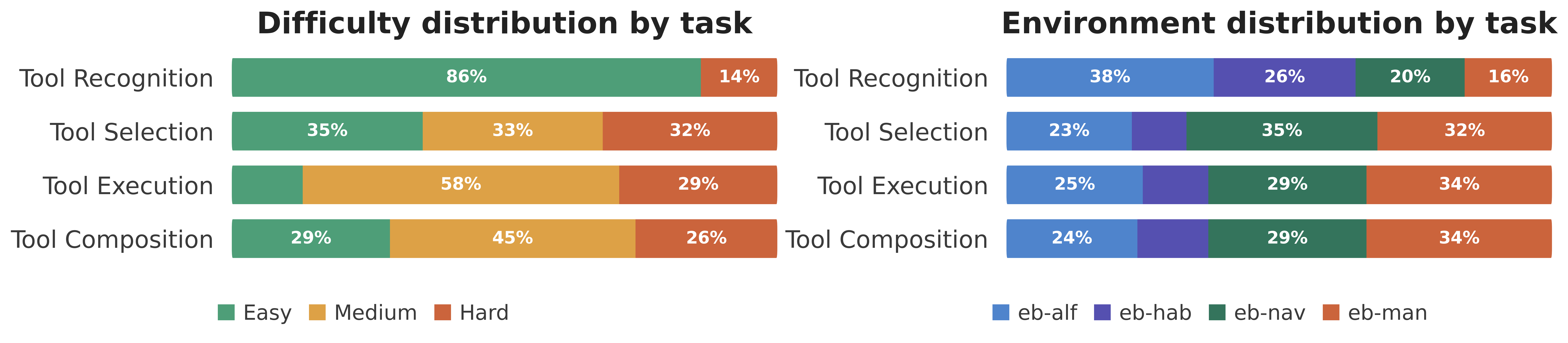}
\caption{Distribution of EmbodiedToolBench samples across tasks, environments, and tool-use dimensions.}
\label{fig:embodiedtool_distribution}
\end{figure*}

Figure~\ref{fig:embodiedtool_distribution} further illustrates the dataset distribution across difficulty levels and embodied environments. The difficulty profiles vary meaningfully across tasks, reflecting the distinct cognitive demands of each. \textit{Tool-Need Recognition} is heavily skewed toward easy instances (86\%), consistent with its binary decision nature. In contrast, \textit{Tool Selection} achieves a well-balanced distribution across easy, medium, and hard cases (35\%/33\%/32\%), ensuring comprehensive coverage of selection complexity. \textit{Tool Execution} and \textit{Tool-Chain Composition} are predominantly composed of medium and hard instances (87\% and 71\%, respectively), reflecting the greater contextual and compositional demands these tasks impose. Across all four tasks, the difficulty design ensures that no single difficulty level dominates the benchmark as a whole, providing a balanced and challenging evaluation suite.
The environment distribution is similarly well-balanced. \textit{Tool-Need Recognition} draws from all four environments, with EB-ALFRED contributing the largest share (38\%). The remaining three tasks exhibit a more even spread across EB-ALFRED, EB-Habitat, EB-Navigation, and EB-Manipulation, with no single environment accounting for more than 35\% in any task. Together, these distributions confirm that \textsc{EmbodiedToolBench} is designed with both deliberate difficulty stratification and broad environmental diversity, enabling a rigorous and representative evaluation of embodied tool-use capabilities.

% Figure~\ref{fig:embodiedtool_distribution_2} shows how often individual tools appear as gold answers across the four tasks. The distribution is clearly long-tailed: a small number of tools account for a large fraction of the benchmark, while many others appear less frequently. Grounding DINO is the most frequent tool overall, with 138 occurrences, and is especially dominant in tool composition (94). Contact-GraspNet and \texttt{navigate\_to\_goal\_pose} are also heavily represented, with 85 and 83 total occurrences, respectively, again driven largely by tool composition. By contrast, tools such as Grounded-SAM / SAM 2 and AnyGrasp appear in more specialized settings and are concentrated in a smaller number of tasks. At the same time, the benchmark is not dominated by a single task-specific pattern: some tools, such as HOV-SG (nav), FoundationPose, and \texttt{query\_3d\_scene\_graph}, are distributed across awareness, selection, and usage, indicating that the dataset covers both reusable cross-task tools and highly specialized tools. This distribution further supports the diversity of EmbodiedToolBench at the level of concrete tool instances.

% \begin{figure*}[htb]
% \centering
% \includegraphics[width=0.99\textwidth]{figures/embodiedtool_table_ref_1to1_nobg.png}
% \caption{Additional distribution statistics of EmbodiedToolBench across task-specific dimensions.}
% \label{fig:embodiedtool_distribution_2}
% \end{figure*}

\section{Additional Experiments}
\label{app:addiational_experiments}
\subsection{Implementation and Evaluation Details}
\label{app:implementation}
\paragraph{Compute Resources.} The compute resources used in this work consist of API-based evaluation resources and local GPU resources. Closed-source models are evaluated through their official APIs, with an approximate total cost of \$3000 for model invocation and benchmark evaluation. Specifically, we use a server with 8 NVIDIA A800 GPUs for tool inference and open-source component deployment.

\paragraph{Model Details.} Due to space limitations, we report abbreviated model names in the main paper.
Table~\ref{tab:model_details} provides the full model identifiers, the
corresponding abbreviations used in our experiments, model providers, and
parameter scales. For closed-source models, the exact number of parameters is
not publicly disclosed by the corresponding providers.

\begin{table}[t]
\centering
\caption{%
  Details of the models used in our experiments.
  \textbf{Abbr.}\ denotes the abbreviated model name used throughout the paper.
  $^\dagger$Exact parameter counts are not publicly disclosed by the provider.
  $^\ddagger$Mixture-of-Experts (MoE) architecture; only 3\,B parameters are
  activated per forward pass.
}
\label{tab:model_details}
\small
\setlength{\tabcolsep}{8pt}
\renewcommand{\arraystretch}{1.15}
\begin{tabular}{@{}llll@{}}
\toprule
\textbf{Model Identifier} &
\textbf{Abbr.} &
\textbf{Provider} &
\textbf{Parameter Scale} \\
\midrule
\multicolumn{4}{@{}l}{\textit{Closed-Source Models}} \\[2pt]
\texttt{gpt-4o}                        & GPT-4o       & OpenAI    & Not disclosed$^\dagger$ \\
\texttt{gpt-5-chat}                    & GPT-5        & OpenAI    & Not disclosed$^\dagger$ \\[4pt]
\texttt{claude-3-7-sonnet-20250219}    & Claude-3.7   & Anthropic & Not disclosed$^\dagger$ \\
\texttt{claude-sonnet-4-6}             & Claude-4.6   & Anthropic & Not disclosed$^\dagger$ \\
\midrule
\multicolumn{4}{@{}l}{\textit{Open-Source Models}} \\[2pt]
\texttt{Qwen3-VL-32B-Instruct}         & Qwen3-32B    & Alibaba   & ${\sim}$32\,B (dense) \\
\texttt{Qwen3-VL-8B-Instruct}          & Qwen3-8B     & Alibaba   & ${\sim}$8\,B (dense)  \\
\texttt{Qwen2.5-VL-32B-Instruct}       & Qwen2.5-32B  & Alibaba   & 32\,B (dense)         \\
\texttt{Qwen3.5-35B-A3B-Instruct}      & Qwen3.5-35B  & Alibaba   & 35\,B total, 3\,B active$^\ddagger$ \\
\bottomrule
\end{tabular}
\end{table}

\subsection{Real-world Robotic Setup}
\label{app:real_world_robot}
\begin{wrapfigure}{r}{0.55\textwidth}
    \centering
    \vspace{-10pt}
    \includegraphics[width=0.55\textwidth]{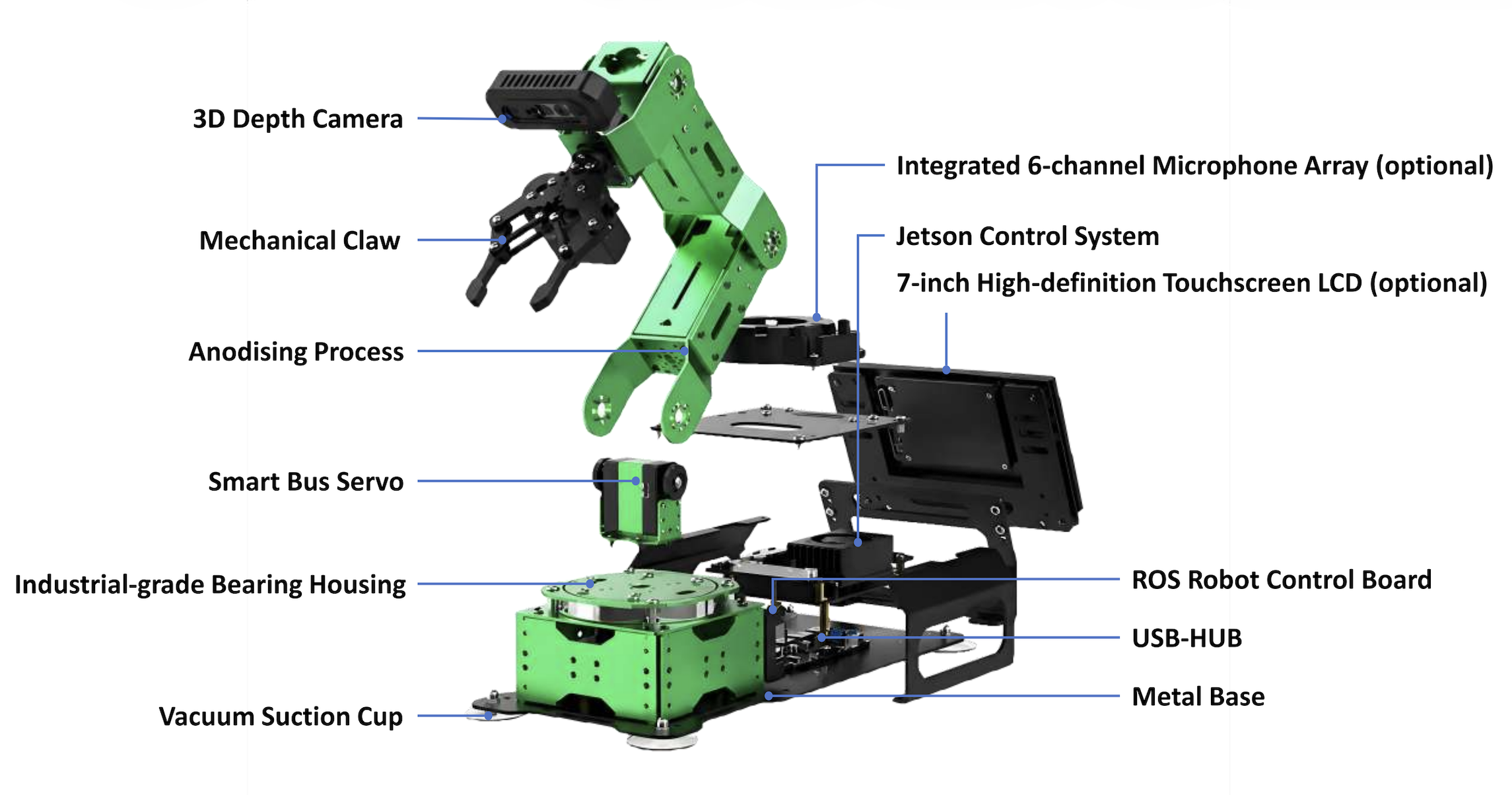}
    \caption{Overview of the high-performance robotic hardware platform.
    }
    \label{fig:hardware_overview}
    \vspace{-10pt}
\end{wrapfigure}
We conduct real-world experiments on a tabletop robotic manipulation platform equipped with a 6-DoF bus-servo robotic arm, a parallel robotic gripper, and a Gemini Plus RGB-D camera for visual and depth perception. The camera provides synchronized RGB and depth observations within a working range of 0.25--2.5 m, enabling object localization and scene understanding in tabletop environments. The arm is driven by high-torque smart bus servos and controlled through an STM32-based robotic controller, which handles low-level actuation and servo communication. High-level computation and policy execution are supported by an onboard Jetson control system, while a ROS-compatible control board and USB hub provide communication interfaces for perception, control, and peripheral devices. The whole system is mounted on a rigid metal base with vacuum suction cups to improve stability during manipulation.

\subsection{Statistical Significance Analysis}
\label{app:statistical_significance}
% Required packages:
% \usepackage{booktabs}
% \usepackage{colortbl}
% \usepackage{xcolor}
% \usepackage{graphicx}
% \usepackage{array}

\definecolor{groupbg}{RGB}{244,244,244}
\definecolor{toolbg}{RGB}{247,250,255}
\definecolor{avgbg}{RGB}{218,248,211}

\newcommand{\best}[1]{\textbf{#1}}
\newcommand{\worst}[1]{\underline{#1}}
\newcommand{\gainup}[1]{\makebox[4.3em][c]{$\uparrow\,#1$}}

\begin{table*}[t]
\centering
\caption{%
Performance comparison across four embodied benchmarks over three independent runs.
Results are reported as mean $\pm$ standard deviation.
\textbf{Bold}: best result per column;
\underline{underlined}: worst result;
\colorbox{avgbg}{Avg. Gain}: mean improvement of ``w/ tool'' over ``w/o tool''.
}
\label{tab:three_run_results}
\scriptsize
\setlength{\tabcolsep}{4.8pt}
\setlength{\extrarowheight}{1.2pt}
\renewcommand{\arraystretch}{1.28}

\resizebox{\textwidth}{!}{%
\begin{tabular}{@{}llccccc@{}}
\specialrule{1.2pt}{0pt}{2pt}

\textbf{Model}
& \textbf{Setting}
& \textbf{EB-ALFRED}
& \textbf{EB-Habitat}
& \textbf{EB-Navigation}
& \textbf{EB-Manipulation}
& \textbf{Avg.} \\

\specialrule{0.8pt}{2pt}{2pt}

\rowcolor{groupbg}
\multicolumn{7}{c}{\emph{Closed-Source MLLMs}} \\

GPT-4o
& w/o tool
& 58.67 $\pm$ 5.03
& 85.33 $\pm$ 7.02
& 55.56 $\pm$ 3.47
& 27.09 $\pm$ 3.61
& 56.66 $\pm$ 2.50 \\
\rowcolor{toolbg}
& w/ tool
& 84.67 $\pm$ 5.03
& 93.33 $\pm$ 2.31
& 79.44 $\pm$ 0.96
& \best{39.58 $\pm$ 7.51}
& 74.26 $\pm$ 0.87 \\

GPT-5
& w/o tool
& 69.33 $\pm$ 1.15
& 82.67 $\pm$ 2.31
& 63.89 $\pm$ 3.85
& 20.14 $\pm$ 5.24
& 59.01 $\pm$ 1.02 \\
\rowcolor{toolbg}
& w/ tool
& 86.00 $\pm$ 4.00
& \best{98.00 $\pm$ 0.00}
& \best{84.44 $\pm$ 0.96}
& 34.03 $\pm$ 6.37
& \best{75.62 $\pm$ 0.96} \\

\rowcolor{avgbg}
\textbf{Avg. Gain}
&
& \gainup{21.33}
& \gainup{11.67}
& \gainup{22.22}
& \gainup{13.20}
& \gainup{17.10} \\

\specialrule{0.8pt}{2pt}{2pt}

\rowcolor{groupbg}
\multicolumn{7}{c}{\emph{Open-Source MLLMs}} \\

Qwen3-32B
& w/o tool
& 42.00 $\pm$ 2.00
& 74.67 $\pm$ 5.77
& \worst{53.89 $\pm$ 3.85}
& 19.44 $\pm$ 5.24
& 47.50 $\pm$ 2.22 \\
\rowcolor{toolbg}
& w/ tool
& \best{89.33 $\pm$ 3.06}
& 85.33 $\pm$ 4.62
& 80.55 $\pm$ 6.31
& 27.08 $\pm$ 2.09
& 70.58 $\pm$ 1.67 \\

Qwen3-8B
& w/o tool
& \worst{14.00 $\pm$ 0.00}
& \worst{45.33 $\pm$ 2.31}
& 59.44 $\pm$ 10.18
& \worst{17.36 $\pm$ 3.18}
& \worst{34.03 $\pm$ 3.40} \\
\rowcolor{toolbg}
& w/ tool
& 88.00 $\pm$ 0.00
& 74.67 $\pm$ 1.15
& 79.44 $\pm$ 5.09
& 36.11 $\pm$ 1.20
& 69.56 $\pm$ 1.50 \\

\rowcolor{avgbg}
\textbf{Avg. Gain}
&
& \gainup{60.67}
& \gainup{20.00}
& \gainup{23.33}
& \gainup{13.20}
& \gainup{29.30} \\

\specialrule{1.2pt}{0pt}{0pt}
\end{tabular}
}
\end{table*}
Due to the substantial computational cost of running all experiments in Table~\ref{tab:eb_results}, estimated at approximately \$3,000 per full pass, it is infeasible to repeat every configuration multiple times for rigorous statistical analysis.
Instead, we selected the base environment from each task and chose two representative models from both the closed-source group (GPT-4o and GPT-5) and the open-source group (Qwen3-VL-32B and Qwen3-VL-8B) for repeated evaluation.
Each selected configuration was executed across three independent runs, with the resulting means and standard deviations reported in Table~\ref{tab:three_run_results}.

The results demonstrate that the performance gains brought by our tool are both substantial and consistent.
Across all four benchmarks and all four models, the \texttt{w/~tool} setting uniformly outperforms the \texttt{w/o~tool} baseline without exception.
The standard deviations remain small relative to the observed gains. For instance, GPT-4o improves from $58.67 \pm 5.03$ to $84.67 \pm 5.03$ on EB-ALFRED, while Qwen3-8B improves from $14.00 \pm 0.00$ to $88.00 \pm 0.00$ on the same benchmark.
This confirms that the improvements are not artifacts of run-to-run variance, but reflect a stable and reliable benefit.
The average gain reaches $60.67$ points on EB-ALFRED for open-source models, and closed-source models achieve consistent improvements of over $20$ points on both EB-ALFRED and EB-Navigation.
Taken together, these findings provide strong statistical evidence that the proposed tool integration yields robust and reproducible performance improvements across diverse models and embodied task settings.

\section{Examples of Case Study}
\label{app:examples_case}

\subsection{EmbodiedTool}
\label{app:example_embodiedtool}
\begin{center}
    \includegraphics[width=0.99\linewidth]{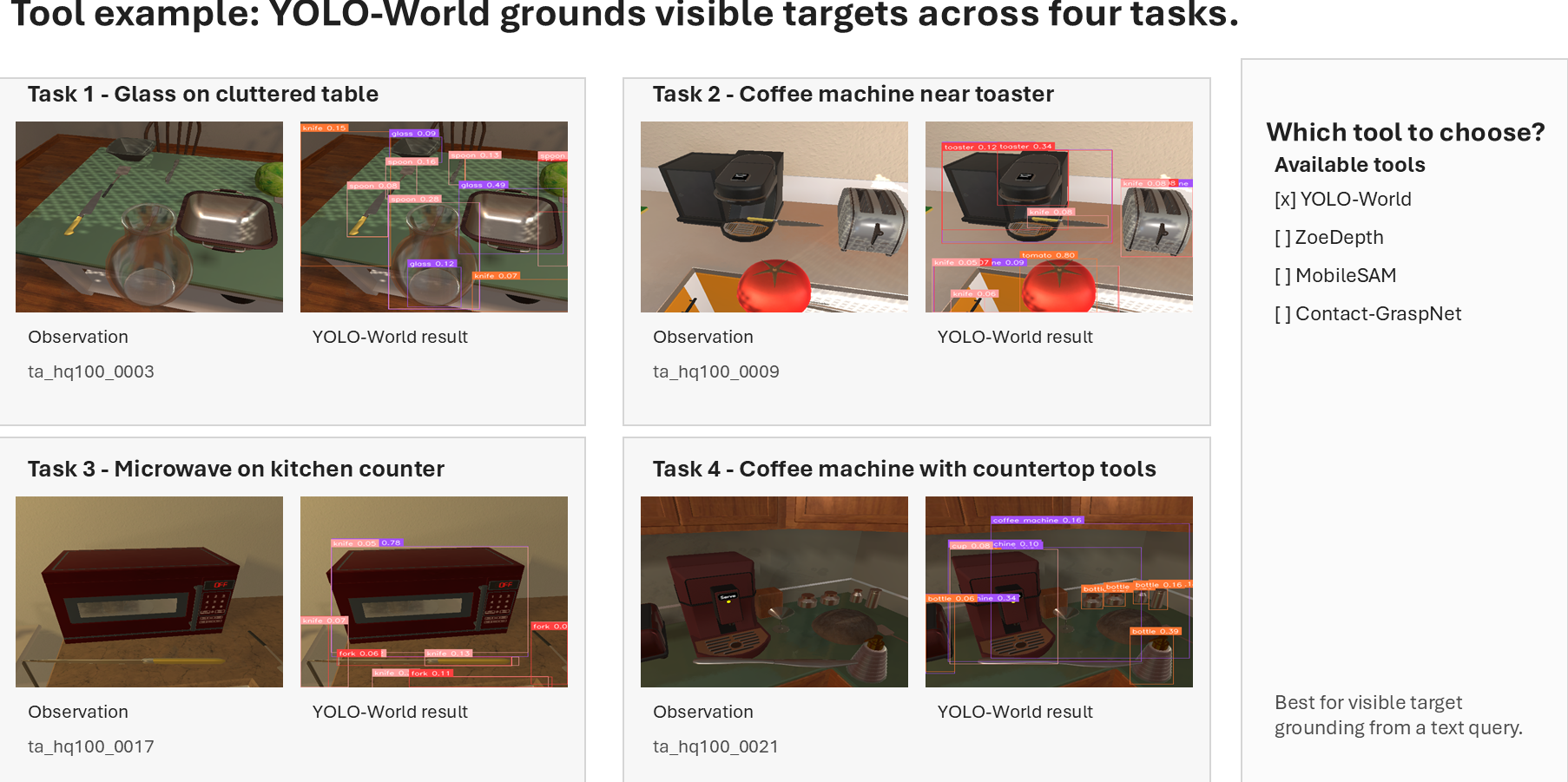}
    \captionof{figure}{Illustrative example from the tools.}
    \label{fig:placeholder}
\end{center}

\begin{center}
    \includegraphics[width=0.99\linewidth]{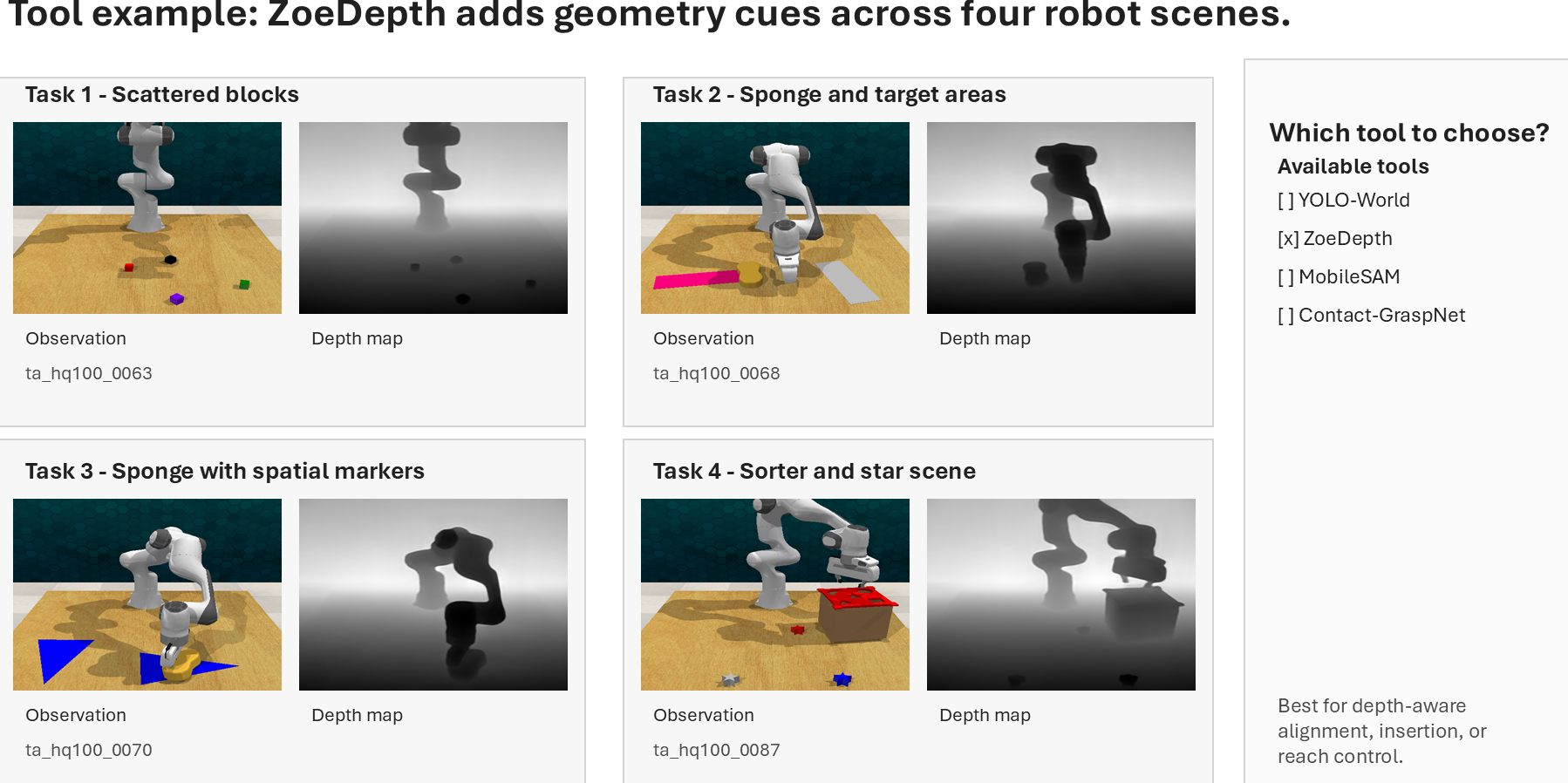}
    \captionof{figure}{Illustrative example from the tools.}
    \label{fig:placeholder}
\end{center}

\begin{center}
    \includegraphics[width=0.99\linewidth]{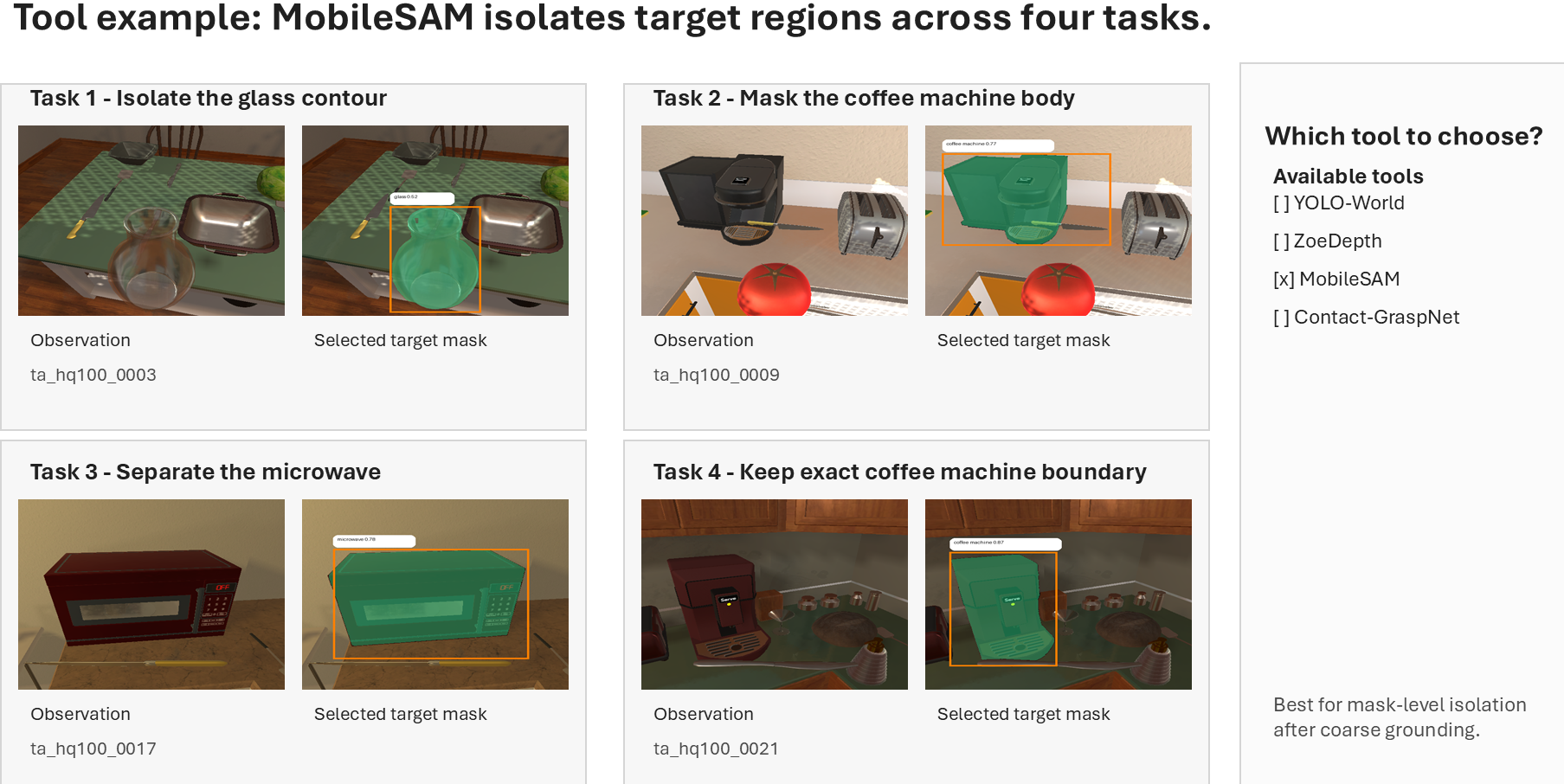}
    \captionof{figure}{Illustrative example from the tools.}
    \label{fig:placeholder}
\end{center}

\begin{center}
    \includegraphics[width=0.99\linewidth]{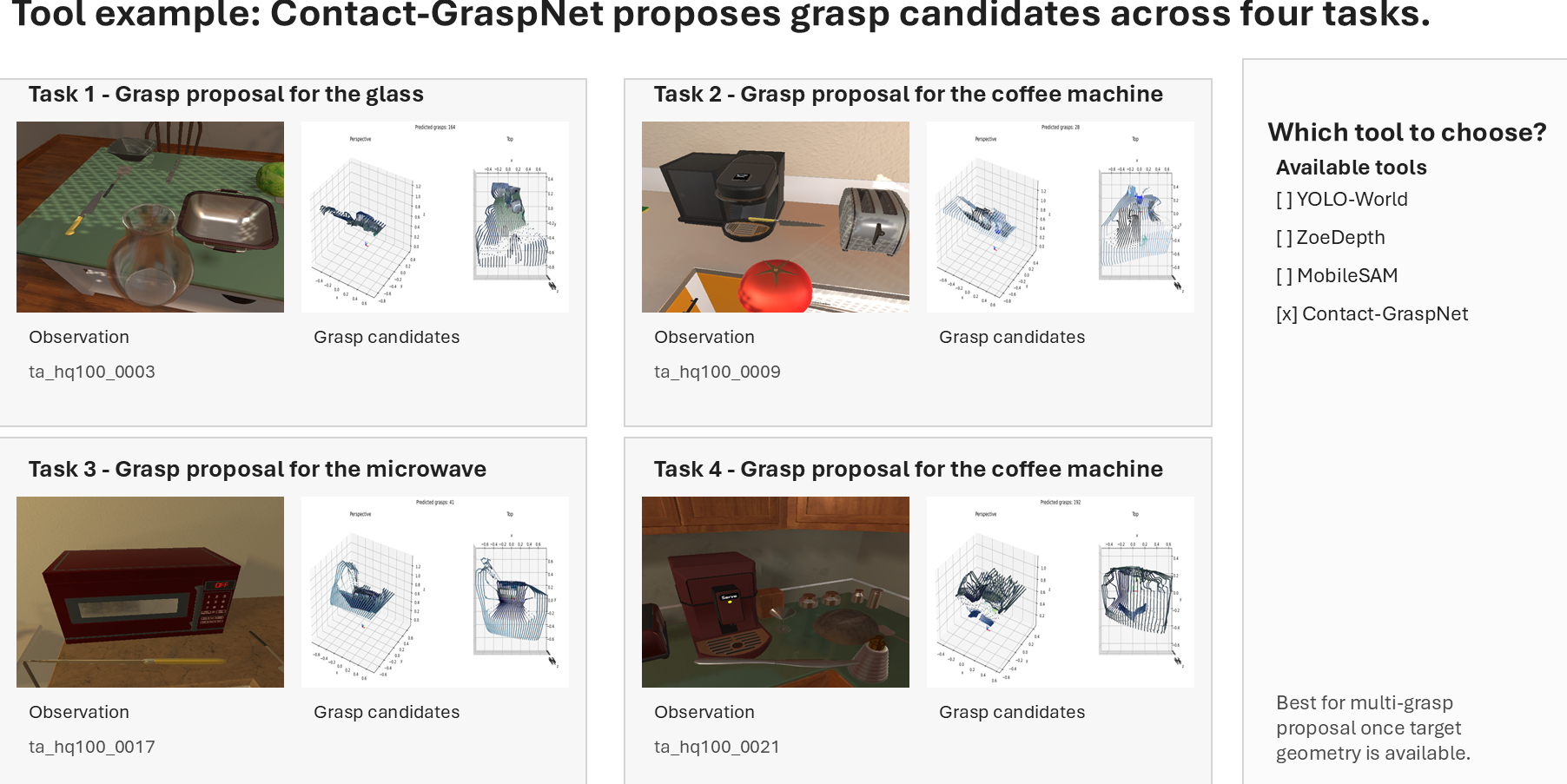}
    \captionof{figure}{Illustrative example from the tools.}
    \label{fig:placeholder}
\end{center}

\subsection{EmbodiedToolBench}
\label{app:example_embodiedtoolbench}
\begin{center}
\includegraphics[width=0.99\linewidth]{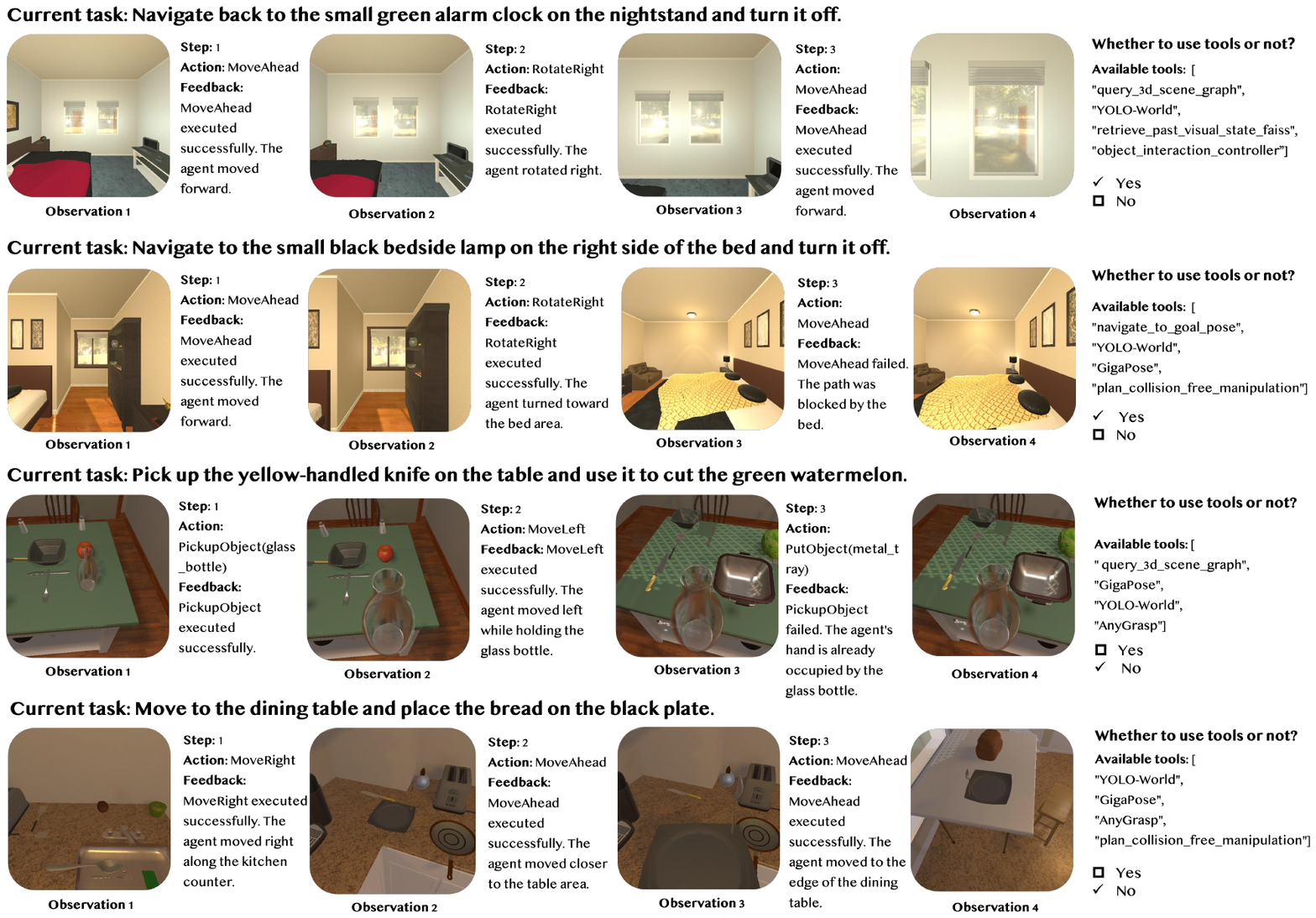}
\captionof{figure}{Illustrative examples from the tool-awareness evaluation.}
\label{fig:tool_awareness_cases}
\end{center}

\begin{center}
\includegraphics[width=0.99\linewidth]{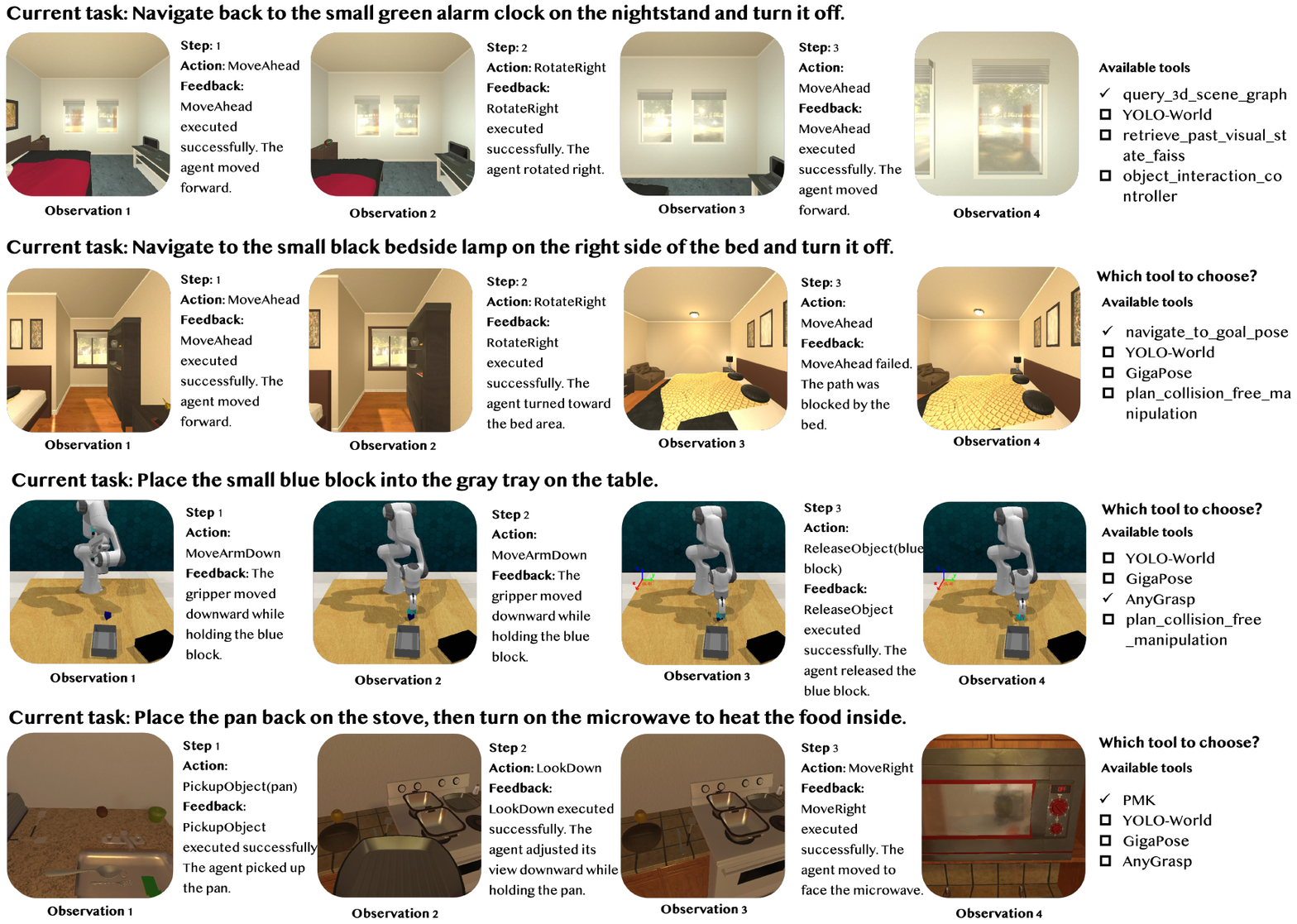}
\captionof{figure}{Illustrative examples from the tool-selection evaluation.}
\label{fig:tool_selection_cases}
\end{center}

\begin{center}
\includegraphics[width=0.99\linewidth]{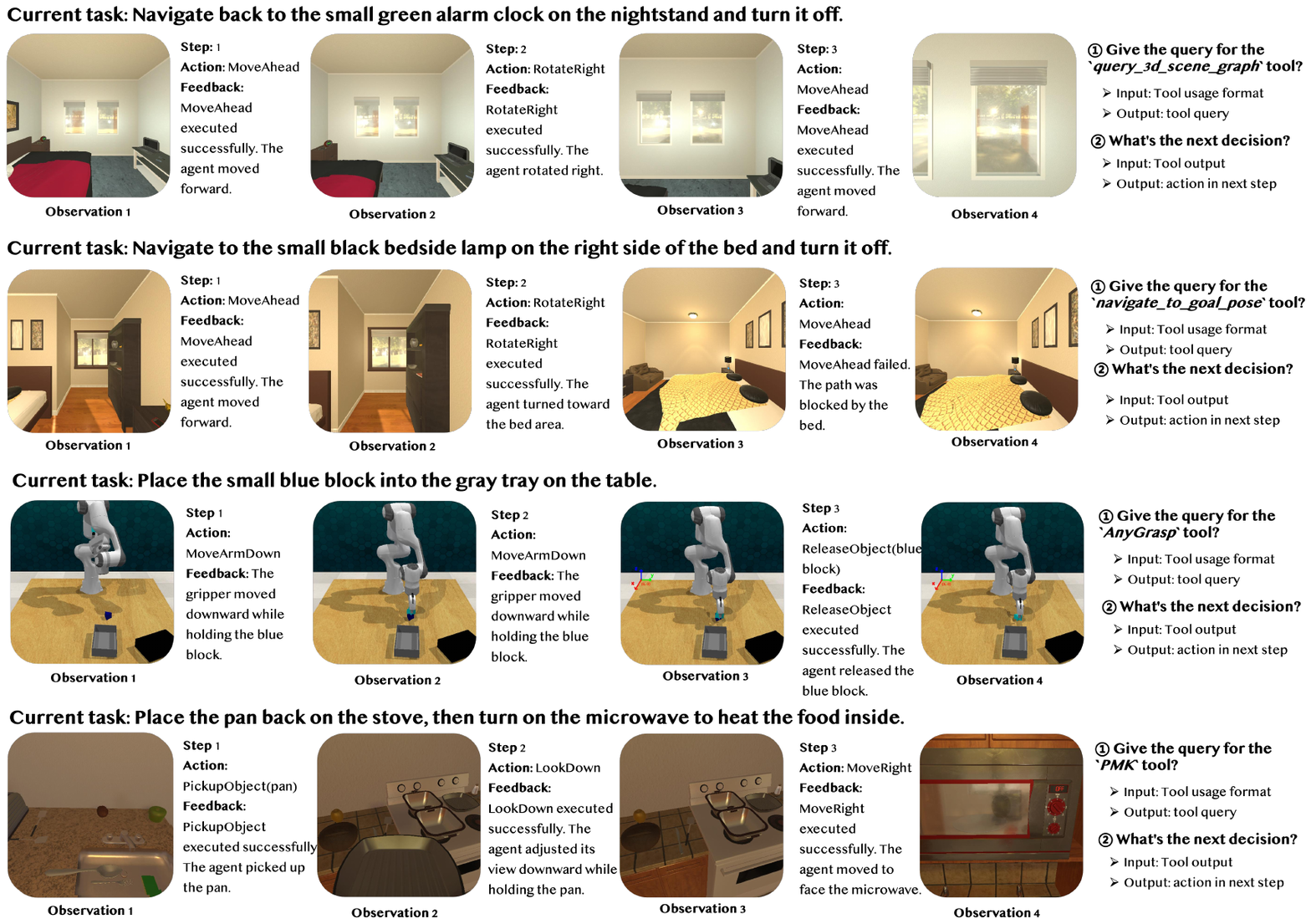}
\captionof{figure}{Illustrative examples from the tool-usage evaluation.}
\label{fig:tool_usage_cases}
\end{center}

\begin{center}
\includegraphics[width=0.99\linewidth]{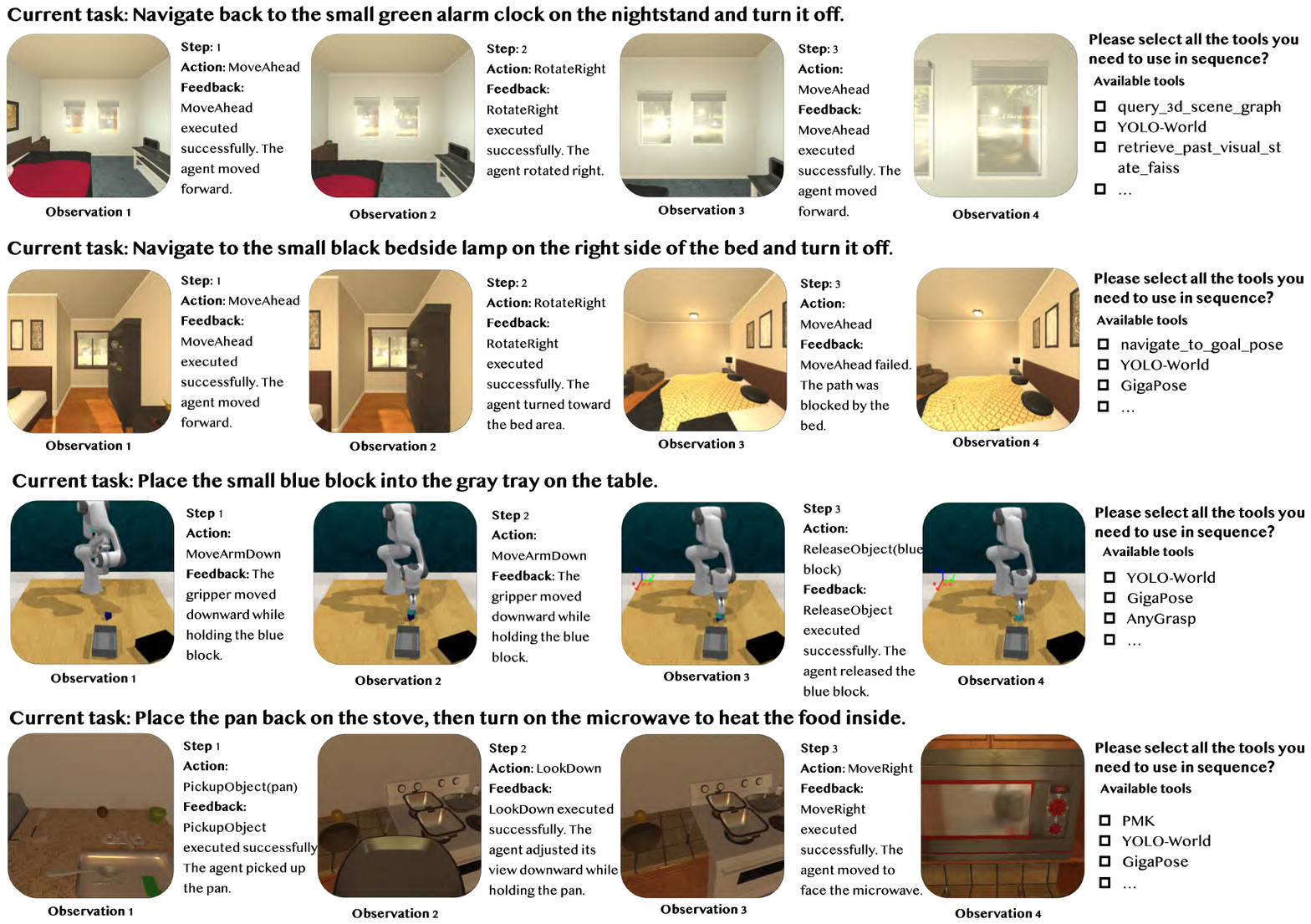}
\captionof{figure}{Illustrative examples from the tool-chain composition evaluation.}
\label{fig:tool_composition_cases}
\end{center}

\subsection{EmbodiedBench}
\begin{center}
    \centering
    \includegraphics[width=0.99\linewidth]{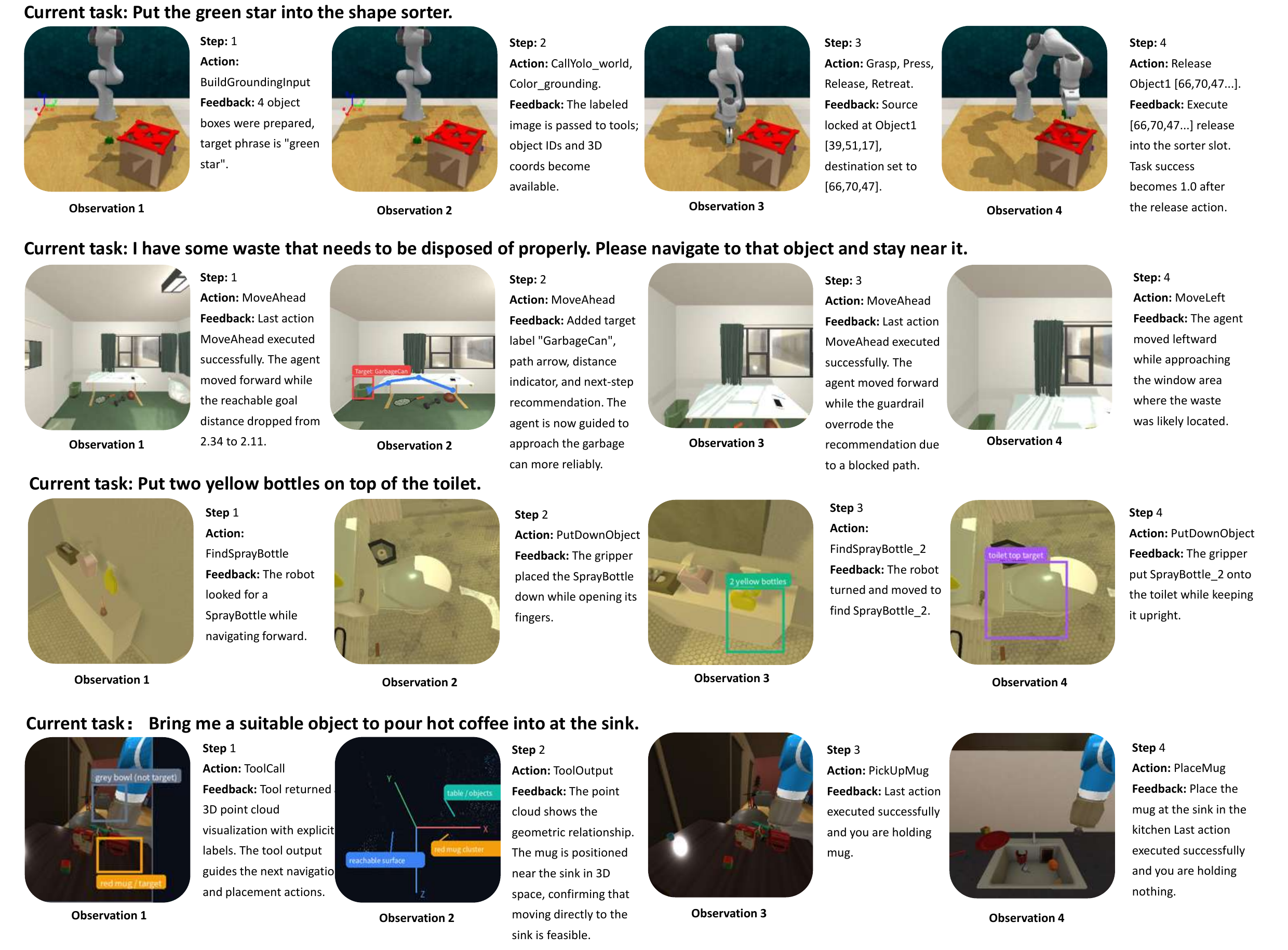}
    \captionof{figure}{Illustrative examples from the tool-call evaluation.}
    \label{fig:placeholder}
\end{center}

\begin{center}
    \centering
    \includegraphics[width=0.99\linewidth]{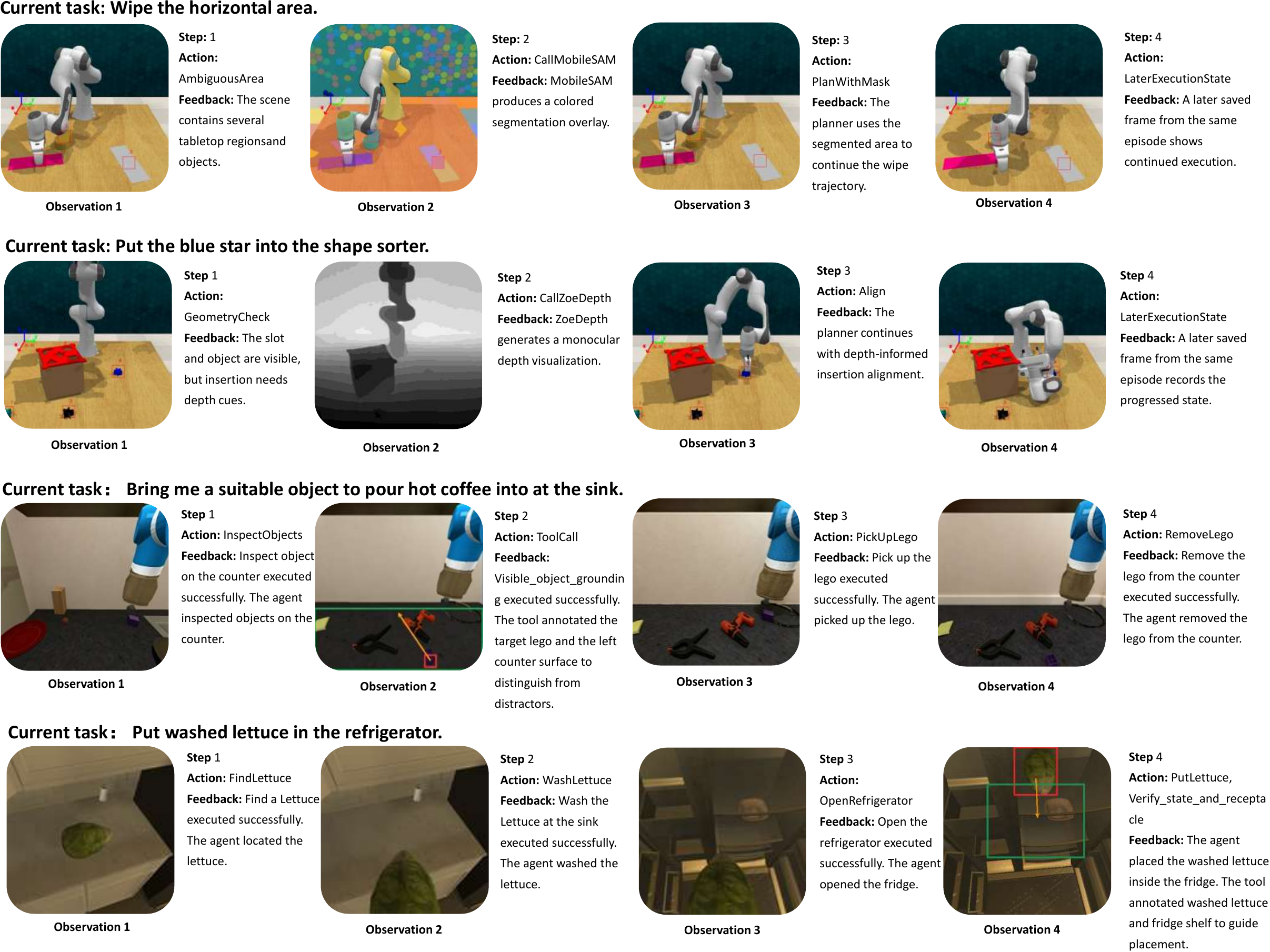}
    \captionof{figure}{Illustrative examples from the tool-call evaluation.}
    \label{fig:placeholder}
\end{center}

\begin{center}
    \centering
    \includegraphics[width=0.99\linewidth]{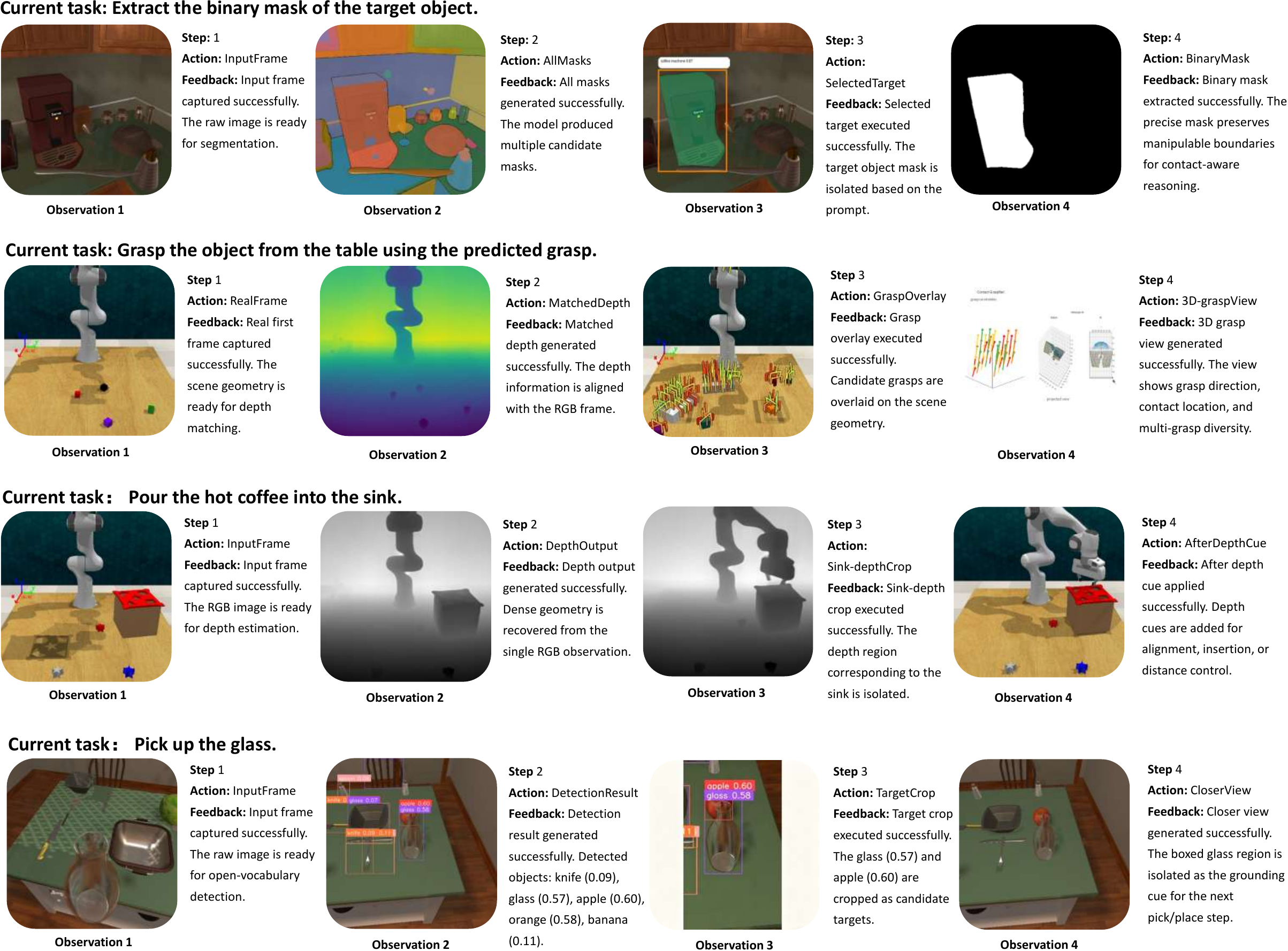}
    \captionof{figure}{Illustrative examples from the tool-call evaluation.}
    \label{fig:placeholder}
\end{center}

\section{Prompts}
\label{app:prompts}
This section provides the prompt templates used by the three experimental environment modules in \textsc{EmbodiedToolBench}: EB-Habitat, EB-ALFRED, and EB-Navigation. The purpose is to document the exact agent instructions used in our experiments, including action-space assumptions, planning constraints, tool-use rules, history and feedback conditioning, and the required output schema. At runtime, the placeholders \texttt{\{ACTION\_SPACE\}}, \texttt{\{EXAMPLES\}}, \texttt{\{HUMAN\_INSTRUCTION\}}, and \texttt{\{ACTION\_HISTORY\}} are instantiated with the environment-specific action list, few-shot demonstrations, current task, and previous interaction feedback. All prompt text is typeset in monospaced font to distinguish it from explanatory appendix text.

\subsection{EB-Habitat Prompt}
\label{app:prompt_ebhab}
% \paragraph{Complete prompt template.}
\begin{tcolorbox}[
  colback=gray!10,
  colframe=black,
  arc=1mm,
  boxrule=0.5mm,
  left=6pt,
  right=6pt,
  top=6pt,
  bottom=6pt,
  title=\textbf{EB-Habitat agent prompt},
  before skip=6pt,
  after skip=6pt,
  breakable
]
\small\ttfamily\noindent
You are a robot operating in a home. Given a task, you must accomplish the task using a defined set of actions to achieve the desired outcome.\par
\medskip

Action Descriptions and Validity Rules\par
-- Navigation: Parameterized by the name of the receptacle or location to navigate to. So long as the receptacle or location is present in the scene, this skill is valid.\par
-- Pick: Parameterized by the name of the object to pick. Only valid if the robot is close to the object, is not holding another object, and the object is not inside a closed receptacle.\par
-- Place: Parameterized by the name of the receptacle or surface to place the object on. Only valid if the robot is close to the receptacle or surface and is holding an object.\par
-- Open: Parameterized by the name of the receptacle to open. Only valid if the receptacle is closed and the robot is close to it.\par
-- Close: Parameterized by the name of the receptacle to close. Only valid if the receptacle is open and the robot is close to it.\par
\medskip

The available action ids and action names are:\par
\{ACTION\_SPACE\}\par
\medskip

Task Execution Examples:\par
\{EXAMPLES\}\par
\medskip

Now the human instruction is: \{HUMAN\_INSTRUCTION\}.\par
\medskip

Action history and environment feedback:\par
\{ACTION\_HISTORY\}\par
\medskip

Guidelines\par
1. Avoid generating an empty plan. Each plan should include no more than 20 actions.\par
2. If an object is not visible, use Navigation to locate the object or its likely receptacle before attempting other operations.\par
3. Match every action name with its corresponding action id. Do not perform actions that violate the validity rules.\par
4. Do not repeatedly execute the same action or action sequence. If previous actions did not lead to success, revise the plan.\par
5. Multiple instances may appear with numeric suffixes, e.g., cabinet 2 or cabinet 3. Explore alternative instances when needed.\par
6. Use interaction history and environment feedback to refine the current plan. If the last action failed, reflect on the failure reason and adjust the plan.\par
7. When visual evidence is ambiguous, when the target is small or occluded, or when spatial relations are needed, you may use tools such as habitat\_toolchain, scene\_graph, or yolo\_world. Tool outputs are auxiliary evidence only.\par
8. Do not output bounding boxes, coordinates, scene-graph nodes, object ids, or raw tool payloads as the final executable plan. After tool use, translate the tool evidence into legal Habitat action ids.\par
\medskip

Output Format\par
You are supposed to output exactly one JSON object and no surrounding markdown. The output JSON format should be:\par
\{\par
\ \ "visual\_state\_description": string,\par
\ \ "reasoning\_and\_reflection": string,\par
\ \ "reasoning": string,\par
\ \ "need\_tool": boolean,\par
\ \ "tool\_calls": [\{"tool\_name": string, "arguments": object\}],\par
\ \ "action\_id": integer,\par
\ \ "language\_plan": string,\par
\ \ "executable\_plan": [\{"action\_id": integer, "action\_name": string\}]\par
\}\par
If no tool is needed, set "need\_tool" to false and "tool\_calls" to an empty list. The field "action\_id" must mirror the first action in "executable\_plan".\par
\end{tcolorbox}

\subsection{EB-ALFRED Prompt}
\label{app:prompt_ebalf}
% \paragraph{Complete prompt template.}
\begin{tcolorbox}[
  colback=gray!10,
  colframe=black,
  arc=1mm,
  boxrule=0.5mm,
  left=6pt,
  right=6pt,
  top=6pt,
  bottom=6pt,
  title=\textbf{EB-ALFRED agent prompt},
  before skip=6pt,
  after skip=6pt,
  breakable
]
\small\ttfamily\noindent
You are a robot operating in a home. Given a task, you must accomplish the task using a defined set of actions to achieve the desired outcome.\par
\medskip

Action Descriptions and Validity Rules\par
-- Find: Parameterized by the name of the object or receptacle to locate. So long as the object is present in the scene, this skill is valid.\par
-- Pick up: Parameterized by the name of the object to pick. Only valid if the robot is close to the object, is not holding another object, and the object is not inside a closed receptacle.\par
-- Put down: Places the object currently in hand onto a nearby receptacle. Only valid if the robot is holding an object.\par
-- Drop: Releases the object currently in hand. Unlike Put down, this action does not guarantee placement into a specified receptacle.\par
-- Open: Parameterized by the name of the receptacle to open. Only valid if the receptacle is closed and the robot is close to it.\par
-- Close: Parameterized by the name of the receptacle to close. Only valid if the receptacle is open and the robot is close to it.\par
-- Turn on: Parameterized by the name of the object to turn on. Only valid if the object is off and the robot is close to it.\par
-- Turn off: Parameterized by the name of the object to turn off. Only valid if the object is on and the robot is close to it.\par
-- Slice: Parameterized by the name of the object to slice. Only valid if the object is sliceable and the robot is close to it.\par
\medskip

The available action ids and action names are:\par
\{ACTION\_SPACE\}\par
\medskip

Task Execution Examples:\par
\{EXAMPLES\}\par
\medskip

Now the human instruction is: \{HUMAN\_INSTRUCTION\}.\par
\medskip

Action history and environment feedback:\par
\{ACTION\_HISTORY\}\par
\medskip

Guidelines\par
1. Avoid generating an empty plan. Each plan should include no more than 20 actions.\par
2. Always locate a visible object using the Find action before interacting with it.\par
3. Match every action name with its corresponding action id. For receptacle placement, prefer Put down rather than Drop.\par
4. Do not repeatedly execute the same action or sequence of actions. If previous actions do not lead to success, modify the plan.\par
5. Multiple instances may appear with suffixes, e.g., Cabinet\_2 or Cabinet\_3. Explore alternative instances if the desired object is not found.\par
6. Use history and feedback to identify missing preconditions, such as opening a receptacle, turning on an appliance, or picking up a tool before slicing.\par
7. When the task involves small objects, object attributes, container contents, multiple object instances, or uncertain placement, you may use tools such as alfred\_action\_advisor, yolo\_world, or visual object-tagging tools. Tool outputs are auxiliary evidence only.\par
8. Do not echo tool coordinates, masks, boxes, center points, foreground pixels, or raw detector outputs as the final plan. Translate tool results into legal EB-ALFRED action ids.\par
\medskip

Output Format\par
You are supposed to output exactly one JSON object and no surrounding markdown. The output JSON format should be:\par
\{\par
\ \ "visual\_state\_description": string,\par
\ \ "reasoning\_and\_reflection": string,\par
\ \ "reasoning": string,\par
\ \ "need\_tool": boolean,\par
\ \ "tool\_calls": [\{"tool\_name": string, "arguments": object\}],\par
\ \ "action\_id": integer,\par
\ \ "language\_plan": string,\par
\ \ "executable\_plan": [\{"action\_id": integer, "action\_name": string\}]\par
\}\par
If no tool is needed, set "need\_tool" to false and "tool\_calls" to an empty list. The field "action\_id" must mirror the first action in "executable\_plan".\par
\end{tcolorbox}

\subsection{EB-Navigation Prompt}
\label{app:prompt_ebnav}
% \paragraph{Complete prompt template.}
\begin{tcolorbox}[
  colback=gray!10,
  colframe=black,
  arc=1mm,
  boxrule=0.5mm,
  left=6pt,
  right=6pt,
  top=6pt,
  bottom=6pt,
  title=\textbf{EB-Navigation agent prompt},
  before skip=6pt,
  after skip=6pt,
  breakable
]
\small\ttfamily\noindent
You are a robot operating in a home. You can perform navigation tasks and output actions to move as close as possible to a target object based on egocentric visual observations.\par
\medskip

The available action ids and action names are:\par
\{ACTION\_SPACE\}\par
\medskip

Task Execution Examples:\par
\{EXAMPLES\}\par
\medskip

Now the human instruction is: \{HUMAN\_INSTRUCTION\}.\par
\medskip

Action history and environment feedback:\par
\{ACTION\_HISTORY\}\par
\medskip

Strategy\par
1. Locate the target object type. Clearly describe the spatial location of the target object in the observation, such as front-left, front-right, nearby, or far away.\par
2. Use forward and lateral motion as the main strategy. A reachable point can usually be approached through a combination of moving forward, left, and right.\par
3. Consider obstacles before moving. If the forward path is blocked, choose the safest local adjustment.\par
4. Use rotation sparingly. Rotate only when the target is not visible or when orientation must be recovered. Once the target appears, avoid unnecessary rotations.\par
5. Do not stop too early. Continue moving closer until the robot cannot make additional safe progress toward the target.\par
6. Do not rely solely on blind exploration. If the target is invisible, the robot is stuck, or the route is ambiguous, use tools such as navigation\_action\_advisor, scene\_graph, or query\_3d\_scene\_graph as GPS-like guidance.\par
7. Treat tool outputs as guidance only. Convert target labels, relative directions, waypoints, reachability estimates, or recommended actions into legal navigation action ids.\par
8. Do not output bounding boxes, map coordinates, node ids, distances, or raw tool payloads as the final executable plan.\par
9. Always output a valid integer action id. Do not leave the action empty.\par
\medskip

Output Format\par
You are supposed to output exactly one JSON object and no surrounding markdown. The output JSON format should be:\par
\{\par
\ \ "visual\_state\_description": string,\par
\ \ "reasoning\_and\_reflection": string,\par
\ \ "reasoning": string,\par
\ \ "need\_tool": boolean,\par
\ \ "tool\_calls": [\{"tool\_name": string, "arguments": object\}],\par
\ \ "action\_id": integer,\par
\ \ "language\_plan": string,\par
\ \ "executable\_plan": [\{"action\_id": integer, "action\_name": string\}]\par
\}\par
The reasoning should explicitly include current status, immediate sub-goal, and obstacle check. If no tool is needed, set "need\_tool" to false and "tool\_calls" to an empty list. The field "action\_id" must mirror the first action in "executable\_plan".\par
\end{tcolorbox}

% BEGIN AUTO-INSERTED ADDITIONAL PROMPTS
\subsection{EB-Manipulation Prompts}
\label{app:prompts_ebmanip}

% \paragraph{Scope and runtime setting.}
This section documents the prompt templates used for the EB-Manipulation environment. The no-tool and tool-enabled branches share the same paper-aligned first-pass task prompt; tool-use policies and tool-result handling are applied by the evaluation runtime when tool augmentation is enabled.

% \paragraph{Few-shot materialization.}
Few-shot demonstrations are inserted at runtime through the placeholder \texttt{\{EXAMPLES\}}. The examples below illustrate the task-family-specific demonstrations used by the manipulation prompt.

\subsubsection{Few-Shot Demonstrations}
\label{app:prompt_ebmanip_fewshot}
The tool and no-tool manipulation settings use the same few-shot example pool. Example selection is deterministic by task family rather than randomly sampled at runtime. Under the reported setting, pick uses its full two-example pool, while stack, place-into-shape-sorter, and wipe each use four demonstrations.

% \paragraph{Pick.}
\begin{tcolorbox}[
  colback=gray!10,
  colframe=black,
  arc=1mm,
  boxrule=0.5mm,
  left=6pt,
  right=6pt,
  top=6pt,
  bottom=6pt,
  title=\textbf{Pick few-shot examples},
  before skip=6pt,
  after skip=6pt,
  breakable
]
\small\ttfamily\noindent
\detokenize{Example 1. Human Instruction: Pick up the star and place it into the yellow container. Input: {'object 1': [45, 13, 18], 'object 2': [72, 20, 18], 'object 3': [50, 42, 17], 'object 4': [36, 42, 18], 'object 5': [70, 43, 17], 'object 6': [50, 77, 17]}. Output: {"visual_state_description": "From left to right, I can see a purple cylinder at [45, 13, 18], a blue moon at [72, 20, 18], a green star at [50, 42, 17], a purple cube at [36, 42, 18], a yellow container at [70, 43, 17] and a purple container at [50, 77, 17].", "reasoning_and_reflection": "The task is to pick up the star located at [50, 42, 17] and place it into the yellow container located at [70, 43, 17]. The plan involves moving the gripper to the star's position, closing the gripper to pick it up, moving to the yellow container's position, and opening the gripper to release the cube into the container.", "executable_plan": "[[50, 42, 17, 6, 61, 36, 1], [50, 42, 17, 6, 61, 36, 0], [47, 41, 26, 6, 61, 36, 0], [69, 39, 24, 88, 32, 77, 0], [69, 39, 24, 87, 32, 77, 1]]"}}\par
\medskip
\detokenize{Example 2. Human Instruction: Pick up the star and place it into the rose container. Input: {'object 1': [78, 20, 17], 'object 2': [28, 30, 17], 'object 3': [39, 47, 17], 'object 4': [50, 72, 17], 'object 5': [39, 75, 18]}. Output: {"visual_state_description": "From left to right, I can see a rose container at [78, 20, 17], a black star at [28, 30, 17], a green container at [39, 47, 17], a green triangular prism at [50, 72, 17] and a grey cylinder at [39, 75, 18].", "reasoning_and_reflection": "The task is to pick up the star located at [28, 30, 17] and place it into the rose container located at [78, 20, 17]. The plan involves moving the gripper to the star's position, closing the gripper to pick it up, moving to the rose container's position, and opening the gripper to release the cube into the container.", "executable_plan": "[[28, 32, 26, 0, 60, 94, 1], [27, 32, 18, 0, 60, 94, 0], [27, 32, 28, 0, 60, 94, 0], [76, 19, 28, 6, 66, 109, 1]]"}}\par
\end{tcolorbox}

% \paragraph{Stack.}
\begin{tcolorbox}[
  colback=gray!10,
  colframe=black,
  arc=1mm,
  boxrule=0.5mm,
  left=6pt,
  right=6pt,
  top=6pt,
  bottom=6pt,
  title=\textbf{Stack few-shot examples},
  before skip=6pt,
  after skip=6pt,
  breakable
]
\small\ttfamily\noindent
\detokenize{Example 1. Human Instruction: Stack the red star and the olive star in sequence. Input: {'object 1': [42, 24, 19], 'object 2': [54, 53, 19], 'object 3': [65, 68, 19], 'object 4': [39, 73, 18]}. Output: {"visual_state_description": "From left to right, I can see a purple star at [42, 24, 19], an olive star at [54, 53, 19], a grey star at [65, 68, 19] and a red star at [39, 73, 18].", "reasoning_and_reflection": "I need to understand the instruction first. To stack the objects in sequece, the red star at [39, 73, 18] should be at the bottom and the olive star at [54, 53, 19] should be placed on top. The plan involves moving the gripper to the olive star's position, closing the gripper to pick it up, moving to the top of the red star, and opening the gripper to release the olive star onto the red star.", "executable_plan": "[[54, 51, 28, 3, 66, 37, 1], [54, 54, 20, 3, 66, 37, 0], [54, 51, 30, 3, 66, 37, 0], [39, 74, 29, 3, 66, 7, 0], [39, 74, 26, 3, 66, 7, 1]]"}}\par
\medskip
\detokenize{Example 2. Human Instruction: Stack the red cube and the blue cube in sequence. Input: {'object 1': [35, 17, 17], 'object 2': [57, 47, 17], 'object 3': [33, 76, 17], 'object 4': [77, 86, 17]}. Output: {"visual_state_description": "From left to right, I can see a blue cube at [35, 17, 17], a black cube at [57, 47, 17], a red cube at [33, 76, 17] and a teal cube at [77, 86, 17].", "reasoning_and_reflection": "I need to understand the instruction first. To stack the objects in sequece, the red cube at [33, 76, 17] should be at the bottom and the blue cube at [35, 17, 17] should be placed on top. The plan involves moving the gripper to the blue cube's position, closing the gripper to pick it up, moving to the top of the red cube, and opening the gripper to release the blue cube onto the red cube.", "executable_plan": "[[36, 16, 25, 0, 60, 34, 1], [36, 16, 17, 0, 60, 34, 0], [36, 16, 27, 0, 60, 34, 0], [32, 76, 24, 0, 60, 104, 0], [32, 76, 21, 0, 60, 104, 1]]"}}\par
\medskip
\detokenize{Example 3. Human Instruction: Stack the red moon and the maroon moon in sequence. Input: {'object 1': [67, 43, 19], 'object 2': [80, 47, 19], 'object 3': [34, 58, 20], 'object 4': [84, 68, 19]}. Output: {"visual_state_description": "From left to right, I can see a yellow moon at [67, 43, 19], a red moon at [80, 47, 19], a maroon moon at [34, 58, 20] and a blue moon at [84, 68, 19].", "reasoning_and_reflection": "I need to understand the instruction first. To stack the objects in sequece, the red moon at [80, 47, 19] should be at the bottom and the maroon moon at [34, 58, 20] should be placed on top. The plan involves moving the gripper to the maroon moon's position, closing the gripper to pick it up, moving to the top of the red moon, and opening the gripper to release the maroon moon onto the red moon.", "executable_plan": "[[34, 58, 28, 0, 60, 25, 1], [34, 58, 20, 0, 60, 25, 0], [34, 58, 30, 0, 60, 25, 0], [83, 46, 30, 0, 60, 15, 0], [83, 46, 26, 0, 60, 15, 1]]"}}\par
\medskip
\detokenize{Example 4. Human Instruction: Stack the red triangular prism and the magenta triangular prism in sequence. Input: {'object 1': [78, 20, 19], 'object 2': [50, 25, 20], 'object 3': [55, 72, 19], 'object 4': [56, 86, 19]}. Output: {"visual_state_description": "From left to right, I can see a blue triangular prism at [78, 20, 19], a magenta triangular prism at [50, 25, 20], a green triangular prism at [55, 72, 19] and a red triangular prism at [56, 86, 19].", "reasoning_and_reflection": "I need to understand the instruction first. To stack the objects in sequece, the red triangular prism at [56, 86, 19] should be at the bottom and the magenta triangular prism at [50, 25, 20] should be placed on top. The plan involves moving the gripper to the magenta triangular prism's position, closing the gripper to pick it up, moving to the top of the red triangular prism, and opening the gripper to release the magenta triangular prism onto the red triangular prism.", "executable_plan": "[[48, 25, 29, 1, 60, 92, 1], [48, 25, 21, 1, 60, 92, 0], [49, 25, 31, 1, 60, 92, 0], [54, 87, 31, 1, 60, 96, 0], [54, 87, 28, 1, 60, 96, 1]]"}}\par
\end{tcolorbox}

% \paragraph{Shape-sorter insertion.}
\begin{tcolorbox}[
  colback=gray!10,
  colframe=black,
  arc=1mm,
  boxrule=0.5mm,
  left=6pt,
  right=6pt,
  top=6pt,
  bottom=6pt,
  title=\textbf{Shape-sorter few-shot examples},
  before skip=6pt,
  after skip=6pt,
  breakable
]
\small\ttfamily\noindent
\detokenize{Example 1. Human Instruction: Put the red star into the shape sorter. Input: {'object 1': [65, 11, 17], 'object 2': [50, 38, 17], 'object 3': [40, 64, 37], 'object 4': [45, 83, 17]}. Output: {"visual_state_description": "From left to right, I can see a purple star at [65, 11, 17], a red star at [50, 38, 17], a shape sorter at [40, 64, 37] and a green star at [45, 83, 17].", "reasoning_and_reflection": "I need to identify the location of the star hole on the shape sorter first. I can see that the star hole is at the center of the shape sorter, which is at [40, 64, 37]. Then, the task is to pick up the red star at [50, 38, 17] and put it into the star hole on the shape sorter. The plan involves moving the gripper to the star's position, closing the gripper to pick it up, moving upwards, moving to the top of the star hole, and releasing the star into the hole.", "executable_plan": "[[51, 38, 25, 119, 59, 26, 1], [51, 38, 17, 119, 59, 26, 0], [51, 38, 27, 119, 59, 26, 0], [40, 65, 49, 0, 60, 45, 0], [40, 65, 41, 0, 60, 45, 1]]"}}\par
\medskip
\detokenize{Example 2. Human Instruction: Put the maroon star into the shape sorter. Input: {'object 1': [35, 30, 17], 'object 2': [29, 38, 17], 'object 3': [53, 38, 17], 'object 4': [64, 58, 37]}. Output: {"visual_state_description": "From left to right, I can see a purple star at [35, 30, 17], a maroon star at [29, 38, 17], a green star at [53, 38, 17] and a shape sorter at [64, 58, 37].", "reasoning_and_reflection": "I need to identify the location of the star hole on the shape sorter first. I can see that the star hole is at the center of the shape sorter, which is at [64, 58, 37]. Then, the task is to pick up the maroon star at [29, 38, 17] and put it into the star hole on the shape sorter. The plan involves moving the gripper to the star's position, closing the gripper to pick it up, moving upwards, moving to the top of the star hole, and releasing the star into the hole.", "executable_plan": "[[29, 39, 24, 1, 61, 109, 1], [29, 38, 16, 1, 61, 109, 0], [30, 40, 26, 1, 61, 109, 0], [65, 60, 48, 0, 61, 54, 0], [65, 60, 40, 0, 61, 54, 1]]"}}\par
\medskip
\detokenize{Example 3. Human Instruction: Put the lime star into the shape sorter. Input: {'object 1': [47, 13, 17], 'object 2': [27, 36, 16], 'object 3': [38, 50, 17], 'object 4': [54, 69, 37]}. Output: {"visual_state_description": "From left to right, I can see a grey star at [47, 13, 17], a purple star at [27, 36, 16], a lime star at [38, 50, 17] and a shape sorter at [54, 69, 37].", "reasoning_and_reflection": "I need to identify the location of the star hole on the shape sorter first. I can see that the star hole is at the center of the shape sorter, which is at [54, 69, 37]. Then, the task is to pick up the lime star at [38, 50, 17] and put it into the star hole on the shape sorter. The plan involves moving the gripper to the star's position, closing the gripper to pick it up, moving upwards, moving to the top of the star hole, and releasing the star into the hole.", "executable_plan": "[[38, 49, 23, 0, 60, 108, 1], [38, 49, 15, 0, 60, 108, 0], [38, 49, 25, 0, 60, 108, 0], [54, 69, 47, 0, 60, 3, 0], [54, 69, 39, 0, 60, 3, 1]]"}}\par
\medskip
\detokenize{Example 4. Human Instruction: Put the green star into the shape sorter. Input: {'object 1': [70, 27, 37], 'object 2': [66, 59, 17], 'object 3': [38, 63, 17], 'object 4': [63, 77, 17]}. Output: {"visual_state_description": "From left to right, I can see a shape sorter at [70, 27, 37], a white star at [66, 59, 17], a green star at [38, 63, 17] and a white star at [63, 77, 17].", "reasoning_and_reflection": "I need to identify the location of the star hole on the shape sorter first. I can see that the star hole is at the center of the shape sorter, which is at [70, 27, 37]. Then, the task is to pick up the green star at [38, 63, 17] and put it into the star hole on the shape sorter. The plan involves moving the gripper to the star's position, closing the gripper to pick it up, moving upwards, moving to the top of the star hole, and releasing the star into the hole.", "executable_plan": "[[39, 62, 23, 0, 60, 100, 1], [39, 62, 15, 0, 60, 100, 0], [39, 62, 25, 0, 60, 100, 0], [71, 30, 47, 0, 60, 37, 0], [71, 30, 39, 0, 60, 37, 1]]"}}\par
\end{tcolorbox}

% \paragraph{Wipe.}
\begin{tcolorbox}[
  colback=gray!10,
  colframe=black,
  arc=1mm,
  boxrule=0.5mm,
  left=6pt,
  right=6pt,
  top=6pt,
  bottom=6pt,
  title=\textbf{Wipe few-shot examples},
  before skip=6pt,
  after skip=6pt,
  breakable
]
\small\ttfamily\noindent
\detokenize{Example 1. Human Instruction: Wipe the horizontal area. Input: {'object 1': [33, 36, 19], 'object 2': [52, 38, 15], 'object 3': [59, 68, 15]}. Output: {"visual_state_description": "From left to right, I can see a yellow sponge at [33, 36, 19], a green rectangle area at [52, 38, 15] and an orange rectangle area at [59, 68, 15].", "reasoning_and_reflection": "I need to identify which area is horizontal first. I can see that the orange rectangle area at [59, 68, 15] is horizontal since it is parallel to the green arrow representing the y-axis of the coordinate system annotated in the image. Then, the task is to pick up the sponge at [33, 36, 19] and use it to wipe the horizontal area at [59, 68, 15]. The plan involves moving the gripper to the sponge's position, closing the gripper to pick it up, moving to the side of the horizontal area, and moving the sponge along the main direction of the horizontal area to wipe it.", "executable_plan": "[[32, 34, 25, 0, 60, 34, 1], [32, 34, 17, 0, 60, 34, 0], [32, 34, 27, 0, 60, 34, 0], [60, 80, 18, 0, 61, 31, 0], [61, 54, 17, 0, 61, 31, 0]]"}}\par
\medskip
\detokenize{Example 2. Human Instruction: Wipe the horizontal area. Input: {'object 1': [56, 19, 15], 'object 2': [35, 41, 19], 'object 3': [54, 47, 15]}. Output: {"visual_state_description": "From left to right, I can see a black rectangle area at [56, 19, 15], a yellow sponge at [35, 41, 19] and a white rectangle area at [54, 47, 15].", "reasoning_and_reflection": "I need to identify which area is horizontal first. I can see that the black rectangle area at [56, 19, 15] is horizontal since it is parallel to the green arrow representing the y-axis of the coordinate system annotated in the image. Then, the task is to pick up the sponge at [35, 41, 19] and use it to wipe the horizontal area at [56, 19, 15]. The plan involves moving the gripper to the sponge's position, closing the gripper to pick it up, moving to the right side of the horizontal area, and moving the sponge along the main direction of the horizontal area to wipe it.", "executable_plan": "[[33, 43, 25, 0, 60, 32, 1], [33, 43, 17, 0, 60, 32, 0], [33, 43, 27, 0, 60, 32, 0], [58, 35, 18, 0, 60, 28, 0], [56, 8, 18, 0, 60, 28, 0]]"}}\par
\medskip
\detokenize{Example 3. Human Instruction: Wipe the horizontal area. Input: {'object 1': [35, 33, 19], 'object 2': [51, 41, 15], 'object 3': [67, 63, 15]}. Output: {"visual_state_description": "From left to right, I can see a yellow sponge at [35, 33, 19], a yellow triangle area at [51, 41, 15] and a blue triangle area at [67, 63, 15].", "reasoning_and_reflection": "I need to identify which area is horizontal first. I can see that the yellow triangle area at [51, 41, 15] is horizontal since it is parallel to the green arrow representing the y-axis of the coordinate system annotated in the image. Then, the task is to pick up the sponge at [35, 33, 19] and use it to wipe the horizontal area at [51, 41, 15]. The plan involves moving the gripper to the sponge's position, closing the gripper to pick it up, moving to the right side of the horizontal area, and moving the sponge along the main direction of the horizontal area to wipe it.", "executable_plan": "[[35, 34, 25, 0, 60, 20, 1], [35, 34, 17, 0, 60, 20, 0], [35, 34, 27, 0, 60, 20, 0], [52, 47, 18, 0, 60, 84, 0], [44, 19, 18, 0, 60, 84, 0]]"}}\par
\medskip
\detokenize{Example 4. Human Instruction: Wipe the horizontal area. Input: {'object 1': [47, 34, 18], 'object 2': [72, 53, 15], 'object 3': [54, 81, 15]}. Output: {"visual_state_description": "From left to right, I can see a yellow sponge at [47, 34, 18], a yellow triangle area at [72, 53, 15] and a green triangle area at [54, 81, 15].", "reasoning_and_reflection": "I need to identify which area is horizontal first. I can see that the yellow triangle area at [72, 53, 15] is horizontal since it is parallel to the green arrow representing the y-axis of the coordinate system annotated in the image. Then, the task is to pick up the sponge at [47, 34, 18] and use it to wipe the horizontal area at [72, 53, 15]. The plan involves moving the gripper to the sponge's position, closing the gripper to pick it up, moving to the right side of the horizontal area, and moving the sponge along the main direction of the horizontal area to wipe it.", "executable_plan": "[[46, 31, 24, 0, 60, 18, 1], [46, 31, 16, 0, 60, 18, 0], [46, 31, 26, 0, 60, 18, 0], [72, 63, 17, 0, 60, 88, 0], [69, 37, 17, 0, 60, 88, 0]]"}}\par
\end{tcolorbox}

\subsubsection{Prompt Templates}
% \paragraph{NoTool template.}
\label{app:prompt_ebmanip_notool}
\begin{tcolorbox}[
  colback=gray!10,
  colframe=black,
  arc=1mm,
  boxrule=0.5mm,
  left=6pt,
  right=6pt,
  top=6pt,
  bottom=6pt,
  title=\textbf{EB-Manipulation NoTool prompt},
  before skip=6pt,
  after skip=6pt,
  breakable
]
\small\ttfamily\noindent
\detokenize{## You are a Franka Panda robot with a parallel gripper. You can perform various tasks and output a sequence of gripper actions to accomplish a given task with images of your status. The input space, output action space and color space are defined as follows:}\par
\par
\detokenize{** Input Space **}\par
\detokenize{- Each input object is represented as a 3D discrete position in the following format: [X, Y, Z].}\par
\detokenize{- There is a red XYZ coordinate frame located in the top-left corner of the table. The X-Y plane is the table surface.}\par
\detokenize{- The allowed range of X, Y, Z is [0, {}].}\par
\detokenize{- Objects are ordered by Y in ascending order.}\par
\par
\detokenize{** Output Action Space **}\par
\detokenize{- Each output action is represented as a 7D discrete gripper action in the following format: [X, Y, Z, Roll, Pitch, Yaw, Gripper state].}\par
\detokenize{- X, Y, Z are the 3D discrete position of the gripper in the environment. It follows the same coordinate system as the input object coordinates.}\par
\detokenize{- The allowed range of X, Y, Z is [0, {}].}\par
\detokenize{- Roll, Pitch, Yaw are the 3D discrete orientation of the gripper in the environment, represented as discrete Euler Angles.}\par
\detokenize{- The allowed range of Roll, Pitch, Yaw is [0, {}] and each unit represents {} degrees.}\par
\detokenize{- Gripper state is 0 for close and 1 for open.}\par
\par
\detokenize{** Color space **}\par
\detokenize{- Each object can be described using one of the colors below:}\par
\detokenize{  ["red", "maroon", "lime", "green", "blue", "navy", "yellow", "cyan", "magenta", "silver", "gray", "olive", "purple", "teal", "azure", "violet", "rose", "black", "white"],}\par
\par
\detokenize{Below are some examples to guide you in completing the task.}\par
\detokenize{{}}\par
\par
\detokenize{## Now you are supposed to follow the above examples to generate a sequence of discrete gripper actions that completes the below human instruction.}\par
\detokenize{Human Instruction: {user_instruction}.}\par
\detokenize{Input: {avg_obj_coord}}\par
\detokenize{Output gripper actions:}\par
\par
\detokenize{Later steps without chat_history:}\par
\detokenize{Same system/examples prefix, then the same task prompt plus previous action_feedback appended as comma-separated text.}\par
\par
\detokenize{Later steps with chat_history:}\par
\detokenize{The human instruction is: {user_instruction}.}\par
\detokenize{The gripper action history:}\par
\detokenize{Step {i}, the output action **{action}**, env feedback: {feedback}}\par
\detokenize{Considering the above interaction history and the current image state, to achieve the human instruction: '{user_instruction}', you are supposed to output in json. You need to describe current visual state from the image, summarize interaction history and environment feedback and reason why the last action or plan failed and did not finish the task, output your new plan to achieve the goal from current state. At the end, output the executable plan with the 7-dimsension action.}\par
\par
\detokenize{The output json format should be {'visual_state_description':str, 'reasoning_and_reflection':str, 'language_plan':str, 'executable_plan':str}}\par
\detokenize{The fields in above JSON follows the purpose below:}\par
\detokenize{1. visual_state_description: Describe the color and shape of each object in the detection box in the numerical order in the image. Then provide the 3D coordinates of the objects chosen from input.}\par
\detokenize{2. reasoning_and_reflection: Reason about the overall plan that needs to be taken on the target objects, and reflect on the previous actions taken if available.}\par
\detokenize{3. language_plan: A list of natural language actions to achieve the user instruction. Each language action is started by the step number and the language action name.}\par
\detokenize{4. executable_plan: A list of discrete actions needed to achieve the user instruction, with each discrete action being a 7-dimensional discrete action.}\par
\detokenize{5. keep your plan efficient and concise.}\par
\detokenize{!! When generating content for JSON strings, avoid using any contractions or abbreviated forms (like 's, 're, 've, 'll, 'd, n't) that use apostrophes. Instead, write out full forms (is, are, have, will, would, not) to prevent parsing errors in JSON. Please do not output any other thing more than the above-mentioned JSON, do not include ```json and ```!!!.}\par
\end{tcolorbox}

% \paragraph{Tool first-pass template.}
\label{app:prompt_ebmanip_tool}
\begin{tcolorbox}[
  colback=gray!10,
  colframe=black,
  arc=1mm,
  boxrule=0.5mm,
  left=6pt,
  right=6pt,
  top=6pt,
  bottom=6pt,
  title=\textbf{EB-Manipulation Tool first-pass prompt},
  before skip=6pt,
  after skip=6pt,
  breakable
]
\small\ttfamily\noindent
\detokenize{Tool first-pass task prompt template, from tool_manip_planner.py::process_prompt}\par
\par
\detokenize{Current full runs use:}\par
\detokenize{- paper_align=True}\par
\detokenize{- tool_runtime_enabled=True}\par
\detokenize{- tool_prompt_hidden=True}\par
\par
\detokenize{Therefore the visible first-pass task prompt is effectively the paper-aligned baseline task prompt:}\par
\detokenize{{paper_aligned_system_prompt.format(VOXEL_SIZE, VOXEL_SIZE, rotation_bins, ROTATION_RESOLUTION, examples)}}\par
\par
\detokenize{## Now you are supposed to follow the above examples to generate a sequence of discrete gripper actions that completes the below human instruction.}\par
\detokenize{Human Instruction: {user_instruction}.}\par
\detokenize{Input: {avg_obj_coord}}\par
\detokenize{Output gripper actions:}\par
\par
\detokenize{If tool_prompt_hidden=False, build_tool_usage_prompt(enabled_tools, force_call=tool_debug_force_call) is appended after "Output gripper actions:". In current repeat-full runs it is hidden, so it is not visible to the model in the first prompt.}\par
\par
\detokenize{After first model response, the runtime may synthesize or normalize tool_calls, execute enabled tools, and either:}\par
\detokenize{1. append build_tool_result_prompt(tool_results) for a second model pass; or}\par
\detokenize{2. bypass the second model pass and use deterministic repair/guard logic.}\par
\par
\detokenize{The output json format should be {'visual_state_description':str, 'reasoning_and_reflection':str, 'language_plan':str, 'executable_plan':
List[List[number]]}}\par
\detokenize{The fields in above JSON follows the purpose below:}\par
\detokenize{1. visual_state_description: Describe the color and shape of each object in the detection box in the numerical order in the image. Then provide the 3D coordinates of the objects chosen from input.}\par
\detokenize{2. reasoning_and_reflection: Reason about the overall plan that needs to be taken on the target objects, and reflect on the previous actions taken if available.}\par
\detokenize{3. language_plan: A list of natural language actions to achieve the user instruction. Each language action is started by the step number and the language action name.}\par
\detokenize{4. executable_plan: A list of discrete actions needed to achieve the user instruction, with each discrete action being a 7-dimensional discrete action.}\par
\detokenize{5. keep your plan efficient and concise.}\par
\detokenize{6. Also include the tool fields reasoning, need_tool, tool_calls, and action_id when tool use is helpful. For manipulation, executable_plan remains the action source.}\par
\detokenize{7. CRITICAL FORMAT RULE: Your executable_plan MUST be a pure nested list of numbers, for example: [[1, 0, 0, 0, 0, 0, 0]]. DO NOT wrap the inner arrays in dictionaries or strings like {'action': '[...]'}.}\par
\detokenize{8. KINEMATICS WARNING: Tool results such as 2D detections, masks, or depth summaries are only grounding hints. You MUST NOT use 2D pixels or raw tool metadata directly as 3D (X, Y, Z) coordinates for the robotic arm. Your executable_plan MUST REMAIN a nested list of 7 numbers like [[X, Y, Z, Roll, Pitch, Yaw, Gripper]]. DO NOT invent new object schemas with keys like action_id or parameters.}\par
\detokenize{9. EMERGENCY FALLBACK RULE: If you cannot produce full JSON, output ONLY one bare 7-number array like [0.05, 0.0, -0.02, 0, 0, 0, 1]. Do not output any words before or after the array.}\par
\detokenize{!! When generating content for JSON strings, avoid using any contractions or abbreviated forms (like 's, 're, 've, 'll, 'd, n't) that use apostrophes. Instead, write out full forms (is, are, have, will, would, not) to prevent parsing errors in JSON. Please do not output any other thing more than the above-mentioned JSON, do not include ```json and ```!!!.}\par
\par
\detokenize{[NOTE] This template is concatenated with TOOL_PROMPT_GUIDE at runtime. See tool/05_TOOL_PROMPT_GUIDE.txt.}\par
\end{tcolorbox}

\subsubsection{Runtime Components}
\label{app:prompt_ebmanip_tool_runtime}
% \paragraph{Hidden tool-usage policy.}
\begin{tcolorbox}[
  colback=gray!10,
  colframe=black,
  arc=1mm,
  boxrule=0.5mm,
  left=6pt,
  right=6pt,
  top=6pt,
  bottom=6pt,
  title=\textbf{Hidden tool-use policy source text},
  before skip=6pt,
  after skip=6pt,
  breakable
]
\small\ttfamily\noindent
\detokenize{Tool usage policy:}\par
\detokenize{Enabled tools:}\par
\detokenize{- color_grounding: color-attribute grounding over candidate object boxes when same-shape objects have generic labels. Trigger Condition: Use ONLY WHEN the instruction names a color and multiple same-shape candidate boxes remain ambiguous, especially when grounding_repair only sees generic labels such as object 1/object 2. Do NOT call it when color is irrelevant or no candidate boxes are available.}\par
\detokenize{- grounding_repair: candidate target grounding repair over detector/object-coordinate proposals for noisy labels or source/destination swaps. Trigger Condition: Use ONLY WHEN target labels, detector outputs, or object-coordinate candidates conflict and you need a ranked target choice before acting.}\par
\detokenize{- path_feasibility: lightweight normalized-workspace and straight-line feasibility guard for planned manipulation actions. Trigger Condition: Use ONLY WHEN a proposed 7D action or target pose may be outside the normalized workspace, too large a jump, too low/high, or likely to need replanning before execution.}\par
\detokenize{- motion_repair: action-sequence sanitizer for malformed, repeated, or looping 7D manipulation trajectories. Trigger Condition: Use ONLY WHEN the executable_plan repeats nearly identical 7D actions, loops after failure, or contains malformed/out-of-range action values.}\par
\detokenize{Default to your own executable plan; call a tool only for a concrete unresolved issue or recovery after failed execution.}\par
\detokenize{You may call more than one tool in the same response when multiple checks are needed, but you MUST NOT exceed 3 tools in a single step.}\par
\detokenize{Only call tools when absolutely necessary, because unnecessary tool calls increase context load and can hurt planning quality.}\par
\detokenize{When the visual state is clear enough to produce a valid executable_plan, rely on the model plan and leave tools unused.}\par
\detokenize{For colored same-shape manipulation targets with generic object labels, call color_grounding with the current candidate boxes; use it only after ambiguity is concrete, and trust it only when confidence and top-2 margin are strong.}\par
\detokenize{grounding_repair, motion_repair, and path_feasibility are planning-support tools for manipulation only; call them with concrete candidates, executable_plan/actions, or target_pose/action arguments, and use their output as evidence for the second-pass plan rather than as a success signal.}\par
\detokenize{When using clahe_filter or feature_squeezer, you DO NOT need to provide an image_path. The system will automatically apply them to the current visual observation.}\par
\detokenize{For manipulation tasks that require executable_plan, include a conservative candidate executable_plan even when need_tool=true; the tool result will verify or refine it in the second pass.}\par
\detokenize{If you call a tool, the first response should request the tool. After tool results are returned, produce a second final JSON decision.}\par
\end{tcolorbox}

% \paragraph{Second-pass tool-result prompt template.}
\begin{tcolorbox}[
  colback=gray!10,
  colframe=black,
  arc=1mm,
  boxrule=0.5mm,
  left=6pt,
  right=6pt,
  top=6pt,
  bottom=6pt,
  title=\textbf{Second-pass tool-result prompt},
  before skip=6pt,
  after skip=6pt,
  breakable
]
\small\ttfamily\noindent
\detokenize{Tool results:}\par
\detokenize{{serialized_tool_results}}\par
\par
\detokenize{Use the tool results above to revise your reasoning. Now output the final JSON decision for the current step. If the tool changed your belief, update reasoning, language_plan, executable_plan, and action_id accordingly.}\par
\par
\detokenize{For manipulation runs, the generic build_tool_result_prompt branch is used by default, so no Habitat-specific or ALFRED-specific hard rule is automatically inserted. The runtime may additionally append runtime_guardrail text, or bypass the second pass entirely and use deterministic repair/guard logic.}\par
\end{tcolorbox}

\paragraph{Artifact availability.}
The second-pass prompt is constructed dynamically from the tool results returned at a given step. Consequently, the template below specifies the reusable prompt structure, while the concrete serialized tool outputs vary across episodes and steps.

\subsubsection{Run Configuration Summary}
\label{app:prompt_ebmanip_configs}
% \paragraph{Prompt-related configuration.}
\begin{tcolorbox}[
  colback=gray!10,
  colframe=black,
  arc=1mm,
  boxrule=0.5mm,
  left=6pt,
  right=6pt,
  top=6pt,
  bottom=6pt,
  title=\textbf{Prompt-related run configuration},
  before skip=6pt,
  after skip=6pt,
  breakable
]
\small\ttfamily\noindent
\detokenize{Common configuration fields: model_name, model_type, eval_sets, n_shots, detection_box, resolution, visual_icl, and selected task indexes.}\par
\medskip
\detokenize{No-tool setting: the model receives the paper-aligned manipulation prompt, the current task instruction, object coordinates, optional history, and the required JSON output schema.}\par
\medskip
\detokenize{Tool-enabled setting: the first-pass prompt remains paper-aligned; tool policies, tool calls, tool results, and repair or guard rules are applied by the runtime before producing the final executable plan.}\par
\end{tcolorbox}

\subsection{Tool-Evaluation Prompts}
\label{app:prompts_tool_eval}
The following prompts are used by the tool-awareness, tool-selection, tool-usage, and tool-composition evaluation scripts.
\subsubsection{Tool-Evaluation Prompt Overview}
\label{app:prompt_tool_eval_readme}
\begin{tcblisting}{
  colback=gray!10,
  colframe=black,
  arc=1mm,
  boxrule=0.5mm,
  left=6pt,
  right=6pt,
  top=6pt,
  bottom=6pt,
  title=\textbf{Tool-Evaluation Prompt Overview},
  before skip=6pt,
  after skip=6pt,
  breakable,
  listing only,
  listing options={basicstyle=\small\ttfamily,breaklines=true,columns=fullflexible}
}
# Embodied Tool Benchmark Evaluation Prompts

This bundle documents the prompts used by the tool-evaluation tasks.

Files:

- Task 1: Tool Awareness / Tool-Need Recognition.
- Task 2: Tool Selection Accuracy.
- Task 3: Tool Usage, two-stage tool invocation and tool-conditioned action.
- Task 4: Tool Composition.

Notes:

- The model receives the listed system prompt plus a JSON/text user message and observation images.
- Hidden labels, ground truth, metadata used only for scoring, and evaluation metrics are not exposed to the model.
- For Task 3, the evaluation code uses dynamic per-tool prompts generated by `build_stage1_system_prompt()` and `build_stage2_system_prompt()`, not the older top-level constant prompt strings.

\end{tcblisting}
\subsubsection{Task 1: Tool-Need Recognition Prompt}
\label{app:prompt_tool_eval_task1_tool_need_recognition}
\begin{tcblisting}{
  colback=gray!10,
  colframe=black,
  arc=1mm,
  boxrule=0.5mm,
  left=6pt,
  right=6pt,
  top=6pt,
  bottom=6pt,
  title=\textbf{Task 1: Tool-Need Recognition Prompt},
  before skip=6pt,
  after skip=6pt,
  breakable,
  listing only,
  listing options={basicstyle=\small\ttfamily,breaklines=true,columns=fullflexible}
}
# Task 1: Tool-Need Recognition Prompt

## System Prompt

```text
You are a strict evaluation assistant. You will receive the current observation image, task, and interaction history. The available tool set is listed below (you can only choose from these tool_id values):
<TOOL_CATALOG_TEXT>

You must output JSON only, without any extra explanation.
Task: Determine whether the current step requires tool usage to improve decision quality.
Output format (strict):
{"need_tool": true} or {"need_tool": false}
Decision rule: If the current step has perception/localization uncertainty and cannot be solved reliably with current information alone, set need_tool=true; if current information is already sufficient and tools would not change the decision, set need_tool=false.
```

`<TOOL_CATALOG_TEXT>` is generated from the tool registry and contains the available `tool_id`, tool name, capability type, and trigger description.

## User Message Template

```text
sample_id: <sample_id>
task: <task>
env: <eval_env>, episode: <source_episode>, step: <source_step>
action_description: <current_or_recent_action_description>
env_feedback: <environment_feedback>
last_action_success: <true|false|null>
Please output the binary need_tool label as JSON.
Interaction history text follows (history images may also be provided):
history_1: <history text>
history_2: <history text>
...
```

History is included only when available and enabled by the evaluator.

## Images Provided

- Current observation image.
- Optional recent history images when enabled.

## Expected Output

```json
{"need_tool": true}
```

or

```json
{"need_tool": false}
```

\end{tcblisting}
\subsubsection{Task 2: Tool Selection Prompt}
\label{app:prompt_tool_eval_task2_tool_selection}
\begin{tcblisting}{
  colback=gray!10,
  colframe=black,
  arc=1mm,
  boxrule=0.5mm,
  left=6pt,
  right=6pt,
  top=6pt,
  bottom=6pt,
  title=\textbf{Task 2: Tool Selection Prompt},
  before skip=6pt,
  after skip=6pt,
  breakable,
  listing only,
  listing options={basicstyle=\small\ttfamily,breaklines=true,columns=fullflexible}
}
# Task 2: Tool Selection Prompt

## System Prompt

```text
You are a strict evaluation assistant for embodied-agent tool selection. You will receive a task, the current observation image, a concise state summary, optional sanitized interaction history, and exactly four candidate tool options.
Select exactly ONE minimally sufficient tool for the current decision step. Do not choose a tool because it may be useful later; choose the tool that resolves the current bottleneck. Only choose from the four candidate tool_id values shown in the user message.
Output JSON only, with this strict format:
{"selected_tool_ids": ["tool_x"]}
```

## User Message Template

For the current dataset format, the evaluator uses `sample["model_input"]` when present:

```text
task: <task>
current_state_summary: <current_state_summary>
Select exactly one minimally sufficient tool from the four candidate tools below.
candidate_tools:
A. <tool_id>: <option_description>
B. <tool_id>: <option_description>
C. <tool_id>: <option_description>
D. <tool_id>: <option_description>
sanitized_history:
history_1: <sanitized history text>
history_2: <sanitized history text>
...
{"selected_tool_ids": ["tool_x"]}
```

If `current_state_summary` or history is absent, those sections are omitted.

## Candidate Tool Requirement

The model must select from the four displayed candidate `tool_id` values only.

## Expected Output

```json
{"selected_tool_ids": ["tool_3"]}
```

The list must contain exactly one tool id.

\end{tcblisting}
\subsubsection{Task 3: Tool Usage Prompt}
\label{app:prompt_tool_eval_task3_tool_usage}
\begin{tcblisting}{
  colback=gray!10,
  colframe=black,
  arc=1mm,
  boxrule=0.5mm,
  left=6pt,
  right=6pt,
  top=6pt,
  bottom=6pt,
  title=\textbf{Task 3: Tool Usage Prompt},
  before skip=6pt,
  after skip=6pt,
  breakable,
  listing only,
  listing options={basicstyle=\small\ttfamily,breaklines=true,columns=fullflexible}
}
# Task 3: Tool Usage Prompt

### Stage 1 System Prompt, Common Part

```text
You are evaluating embodied-agent tool usage.
You receive a task, history, current observation, and exactly one designated tool with its input schema.

Your Stage 1 job is to construct ONE valid and task-relevant tool call for the designated tool.

General output rules:
1. Output JSON only. Do not output explanations, markdown, or extra text.
2. Use exactly this top-level structure:
   {"tool_call":{"tool_id":"...","arguments":{...}}}
3. The tool_id must be the designated tool_id.
4. The arguments object must include every required argument from the designated tool.
5. Do not rename argument keys. Use the keys exactly as defined in the input schema.
6. Keep argument value types consistent with the input schema.
7. When an argument requires the current image, use the current observation image path.
8. When an argument requires history images, use the available history image paths.
9. Tool arguments should contain only the information requested by the argument field. Do not add location, relation, action, or scene context unless that argument explicitly asks for it.
```

The evaluator appends one designated-tool-specific rule block.

### Stage 1 Tool-Specific Rules

#### `tool_3`: Open-Vocabulary Object Grounding / Detection

```text
Designated-tool input rules: open-vocabulary object grounding / detection.
Use this arguments format:
{"image":"<current_observation_image_path>","text_query":"<object_name>"}
The text_query must be only the target object name. Do not include location, relation, action, or scene context.
```

#### `tool_17`: Language-Grounded Navigation Scene Graph

```text
Designated-tool input rules: language-grounded navigation scene graph.
Use this arguments format:
{"current_image":"<current_observation_image_path>","history_images":["<history_image_path>",...],"goal_description":"<object_or_place_name>"}
The goal_description must be only the navigation target object/place name. Do not copy the full task instruction.
```

#### `tool_19`: Scene-Graph Memory Query

```text
Designated-tool input rules: scene-graph memory query.
Use this arguments format:
{"scene_graph":"scene_memory","object_query":"<object_name>","relation_query":"location"}
The scene_graph value is the benchmark memory resource name scene_memory. The relation_query should be location when recovering where the target object was last seen. The object_query must be only the target object name, written as natural lowercase words with spaces; for example, convert ToiletPaper to toilet paper.
```

#### `tool_51`: Goal-Pose Navigation

```text
Designated-tool input rules: goal-pose navigation.
Use this arguments format:
{"current_image":"<current_observation_image_path>","target_pose_hint":"near <target>","robot_state":{"base":"current_pose","gripper":"unchanged"}}
The target_pose_hint should be the short phrase near <target>, not a full action plan.
```

#### `tool_59`: Reactive Obstacle Avoidance

```text
Designated-tool input rules: reactive obstacle avoidance.
Use this arguments format:
{"current_image":"<current_observation_image_path>","blocked_motion":"<unsafe_or_blocked_motion>","goal_description":"<task_goal>"}
The blocked_motion must describe the motion that is unsafe or blocked.
```

#### `tool_75`: Grasp Planning

```text
Designated-tool input rules: grasp planning.
Use this arguments format:
{"current_image":"<current_observation_image_path>","target_object":"<object_to_grasp>","placement_goal":"<destination_or_container>","task_goal":"<task_goal>"}
The target_object must be only the object to grasp. Do not include placement words such as into, onto, place, put, or destination containers.
```

#### `tool_84`: Object Pose Estimation

```text
Designated-tool input rules: object pose estimation.
Use this arguments format:
{"current_image":"<current_observation_image_path>","target_objects":["<object_name>",...],"object_category":"<category>"}
The target_objects list must contain concrete object names only, not task verbs or abstract words such as manipulation, stacking, or wiping.
```

#### Fallback Rule

```text
Designated-tool input rules:
Follow the designated tool's input_schema exactly. Fill each required argument with concise task-relevant values.
```

### Stage 1 Final Instruction

```text
Do not assume access to hidden labels. Infer the concrete argument values only from the task, history, observation, designated tool description, and input schema.
```

### Stage 1 User Message Template

The evaluator serializes this JSON as the user message:

```json
{
  "sample_id": "<sample_id>",
  "task_instruction": "<task_instruction>",
  "history": [
    {
      "step": 0,
      "action": "<previous_action>",
      "feedback": "<feedback>",
      "image_path": "<history_image_path>"
    }
  ],
  "observation": {
    "rgb": "<current_observation_image_path>",
    "robot_state": {
      "gripper": "<open|closed|unknown>",
      "holding": "<object_or_null>"
    }
  },
  "designated_tool": {
    "tool_id": "<tool_id>",
    "tool_name": "<tool_name>",
    "description": "<tool_description>",
    "input_schema": {
      "<argument_name>": "<argument_type_or_description>"
    }
  },
  "output_schema": {
    "tool_call": {
      "tool_id": "<designated_tool_id>",
      "arguments": "<designated_tool_input_schema>"
    }
  }
}
```

### Stage 1 Expected Output

```json
{
  "tool_call": {
    "tool_id": "tool_3",
    "arguments": {
      "image": "images/usage_000001/observation.png",
      "text_query": "pencil"
    }
  }
}
```

## Stage 2: Tool-Conditioned Action

### Stage 2 System Prompt, Common Part

```text
You are evaluating embodied-agent tool-conditioned action.
You receive a task, history, current observation, the designated tool, the model's tool call, and the tool output.

Your Stage 2 job is to predict ONE next high-level embodied action that correctly uses the returned tool result.

General output rules:
1. Output JSON only. Do not output explanations, markdown, or extra text.
2. Use exactly this top-level structure:
   {"next_action":{"action_type":"...","target":"...","reference":{...}}}
3. action_type must be one of the allowed action_type values for the designated tool.
4. target must name the task-relevant object, area, or goal state for the next action.
5. reference must point to the relevant identifier, pose, mask, command, path, or other usable result returned by the tool.
6. Do not invent references that are not present in the tool output.
7. Do not ignore the tool output. The next action must be conditioned on the returned result.
8. Do not rely on confidence score alone. Inspect candidate fields such as confirmation, reachability, temporal validity, safety, feasibility, and pose confidence.
9. If the returned candidates do not contain a unique usable task-relevant result, do not execute a normal downstream action. Use the diagnostic token returned by the tool as the reference value.

Failure/uncertainty action formats:
- Multiple plausible target instances or unresolved grounded area: {"action_type":"inspect_target_area","target":"<task_target>","reference":{"ambiguity_ref":"<diagnostic_token>"}} or {"reference":{"failure_ref":"<diagnostic_token>"}}
- Stale memory or low-confidence pose result: {"action_type":"reject_invalid_tool_result","target":"<task_target>","reference":{"stale_ref":"<diagnostic_token>"}}
- Unsafe local motion, unreachable navigation target, or infeasible grasp set: {"action_type":"replan_from_tool_failure","target":"<task_target>","reference":{"failure_ref":"<diagnostic_token>"}}
- Tool output conflicts with the task target, tool call, history, or requested relation: {"action_type":"reject_invalid_tool_result","target":"<task_target>","reference":{"mismatch_ref":"<diagnostic_token>"}} or, for planner failures, {"action_type":"replan_from_tool_failure","target":"<task_target>","reference":{"mismatch_ref":"<diagnostic_token>"}}
```

The evaluator appends one designated-tool-specific action rule block.

### Stage 2 Tool-Specific Action Rules

#### `tool_3`: Object Grounding

```text
Designated-tool action rules: object grounding.
Allowed action_type values:
- ground_target_and_continue
- inspect_target_area
- reject_invalid_tool_result
If the tool output returns a valid target_ref for the task-relevant object, output:
{"action_type":"ground_target_and_continue","target":"<object_name>","reference":{"target_ref":"<returned_target_ref>"}}
Do not directly output downstream manipulation actions such as pick_up in this stage.
```

#### `tool_17`: Language-Grounded Navigation Scene Graph

```text
Designated-tool action rules: language-grounded navigation scene graph.
Allowed action_type values:
- navigate_to_grounded_area
- inspect_target_area
- reject_invalid_tool_result
If the tool output returns a grounded area reference for the navigation target, output:
{"action_type":"navigate_to_grounded_area","target":"<target_name>","reference":{"area_ref":"<returned_area_ref>"}}
```

#### `tool_19`: Scene-Graph Memory Query

```text
Designated-tool action rules: scene-graph memory query.
Allowed action_type values:
- navigate_to_recovered_location
- reject_invalid_tool_result
If the tool output returns a memory reference for a previously observed object or relation, output:
{"action_type":"navigate_to_recovered_location","target":"<object_name>","reference":{"memory_ref":"<returned_memory_ref>"}}
```

#### `tool_51`: Goal-Pose Navigation

```text
Designated-tool action rules: goal-pose navigation.
Allowed action_type values:
- execute_navigation_command
- replan_from_tool_failure
- reject_invalid_tool_result
If the tool output returns a navigation command reference, output:
{"action_type":"execute_navigation_command","target":"<target_name>","reference":{"command_ref":"<returned_command_ref>"}}
```

#### `tool_59`: Reactive Obstacle Avoidance

```text
Designated-tool action rules: reactive obstacle avoidance.
Allowed action_type values:
- execute_safe_local_motion
- replan_from_tool_failure
- reject_invalid_tool_result
If the tool output returns a safe local motion reference, output:
{"action_type":"execute_safe_local_motion","target":"<task_target>","reference":{"motion_ref":"<returned_motion_ref>"}}
```

#### `tool_75`: Grasp Planning

```text
Designated-tool action rules: grasp planning.
Allowed action_type values:
- execute_grasp
- replan_from_tool_failure
- reject_invalid_tool_result
If the tool output returns a grasp reference, output:
{"action_type":"execute_grasp","target":"<object_to_grasp>","reference":{"grasp_ref":"<returned_grasp_ref>"}}
```

#### `tool_84`: Pose Estimation

```text
Designated-tool action rules: pose estimation.
Allowed action_type values:
- align_manipulation_with_pose
- reject_invalid_tool_result
If the tool output returns one or more pose references, output:
{"action_type":"align_manipulation_with_pose","target":"<pose_target_names>","reference":{"pose_refs":["<returned_pose_ref>",...]}}
```

#### Fallback Rule

```text
Designated-tool action rules:
Use the tool output to choose a valid high-level action. The reference field must contain a returned tool identifier or structured result.
```

### Stage 2 Final Instruction

```text
Do not assume access to hidden labels. Infer the concrete next action only from the task, history, observation, designated tool, tool call, and tool output.
```

### Stage 2 User Message Template

The evaluator serializes this JSON as the user message:

```json
{
  "sample_id": "<sample_id>",
  "task_instruction": "<task_instruction>",
  "history": [
    {
      "step": 0,
      "action": "<previous_action>",
      "feedback": "<feedback>",
      "image_path": "<history_image_path>"
    }
  ],
  "observation": {
    "rgb": "<current_observation_image_path>",
    "robot_state": {
      "gripper": "<open|closed|unknown>",
      "holding": "<object_or_null>"
    }
  },
  "designated_tool": {
    "tool_id": "<tool_id>",
    "tool_name": "<tool_name>",
    "description": "<tool_description>",
    "input_schema": {
      "<argument_name>": "<argument_type_or_description>"
    },
    "output_schema": {
      "<output_name>": "<output_type_or_description>"
    }
  },
  "tool_call": {
    "tool_id": "<tool_id>",
    "arguments": {
      "<argument_name>": "<argument_value>"
    }
  },
  "tool_output": {
    "<tool_return_field>": "<tool_return_value>"
  },
  "output_schema": {
    "next_action": {
      "action_type": "string",
      "target": "string",
      "reference": "object"
    }
  }
}
```

### Stage 2 Expected Output

```json
{
  "next_action": {
    "action_type": "ground_target_and_continue",
    "target": "pencil",
    "reference": {
      "target_ref": "u001_det_c"
    }
  }
}
```

or, when the tool output is unusable/conflicting:

```json
{
  "next_action": {
    "action_type": "reject_invalid_tool_result",
    "target": "pencil",
    "reference": {
      "mismatch_ref": "u001_task_output_mismatch"
    }
  }
}
```

\end{tcblisting}
\subsubsection{Task 4: Tool Composition Prompt}
\label{app:prompt_tool_eval_task4_tool_composition}
\begin{tcblisting}{
  colback=gray!10,
  colframe=black,
  arc=1mm,
  boxrule=0.5mm,
  left=6pt,
  right=6pt,
  top=6pt,
  bottom=6pt,
  title=\textbf{Task 4: Tool Composition Prompt},
  before skip=6pt,
  after skip=6pt,
  breakable,
  listing only,
  listing options={basicstyle=\small\ttfamily,breaklines=true,columns=fullflexible}
}
# Task 4: Tool Composition Prompt

## System Prompt

```text
You are evaluating embodied-agent tool composition planning.
You receive a task instruction, interaction history, current observation, and a candidate list of available tools.

Your job is to output an ordered tool chain for solving the task.
Do not construct tool arguments.
Do not simulate tool outputs.
Do not predict robot actions.
Only choose the ordered sequence of tool IDs that should be composed.

Output JSON only:
{"predicted_tool_sequence":["tool_x","tool_y"]}

Rules:
1. Use only tool_id values from available_tools.
2. Include every tool that is necessary for solving the current task.
3. Do not include tools that are clearly irrelevant.
4. Order tools according to real dependencies: information-producing tools should appear before tools that consume that information.
5. If two tools can be used independently, choose a reasonable order, but do not add extra explanation.
```

## User Message Template

The evaluator serializes this model input as JSON:

```json
{
  "sample_id": "<sample_id>",
  "task_instruction": "<task_instruction>",
  "history": [
    {
      "step": 0,
      "action": "<previous_action>",
      "feedback": "<feedback>"
    }
  ],
  "observation": {
    "rgb": "<current_observation_image_path>",
    "history_rgb": ["<history_image_path>"]
  },
  "available_tools": [
    {
      "tool_id": "tool_x",
      "tool_name": "<tool_name>",
      "description": "<tool_description>"
    }
  ],
  "output_schema": {
    "predicted_tool_sequence": ["tool_id", "..."]
  }
}
```

Fields whose names start with `source_` and `visual_alignment_note` are removed from `observation` before prompting.

## Images Provided

- Current observation image.
- Optional history images when enabled.

## Expected Output

```json
{"predicted_tool_sequence": ["tool_3", "tool_84", "tool_75", "tool_51"]}
```

\end{tcblisting}
% END AUTO-INSERTED ADDITIONAL PROMPTS